%% file: main.tex
\documentclass[11pt,a4paper]{article}
\usepackage[T1]{fontenc}
\usepackage[utf8]{inputenc}
\usepackage{lmodern}
\usepackage[margin=25mm]{geometry}
\usepackage[numbers,sort&compress]{natbib}
\usepackage{authblk}

\input{preamble}

\title{\papertitle}
\ifanonymous
\author{Anonymised for review}
\else
\author[1,2]{Martin Pavlíček\thanks{Corresponding author: \texttt{martin.pavlicek@osu.cz}}}
\author[2]{Martin Kotyrba}
\author[2]{Eva Volná}
\author[1,2]{Tomáš Filip}
\author[1]{Petr Sosík}
\affil[1]{Institute for Research and Applications of Fuzzy Modeling, University of Ostrava, 30.~dubna 22, 701~03 Ostrava, Czech Republic}
\affil[2]{Department of Informatics and Computers, Faculty of Science, University of Ostrava, 30.~dubna 22, 701~03 Ostrava, Czech Republic}
\fi
\date{November 2026}

\begin{document}
\maketitle
\input{sections/abstract}


\input{body}

\bibliographystyle{unsrtnat}
\input{back}

\end{document}

%% file: preamble.tex
\usepackage{microtype}
\usepackage{amsmath,amssymb}
\usepackage{booktabs,tabularx,multirow}
\usepackage{graphicx}
\usepackage{xcolor}
\usepackage[hidelinks]{hyperref}
\usepackage{url}
\newcommand{\nei}{\textsc{nei}}


\makeatletter
\newcommand{\val}[1]{\ifcsname val@#1\endcsname\csname val@#1\endcsname\else\PackageError{main}{Unknown generated value: #1}{}\fi}
\makeatother
\input{generated/values}
\newif\ifshowpending
\showpendingfalse

\newcommand{\papertitle}{Where Do Fact-Checking Errors Arise?\\Separating Retrieval, Search Control and Judgement\\in Evidence-Grounded Verification}
\newif\ifanonymous
\ifdefined\anonymousbuild\anonymoustrue\fi
\hypersetup{pdftitle={Where Do Fact-Checking Errors Arise? Separating Retrieval, Search Control and Judgement in Evidence-Grounded Verification}}
\ifanonymous\else\hypersetup{pdfauthor={Martin Pavlíček, Martin Kotyrba, Eva Volná, Tomáš Filip, Petr Sosík}}\fi

%% file: generated/values.tex
\expandafter\def\csname val@agent.first_host.full.accuracy.a\endcsname{42.0}
\expandafter\def\csname val@agent.first_host.full.accuracy.f\endcsname{66.0}
\expandafter\def\csname val@agent.first_host.full.accuracy.ft\endcsname{51.6}
\expandafter\def\csname val@agent.first_host.full.calls_per_claim.a\endcsname{10.9}
\expandafter\def\csname val@agent.first_host.full.calls_per_claim.f\endcsname{7.7}
\expandafter\def\csname val@agent.first_host.full.calls_per_claim.ft\endcsname{9.2}
\expandafter\def\csname val@agent.first_host.full.damages.a\endcsname{7}
\expandafter\def\csname val@agent.first_host.full.damages.f\endcsname{3}
\expandafter\def\csname val@agent.first_host.full.damages.ft\endcsname{11}
\expandafter\def\csname val@agent.first_host.full.deficit.a\endcsname{6.0}
\expandafter\def\csname val@agent.first_host.full.deficit.f\endcsname{4.0}
\expandafter\def\csname val@agent.first_host.full.deficit.ft\endcsname{5.4}
\expandafter\def\csname val@agent.first_host.full.difference.a\endcsname{$+6.0$}
\expandafter\def\csname val@agent.first_host.full.difference.f\endcsname{$+4.0$}
\expandafter\def\csname val@agent.first_host.full.difference.ft\endcsname{$-5.4$}
\expandafter\def\csname val@agent.first_host.full.failures.a\endcsname{4}
\expandafter\def\csname val@agent.first_host.full.failures.all\endcsname{11}
\expandafter\def\csname val@agent.first_host.full.failures.f\endcsname{0}
\expandafter\def\csname val@agent.first_host.full.failures.ft\endcsname{7}
\expandafter\def\csname val@agent.first_host.full.high.a\endcsname{$+15.0$}
\expandafter\def\csname val@agent.first_host.full.high.f\endcsname{$+10.6$}
\expandafter\def\csname val@agent.first_host.full.high.ft\endcsname{$+4.9$}
\expandafter\def\csname val@agent.first_host.full.low.a\endcsname{$-3.0$}
\expandafter\def\csname val@agent.first_host.full.low.f\endcsname{$-2.0$}
\expandafter\def\csname val@agent.first_host.full.low.ft\endcsname{$-15.8$}
\expandafter\def\csname val@agent.first_host.full.repairs.a\endcsname{13}
\expandafter\def\csname val@agent.first_host.full.repairs.f\endcsname{7}
\expandafter\def\csname val@agent.first_host.full.repairs.ft\endcsname{6}
\expandafter\def\csname val@agent.first_host.interaction.a\endcsname{$+10.0$}
\expandafter\def\csname val@agent.first_host.interaction.f\endcsname{$+5.0$}
\expandafter\def\csname val@agent.first_host.interaction.ft\endcsname{$+0.0$}
\expandafter\def\csname val@agent.first_host.interaction.high.a\endcsname{$+21.0$}
\expandafter\def\csname val@agent.first_host.interaction.high.f\endcsname{$+13.7$}
\expandafter\def\csname val@agent.first_host.interaction.high.ft\endcsname{$+12.1$}
\expandafter\def\csname val@agent.first_host.interaction.low.a\endcsname{$+0.0$}
\expandafter\def\csname val@agent.first_host.interaction.low.f\endcsname{$-3.3$}
\expandafter\def\csname val@agent.first_host.interaction.low.ft\endcsname{$-12.6$}
\expandafter\def\csname val@agent.first_host.interaction.points.a\endcsname{10.0}
\expandafter\def\csname val@agent.first_host.interaction.points.f\endcsname{5.0}
\expandafter\def\csname val@agent.first_host.interaction.points.ft\endcsname{0.0}
\expandafter\def\csname val@agent.first_host.selected.accuracy.a\endcsname{61.0}
\expandafter\def\csname val@agent.first_host.selected.accuracy.f\endcsname{80.0}
\expandafter\def\csname val@agent.first_host.selected.accuracy.ft\endcsname{68.8}
\expandafter\def\csname val@agent.first_host.selected.calls_per_claim.a\endcsname{4.4}
\expandafter\def\csname val@agent.first_host.selected.calls_per_claim.f\endcsname{6.6}
\expandafter\def\csname val@agent.first_host.selected.calls_per_claim.ft\endcsname{7.0}
\expandafter\def\csname val@agent.first_host.selected.damages.a\endcsname{6}
\expandafter\def\csname val@agent.first_host.selected.damages.f\endcsname{5}
\expandafter\def\csname val@agent.first_host.selected.damages.ft\endcsname{10}
\expandafter\def\csname val@agent.first_host.selected.deficit.a\endcsname{4.0}
\expandafter\def\csname val@agent.first_host.selected.deficit.f\endcsname{1.0}
\expandafter\def\csname val@agent.first_host.selected.deficit.ft\endcsname{5.4}
\expandafter\def\csname val@agent.first_host.selected.difference.a\endcsname{$-4.0$}
\expandafter\def\csname val@agent.first_host.selected.difference.f\endcsname{$-1.0$}
\expandafter\def\csname val@agent.first_host.selected.difference.ft\endcsname{$-5.4$}
\expandafter\def\csname val@agent.first_host.selected.failures.a\endcsname{5}
\expandafter\def\csname val@agent.first_host.selected.failures.all\endcsname{14}
\expandafter\def\csname val@agent.first_host.selected.failures.f\endcsname{4}
\expandafter\def\csname val@agent.first_host.selected.failures.ft\endcsname{5}
\expandafter\def\csname val@agent.first_host.selected.high.a\endcsname{$+1.0$}
\expandafter\def\csname val@agent.first_host.selected.high.f\endcsname{$+5.0$}
\expandafter\def\csname val@agent.first_host.selected.high.ft\endcsname{$+2.4$}
\expandafter\def\csname val@agent.first_host.selected.low.a\endcsname{$-10.0$}
\expandafter\def\csname val@agent.first_host.selected.low.f\endcsname{$-6.8$}
\expandafter\def\csname val@agent.first_host.selected.low.ft\endcsname{$-13.2$}
\expandafter\def\csname val@agent.first_host.selected.repairs.a\endcsname{2}
\expandafter\def\csname val@agent.first_host.selected.repairs.f\endcsname{4}
\expandafter\def\csname val@agent.first_host.selected.repairs.ft\endcsname{5}
\expandafter\def\csname val@agent.first_host.store_loss.a\endcsname{19.0}
\expandafter\def\csname val@agent.first_host.store_loss.f\endcsname{14.0}
\expandafter\def\csname val@agent.first_host.store_loss.ft\endcsname{17.2}
\expandafter\def\csname val@agent.full.accuracy.a\endcsname{42.0}
\expandafter\def\csname val@agent.full.accuracy.f\endcsname{66.0}
\expandafter\def\csname val@agent.full.accuracy.ft\endcsname{51.6}
\expandafter\def\csname val@agent.full.calls.a\endcsname{1,088}
\expandafter\def\csname val@agent.full.calls.all\endcsname{2,720}
\expandafter\def\csname val@agent.full.calls.f\endcsname{773}
\expandafter\def\csname val@agent.full.calls.ft\endcsname{859}
\expandafter\def\csname val@agent.full.calls_per_claim.a\endcsname{10.9}
\expandafter\def\csname val@agent.full.calls_per_claim.f\endcsname{7.7}
\expandafter\def\csname val@agent.full.calls_per_claim.ft\endcsname{9.2}
\expandafter\def\csname val@agent.full.claims.wikipedia\endcsname{193}
\expandafter\def\csname val@agent.full.claims_with_rerank.a\endcsname{1}
\expandafter\def\csname val@agent.full.claims_with_rerank.all\endcsname{2}
\expandafter\def\csname val@agent.full.claims_with_rerank.f\endcsname{0}
\expandafter\def\csname val@agent.full.claims_with_rerank.ft\endcsname{1}
\expandafter\def\csname val@agent.full.damages.a\endcsname{7}
\expandafter\def\csname val@agent.full.damages.f\endcsname{3}
\expandafter\def\csname val@agent.full.damages.ft\endcsname{11}
\expandafter\def\csname val@agent.full.deficit.a\endcsname{6.0}
\expandafter\def\csname val@agent.full.deficit.f\endcsname{4.0}
\expandafter\def\csname val@agent.full.deficit.ft\endcsname{5.4}
\expandafter\def\csname val@agent.full.difference.a\endcsname{$+6.0$}
\expandafter\def\csname val@agent.full.difference.f\endcsname{$+4.0$}
\expandafter\def\csname val@agent.full.difference.ft\endcsname{$-5.4$}
\expandafter\def\csname val@agent.full.failures.a\endcsname{4}
\expandafter\def\csname val@agent.full.failures.all\endcsname{11}
\expandafter\def\csname val@agent.full.failures.f\endcsname{0}
\expandafter\def\csname val@agent.full.failures.ft\endcsname{7}
\expandafter\def\csname val@agent.full.high.a\endcsname{$+15.0$}
\expandafter\def\csname val@agent.full.high.f\endcsname{$+10.6$}
\expandafter\def\csname val@agent.full.high.ft\endcsname{$+4.9$}
\expandafter\def\csname val@agent.full.incomplete.a\endcsname{76}
\expandafter\def\csname val@agent.full.incomplete.all\endcsname{181}
\expandafter\def\csname val@agent.full.incomplete.f\endcsname{46}
\expandafter\def\csname val@agent.full.incomplete.ft\endcsname{59}
\expandafter\def\csname val@agent.full.low.a\endcsname{$-3.0$}
\expandafter\def\csname val@agent.full.low.f\endcsname{$-2.0$}
\expandafter\def\csname val@agent.full.low.ft\endcsname{$-15.8$}
\expandafter\def\csname val@agent.full.passages.a\endcsname{7.5}
\expandafter\def\csname val@agent.full.passages.all\endcsname{6.2}
\expandafter\def\csname val@agent.full.passages.f\endcsname{5.3}
\expandafter\def\csname val@agent.full.passages.ft\endcsname{5.8}
\expandafter\def\csname val@agent.full.repairs.a\endcsname{13}
\expandafter\def\csname val@agent.full.repairs.f\endcsname{7}
\expandafter\def\csname val@agent.full.repairs.ft\endcsname{6}
\expandafter\def\csname val@agent.full.stop.controller_budget.a\endcsname{46}
\expandafter\def\csname val@agent.full.stop.controller_budget.all\endcsname{103}
\expandafter\def\csname val@agent.full.stop.controller_budget.f\endcsname{22}
\expandafter\def\csname val@agent.full.stop.controller_budget.ft\endcsname{35}
\expandafter\def\csname val@agent.full.stop.finished.a\endcsname{29}
\expandafter\def\csname val@agent.full.stop.finished.all\endcsname{119}
\expandafter\def\csname val@agent.full.stop.finished.f\endcsname{54}
\expandafter\def\csname val@agent.full.stop.finished.ft\endcsname{36}
\expandafter\def\csname val@agent.full.stop.repeated_action.a\endcsname{25}
\expandafter\def\csname val@agent.full.stop.repeated_action.all\endcsname{71}
\expandafter\def\csname val@agent.full.stop.repeated_action.f\endcsname{24}
\expandafter\def\csname val@agent.full.stop.repeated_action.ft\endcsname{22}
\expandafter\def\csname val@agent.full.stop.repeated_action.wikipedia\endcsname{46}
\expandafter\def\csname val@agent.interaction.a\endcsname{$+6.0$}
\expandafter\def\csname val@agent.interaction.f\endcsname{$+5.0$}
\expandafter\def\csname val@agent.interaction.ft\endcsname{$+2.2$}
\expandafter\def\csname val@agent.interaction.high.a\endcsname{$+16.0$}
\expandafter\def\csname val@agent.interaction.high.f\endcsname{$+13.3$}
\expandafter\def\csname val@agent.interaction.high.ft\endcsname{$+14.7$}
\expandafter\def\csname val@agent.interaction.low.a\endcsname{$-3.0$}
\expandafter\def\csname val@agent.interaction.low.f\endcsname{$-3.5$}
\expandafter\def\csname val@agent.interaction.low.ft\endcsname{$-11.2$}
\expandafter\def\csname val@agent.interaction.points.a\endcsname{6.0}
\expandafter\def\csname val@agent.interaction.points.f\endcsname{5.0}
\expandafter\def\csname val@agent.interaction.points.ft\endcsname{2.2}
\expandafter\def\csname val@agent.selected.accuracy.a\endcsname{65.0}
\expandafter\def\csname val@agent.selected.accuracy.f\endcsname{80.0}
\expandafter\def\csname val@agent.selected.accuracy.ft\endcsname{66.7}
\expandafter\def\csname val@agent.selected.calls.a\endcsname{434}
\expandafter\def\csname val@agent.selected.calls.all\endcsname{1,744}
\expandafter\def\csname val@agent.selected.calls.f\endcsname{644}
\expandafter\def\csname val@agent.selected.calls.ft\endcsname{666}
\expandafter\def\csname val@agent.selected.calls_per_claim.a\endcsname{4.3}
\expandafter\def\csname val@agent.selected.calls_per_claim.f\endcsname{6.4}
\expandafter\def\csname val@agent.selected.calls_per_claim.ft\endcsname{7.2}
\expandafter\def\csname val@agent.selected.claims.wikipedia\endcsname{193}
\expandafter\def\csname val@agent.selected.claims_with_rerank.a\endcsname{1}
\expandafter\def\csname val@agent.selected.claims_with_rerank.all\endcsname{2}
\expandafter\def\csname val@agent.selected.claims_with_rerank.f\endcsname{1}
\expandafter\def\csname val@agent.selected.claims_with_rerank.ft\endcsname{0}
\expandafter\def\csname val@agent.selected.damages.a\endcsname{2}
\expandafter\def\csname val@agent.selected.damages.f\endcsname{5}
\expandafter\def\csname val@agent.selected.damages.ft\endcsname{11}
\expandafter\def\csname val@agent.selected.deficit.a\endcsname{0.0}
\expandafter\def\csname val@agent.selected.deficit.f\endcsname{1.0}
\expandafter\def\csname val@agent.selected.deficit.ft\endcsname{7.5}
\expandafter\def\csname val@agent.selected.difference.a\endcsname{$+0.0$}
\expandafter\def\csname val@agent.selected.difference.f\endcsname{$-1.0$}
\expandafter\def\csname val@agent.selected.difference.ft\endcsname{$-7.5$}
\expandafter\def\csname val@agent.selected.failures.a\endcsname{2}
\expandafter\def\csname val@agent.selected.failures.all\endcsname{12}
\expandafter\def\csname val@agent.selected.failures.f\endcsname{3}
\expandafter\def\csname val@agent.selected.failures.ft\endcsname{7}
\expandafter\def\csname val@agent.selected.high.a\endcsname{$+4.0$}
\expandafter\def\csname val@agent.selected.high.f\endcsname{$+5.4$}
\expandafter\def\csname val@agent.selected.high.ft\endcsname{$+1.1$}
\expandafter\def\csname val@agent.selected.incomplete.a\endcsname{31}
\expandafter\def\csname val@agent.selected.incomplete.all\endcsname{146}
\expandafter\def\csname val@agent.selected.incomplete.f\endcsname{53}
\expandafter\def\csname val@agent.selected.incomplete.ft\endcsname{62}
\expandafter\def\csname val@agent.selected.low.a\endcsname{$-4.0$}
\expandafter\def\csname val@agent.selected.low.f\endcsname{$-7.4$}
\expandafter\def\csname val@agent.selected.low.ft\endcsname{$-16.0$}
\expandafter\def\csname val@agent.selected.passages.a\endcsname{2.5}
\expandafter\def\csname val@agent.selected.passages.all\endcsname{2.1}
\expandafter\def\csname val@agent.selected.passages.f\endcsname{1.9}
\expandafter\def\csname val@agent.selected.passages.ft\endcsname{2.0}
\expandafter\def\csname val@agent.selected.repairs.a\endcsname{2}
\expandafter\def\csname val@agent.selected.repairs.f\endcsname{4}
\expandafter\def\csname val@agent.selected.repairs.ft\endcsname{4}
\expandafter\def\csname val@agent.selected.stop.controller_budget.a\endcsname{0}
\expandafter\def\csname val@agent.selected.stop.controller_budget.all\endcsname{11}
\expandafter\def\csname val@agent.selected.stop.controller_budget.f\endcsname{7}
\expandafter\def\csname val@agent.selected.stop.controller_budget.ft\endcsname{4}
\expandafter\def\csname val@agent.selected.stop.finished.a\endcsname{100}
\expandafter\def\csname val@agent.selected.stop.finished.all\endcsname{183}
\expandafter\def\csname val@agent.selected.stop.finished.f\endcsname{49}
\expandafter\def\csname val@agent.selected.stop.finished.ft\endcsname{34}
\expandafter\def\csname val@agent.selected.stop.repeated_action.a\endcsname{0}
\expandafter\def\csname val@agent.selected.stop.repeated_action.all\endcsname{99}
\expandafter\def\csname val@agent.selected.stop.repeated_action.f\endcsname{44}
\expandafter\def\csname val@agent.selected.stop.repeated_action.ft\endcsname{55}
\expandafter\def\csname val@agent.selected.stop.repeated_action.wikipedia\endcsname{99}
\expandafter\def\csname val@agent.store_loss.a\endcsname{23.0}
\expandafter\def\csname val@agent.store_loss.f\endcsname{14.0}
\expandafter\def\csname val@agent.store_loss.ft\endcsname{15.1}
\expandafter\def\csname val@fresh.agent.accuracy.all\endcsname{54.2}
\expandafter\def\csname val@fresh.agent.accuracy.f\endcsname{65.3}
\expandafter\def\csname val@fresh.agent.accuracy.ft\endcsname{49.4}
\expandafter\def\csname val@fresh.agent.calls_per_claim.all\endcsname{8.8}
\expandafter\def\csname val@fresh.agent.calls_per_claim.f\endcsname{8.6}
\expandafter\def\csname val@fresh.agent.calls_per_claim.ft\endcsname{8.9}
\expandafter\def\csname val@fresh.agent.correct.nei.all\endcsname{90}
\expandafter\def\csname val@fresh.agent.correct.nei.f\endcsname{24}
\expandafter\def\csname val@fresh.agent.correct.nei.ft\endcsname{66}
\expandafter\def\csname val@fresh.agent.correct.refuted.all\endcsname{139}
\expandafter\def\csname val@fresh.agent.correct.refuted.f\endcsname{57}
\expandafter\def\csname val@fresh.agent.correct.refuted.ft\endcsname{82}
\expandafter\def\csname val@fresh.agent.correct.supported.all\endcsname{158}
\expandafter\def\csname val@fresh.agent.correct.supported.f\endcsname{60}
\expandafter\def\csname val@fresh.agent.correct.supported.ft\endcsname{98}
\expandafter\def\csname val@fresh.agent.damages.all\endcsname{86}
\expandafter\def\csname val@fresh.agent.damages.f\endcsname{16}
\expandafter\def\csname val@fresh.agent.damages.ft\endcsname{70}
\expandafter\def\csname val@fresh.agent.deficit.all\endcsname{1.5}
\expandafter\def\csname val@fresh.agent.deficit.f\endcsname{1.9}
\expandafter\def\csname val@fresh.agent.deficit.ft\endcsname{3.0}
\expandafter\def\csname val@fresh.agent.difference.all\endcsname{$-1.5$}
\expandafter\def\csname val@fresh.agent.difference.f\endcsname{$+1.9$}
\expandafter\def\csname val@fresh.agent.difference.ft\endcsname{$-3.0$}
\expandafter\def\csname val@fresh.agent.failures.all\endcsname{35}
\expandafter\def\csname val@fresh.agent.failures.f\endcsname{4}
\expandafter\def\csname val@fresh.agent.failures.ft\endcsname{31}
\expandafter\def\csname val@fresh.agent.high.all\endcsname{$+2.1$}
\expandafter\def\csname val@fresh.agent.high.f\endcsname{$+8.1$}
\expandafter\def\csname val@fresh.agent.high.ft\endcsname{$+1.6$}
\expandafter\def\csname val@fresh.agent.low.all\endcsname{$-5.2$}
\expandafter\def\csname val@fresh.agent.low.f\endcsname{$-4.0$}
\expandafter\def\csname val@fresh.agent.low.ft\endcsname{$-7.7$}
\expandafter\def\csname val@fresh.agent.packet.f\endcsname{81}
\expandafter\def\csname val@fresh.agent.packet.ft\endcsname{152}
\expandafter\def\csname val@fresh.agent.pool.f\endcsname{101}
\expandafter\def\csname val@fresh.agent.pool.ft\endcsname{200}
\expandafter\def\csname val@fresh.agent.repairs.all\endcsname{75}
\expandafter\def\csname val@fresh.agent.repairs.f\endcsname{20}
\expandafter\def\csname val@fresh.agent.repairs.ft\endcsname{55}
\expandafter\def\csname val@fresh.agent.stated_correct.all\endcsname{11}
\expandafter\def\csname val@fresh.claims.all\endcsname{714}
\expandafter\def\csname val@fresh.claims.f\endcsname{216}
\expandafter\def\csname val@fresh.claims.ft\endcsname{498}
\expandafter\def\csname val@fresh.class.nei.all\endcsname{238}
\expandafter\def\csname val@fresh.class.nei.f\endcsname{72}
\expandafter\def\csname val@fresh.class.nei.ft\endcsname{166}
\expandafter\def\csname val@fresh.class.refuted.all\endcsname{238}
\expandafter\def\csname val@fresh.class.refuted.f\endcsname{72}
\expandafter\def\csname val@fresh.class.refuted.ft\endcsname{166}
\expandafter\def\csname val@fresh.class.supported.all\endcsname{238}
\expandafter\def\csname val@fresh.class.supported.f\endcsname{72}
\expandafter\def\csname val@fresh.class.supported.ft\endcsname{166}
\expandafter\def\csname val@fresh.evidence_bearing.f\endcsname{144}
\expandafter\def\csname val@fresh.evidence_bearing.ft\endcsname{344}
\expandafter\def\csname val@fresh.families.all\endcsname{527}
\expandafter\def\csname val@fresh.families.f\endcsname{89}
\expandafter\def\csname val@fresh.families.ft\endcsname{438}
\expandafter\def\csname val@fresh.identical_replies.all\endcsname{714}
\expandafter\def\csname val@fresh.identical_replies.f\endcsname{216}
\expandafter\def\csname val@fresh.identical_replies.ft\endcsname{498}
\expandafter\def\csname val@fresh.single.accuracy.all\endcsname{55.7}
\expandafter\def\csname val@fresh.single.accuracy.f\endcsname{63.4}
\expandafter\def\csname val@fresh.single.accuracy.ft\endcsname{52.4}
\expandafter\def\csname val@fresh.single.calls_per_claim.all\endcsname{1.0}
\expandafter\def\csname val@fresh.single.calls_per_claim.f\endcsname{1.0}
\expandafter\def\csname val@fresh.single.calls_per_claim.ft\endcsname{1.0}
\expandafter\def\csname val@fresh.single.correct.nei.all\endcsname{67}
\expandafter\def\csname val@fresh.single.correct.nei.f\endcsname{13}
\expandafter\def\csname val@fresh.single.correct.nei.ft\endcsname{54}
\expandafter\def\csname val@fresh.single.correct.refuted.all\endcsname{160}
\expandafter\def\csname val@fresh.single.correct.refuted.f\endcsname{59}
\expandafter\def\csname val@fresh.single.correct.refuted.ft\endcsname{101}
\expandafter\def\csname val@fresh.single.correct.supported.all\endcsname{171}
\expandafter\def\csname val@fresh.single.correct.supported.f\endcsname{65}
\expandafter\def\csname val@fresh.single.correct.supported.ft\endcsname{106}
\expandafter\def\csname val@fresh.single.failures.all\endcsname{38}
\expandafter\def\csname val@fresh.single.failures.f\endcsname{5}
\expandafter\def\csname val@fresh.single.failures.ft\endcsname{33}
\expandafter\def\csname val@fresh.single.packet.f\endcsname{108}
\expandafter\def\csname val@fresh.single.packet.ft\endcsname{201}
\expandafter\def\csname val@fresh.single.pool.f\endcsname{111}
\expandafter\def\csname val@fresh.single.pool.ft\endcsname{227}
\expandafter\def\csname val@fresh.single.stated_correct.all\endcsname{14}
\expandafter\def\csname val@fresh.verdict_changed.all\endcsname{0}
\expandafter\def\csname val@fresh.verdict_changed.f\endcsname{0}
\expandafter\def\csname val@fresh.verdict_changed.ft\endcsname{0}
\expandafter\def\csname val@host.changed_correctness.a\endcsname{10}
\expandafter\def\csname val@host.changed_correctness.all\endcsname{18}
\expandafter\def\csname val@host.changed_correctness.f\endcsname{2}
\expandafter\def\csname val@host.changed_correctness.ft\endcsname{6}
\expandafter\def\csname val@host.claims.a\endcsname{100}
\expandafter\def\csname val@host.claims.all\endcsname{293}
\expandafter\def\csname val@host.claims.f\endcsname{100}
\expandafter\def\csname val@host.claims.ft\endcsname{93}
\expandafter\def\csname val@host.correct_to_wrong.a\endcsname{3}
\expandafter\def\csname val@host.correct_to_wrong.all\endcsname{8}
\expandafter\def\csname val@host.correct_to_wrong.f\endcsname{1}
\expandafter\def\csname val@host.correct_to_wrong.ft\endcsname{4}
\expandafter\def\csname val@host.different_request.a\endcsname{16}
\expandafter\def\csname val@host.different_request.all\endcsname{68}
\expandafter\def\csname val@host.different_request.f\endcsname{26}
\expandafter\def\csname val@host.different_request.ft\endcsname{26}
\expandafter\def\csname val@host.first_accuracy.a\endcsname{61.0}
\expandafter\def\csname val@host.first_accuracy.all\endcsname{70.0}
\expandafter\def\csname val@host.first_accuracy.f\endcsname{80.0}
\expandafter\def\csname val@host.first_accuracy.ft\endcsname{68.8}
\expandafter\def\csname val@host.first_correct.a\endcsname{61}
\expandafter\def\csname val@host.first_correct.all\endcsname{205}
\expandafter\def\csname val@host.first_correct.f\endcsname{80}
\expandafter\def\csname val@host.first_correct.ft\endcsname{64}
\expandafter\def\csname val@host.first_failures.a\endcsname{5}
\expandafter\def\csname val@host.first_failures.all\endcsname{14}
\expandafter\def\csname val@host.first_failures.f\endcsname{4}
\expandafter\def\csname val@host.first_failures.ft\endcsname{5}
\expandafter\def\csname val@host.identical_final_request.a\endcsname{84}
\expandafter\def\csname val@host.identical_final_request.all\endcsname{225}
\expandafter\def\csname val@host.identical_final_request.f\endcsname{74}
\expandafter\def\csname val@host.identical_final_request.ft\endcsname{67}
\expandafter\def\csname val@host.mcnemar_p.a\endcsname{0.344}
\expandafter\def\csname val@host.mcnemar_p.all\endcsname{0.815}
\expandafter\def\csname val@host.mcnemar_p.f\endcsname{1.000}
\expandafter\def\csname val@host.mcnemar_p.ft\endcsname{0.688}
\expandafter\def\csname val@host.second_accuracy.a\endcsname{65.0}
\expandafter\def\csname val@host.second_accuracy.all\endcsname{70.6}
\expandafter\def\csname val@host.second_accuracy.f\endcsname{80.0}
\expandafter\def\csname val@host.second_accuracy.ft\endcsname{66.7}
\expandafter\def\csname val@host.second_correct.a\endcsname{65}
\expandafter\def\csname val@host.second_correct.all\endcsname{207}
\expandafter\def\csname val@host.second_correct.f\endcsname{80}
\expandafter\def\csname val@host.second_correct.ft\endcsname{62}
\expandafter\def\csname val@host.second_failures.a\endcsname{2}
\expandafter\def\csname val@host.second_failures.all\endcsname{12}
\expandafter\def\csname val@host.second_failures.f\endcsname{3}
\expandafter\def\csname val@host.second_failures.ft\endcsname{7}
\expandafter\def\csname val@host.verdict_changed.a\endcsname{16}
\expandafter\def\csname val@host.verdict_changed.all\endcsname{28}
\expandafter\def\csname val@host.verdict_changed.f\endcsname{4}
\expandafter\def\csname val@host.verdict_changed.ft\endcsname{8}
\expandafter\def\csname val@host.verdict_changed_on_identical_request.a\endcsname{9}
\expandafter\def\csname val@host.verdict_changed_on_identical_request.all\endcsname{16}
\expandafter\def\csname val@host.verdict_changed_on_identical_request.f\endcsname{2}
\expandafter\def\csname val@host.verdict_changed_on_identical_request.ft\endcsname{5}
\expandafter\def\csname val@host.wrong_to_correct.a\endcsname{7}
\expandafter\def\csname val@host.wrong_to_correct.all\endcsname{10}
\expandafter\def\csname val@host.wrong_to_correct.f\endcsname{1}
\expandafter\def\csname val@host.wrong_to_correct.ft\endcsname{2}
\expandafter\def\csname val@many.adaptive_gate.changes_on_identical_request.all\endcsname{5}
\expandafter\def\csname val@many.adaptive_gate.damages.all\endcsname{5}
\expandafter\def\csname val@many.adaptive_gate.identical_final_request.all\endcsname{245}
\expandafter\def\csname val@many.adaptive_gate.repairs.all\endcsname{5}
\expandafter\def\csname val@many.question_ledger.changes_on_identical_request.all\endcsname{3}
\expandafter\def\csname val@many.question_ledger.damages.all\endcsname{16}
\expandafter\def\csname val@many.question_ledger.identical_final_request.all\endcsname{57}
\expandafter\def\csname val@many.question_ledger.repairs.all\endcsname{18}
\expandafter\def\csname val@many.tool_agent.changes_on_identical_request.all\endcsname{0}
\expandafter\def\csname val@many.tool_agent.damages.all\endcsname{21}
\expandafter\def\csname val@many.tool_agent.identical_final_request.all\endcsname{0}
\expandafter\def\csname val@many.tool_agent.repairs.all\endcsname{26}
\expandafter\def\csname val@one.adaptive_gate.accuracy.a\endcsname{37.0}
\expandafter\def\csname val@one.adaptive_gate.accuracy.f\endcsname{63.0}
\expandafter\def\csname val@one.adaptive_gate.accuracy.ft\endcsname{61.3}
\expandafter\def\csname val@one.adaptive_gate.changes_on_identical_request.a\endcsname{0}
\expandafter\def\csname val@one.adaptive_gate.changes_on_identical_request.all\endcsname{1}
\expandafter\def\csname val@one.adaptive_gate.changes_on_identical_request.f\endcsname{1}
\expandafter\def\csname val@one.adaptive_gate.changes_on_identical_request.ft\endcsname{0}
\expandafter\def\csname val@one.adaptive_gate.damages.a\endcsname{0}
\expandafter\def\csname val@one.adaptive_gate.damages.all\endcsname{1}
\expandafter\def\csname val@one.adaptive_gate.damages.f\endcsname{1}
\expandafter\def\csname val@one.adaptive_gate.damages.ft\endcsname{0}
\expandafter\def\csname val@one.adaptive_gate.difference.a\endcsname{$+1.0$}
\expandafter\def\csname val@one.adaptive_gate.difference.f\endcsname{$+1.0$}
\expandafter\def\csname val@one.adaptive_gate.difference.ft\endcsname{$+4.3$}
\expandafter\def\csname val@one.adaptive_gate.failures.a\endcsname{1}
\expandafter\def\csname val@one.adaptive_gate.failures.f\endcsname{2}
\expandafter\def\csname val@one.adaptive_gate.failures.ft\endcsname{4}
\expandafter\def\csname val@one.adaptive_gate.high.a\endcsname{$+3.0$}
\expandafter\def\csname val@one.adaptive_gate.high.f\endcsname{$+4.7$}
\expandafter\def\csname val@one.adaptive_gate.high.ft\endcsname{$+8.6$}
\expandafter\def\csname val@one.adaptive_gate.identical_final_request.a\endcsname{75}
\expandafter\def\csname val@one.adaptive_gate.identical_final_request.all\endcsname{248}
\expandafter\def\csname val@one.adaptive_gate.identical_final_request.f\endcsname{92}
\expandafter\def\csname val@one.adaptive_gate.identical_final_request.ft\endcsname{81}
\expandafter\def\csname val@one.adaptive_gate.low.a\endcsname{$+0.0$}
\expandafter\def\csname val@one.adaptive_gate.low.f\endcsname{$-2.3$}
\expandafter\def\csname val@one.adaptive_gate.low.ft\endcsname{$+1.0$}
\expandafter\def\csname val@one.adaptive_gate.mcnemar_p.a\endcsname{1.000}
\expandafter\def\csname val@one.adaptive_gate.mcnemar_p.all\endcsname{0.070}
\expandafter\def\csname val@one.adaptive_gate.mcnemar_p.f\endcsname{1.000}
\expandafter\def\csname val@one.adaptive_gate.mcnemar_p.ft\endcsname{0.125}
\expandafter\def\csname val@one.adaptive_gate.repairs.a\endcsname{1}
\expandafter\def\csname val@one.adaptive_gate.repairs.all\endcsname{7}
\expandafter\def\csname val@one.adaptive_gate.repairs.f\endcsname{2}
\expandafter\def\csname val@one.adaptive_gate.repairs.ft\endcsname{4}
\expandafter\def\csname val@one.question_ledger.accuracy.a\endcsname{40.0}
\expandafter\def\csname val@one.question_ledger.accuracy.f\endcsname{62.0}
\expandafter\def\csname val@one.question_ledger.accuracy.ft\endcsname{54.8}
\expandafter\def\csname val@one.question_ledger.changes_on_identical_request.a\endcsname{0}
\expandafter\def\csname val@one.question_ledger.changes_on_identical_request.all\endcsname{0}
\expandafter\def\csname val@one.question_ledger.changes_on_identical_request.f\endcsname{0}
\expandafter\def\csname val@one.question_ledger.changes_on_identical_request.ft\endcsname{0}
\expandafter\def\csname val@one.question_ledger.damages.a\endcsname{6}
\expandafter\def\csname val@one.question_ledger.damages.all\endcsname{19}
\expandafter\def\csname val@one.question_ledger.damages.f\endcsname{3}
\expandafter\def\csname val@one.question_ledger.damages.ft\endcsname{10}
\expandafter\def\csname val@one.question_ledger.difference.a\endcsname{$+4.0$}
\expandafter\def\csname val@one.question_ledger.difference.f\endcsname{$+0.0$}
\expandafter\def\csname val@one.question_ledger.difference.ft\endcsname{$-2.2$}
\expandafter\def\csname val@one.question_ledger.failures.a\endcsname{5}
\expandafter\def\csname val@one.question_ledger.failures.f\endcsname{3}
\expandafter\def\csname val@one.question_ledger.failures.ft\endcsname{6}
\expandafter\def\csname val@one.question_ledger.high.a\endcsname{$+12.0$}
\expandafter\def\csname val@one.question_ledger.high.f\endcsname{$+5.0$}
\expandafter\def\csname val@one.question_ledger.high.ft\endcsname{$+6.9$}
\expandafter\def\csname val@one.question_ledger.identical_final_request.a\endcsname{12}
\expandafter\def\csname val@one.question_ledger.identical_final_request.all\endcsname{54}
\expandafter\def\csname val@one.question_ledger.identical_final_request.f\endcsname{24}
\expandafter\def\csname val@one.question_ledger.identical_final_request.ft\endcsname{18}
\expandafter\def\csname val@one.question_ledger.low.a\endcsname{$-4.0$}
\expandafter\def\csname val@one.question_ledger.low.f\endcsname{$-4.7$}
\expandafter\def\csname val@one.question_ledger.low.ft\endcsname{$-11.1$}
\expandafter\def\csname val@one.question_ledger.mcnemar_p.a\endcsname{0.454}
\expandafter\def\csname val@one.question_ledger.mcnemar_p.all\endcsname{0.875}
\expandafter\def\csname val@one.question_ledger.mcnemar_p.f\endcsname{1.000}
\expandafter\def\csname val@one.question_ledger.mcnemar_p.ft\endcsname{0.815}
\expandafter\def\csname val@one.question_ledger.repairs.a\endcsname{10}
\expandafter\def\csname val@one.question_ledger.repairs.all\endcsname{21}
\expandafter\def\csname val@one.question_ledger.repairs.f\endcsname{3}
\expandafter\def\csname val@one.question_ledger.repairs.ft\endcsname{8}
\expandafter\def\csname val@one.single.dense.accuracy.a\endcsname{35.0}
\expandafter\def\csname val@one.single.dense.accuracy.f\endcsname{65.0}
\expandafter\def\csname val@one.single.dense.accuracy.ft\endcsname{63.4}
\expandafter\def\csname val@one.single.dense.failures.a\endcsname{2}
\expandafter\def\csname val@one.single.dense.failures.f\endcsname{2}
\expandafter\def\csname val@one.single.dense.failures.ft\endcsname{1}
\expandafter\def\csname val@one.single.hybrid.accuracy.a\endcsname{36.0}
\expandafter\def\csname val@one.single.hybrid.accuracy.f\endcsname{62.0}
\expandafter\def\csname val@one.single.hybrid.accuracy.ft\endcsname{57.0}
\expandafter\def\csname val@one.single.hybrid.failures.a\endcsname{0}
\expandafter\def\csname val@one.single.hybrid.failures.f\endcsname{1}
\expandafter\def\csname val@one.single.hybrid.failures.ft\endcsname{4}
\expandafter\def\csname val@one.single.sparse.accuracy.a\endcsname{38.0}
\expandafter\def\csname val@one.single.sparse.accuracy.f\endcsname{66.0}
\expandafter\def\csname val@one.single.sparse.accuracy.ft\endcsname{55.9}
\expandafter\def\csname val@one.single.sparse.failures.a\endcsname{2}
\expandafter\def\csname val@one.single.sparse.failures.f\endcsname{0}
\expandafter\def\csname val@one.single.sparse.failures.ft\endcsname{4}
\expandafter\def\csname val@one.tool_agent.accuracy.a\endcsname{43.0}
\expandafter\def\csname val@one.tool_agent.accuracy.f\endcsname{68.0}
\expandafter\def\csname val@one.tool_agent.accuracy.ft\endcsname{53.8}
\expandafter\def\csname val@one.tool_agent.changes_on_identical_request.a\endcsname{0}
\expandafter\def\csname val@one.tool_agent.changes_on_identical_request.all\endcsname{0}
\expandafter\def\csname val@one.tool_agent.changes_on_identical_request.f\endcsname{0}
\expandafter\def\csname val@one.tool_agent.changes_on_identical_request.ft\endcsname{0}
\expandafter\def\csname val@one.tool_agent.damages.a\endcsname{10}
\expandafter\def\csname val@one.tool_agent.damages.all\endcsname{24}
\expandafter\def\csname val@one.tool_agent.damages.f\endcsname{5}
\expandafter\def\csname val@one.tool_agent.damages.ft\endcsname{9}
\expandafter\def\csname val@one.tool_agent.difference.a\endcsname{$+7.0$}
\expandafter\def\csname val@one.tool_agent.difference.f\endcsname{$+6.0$}
\expandafter\def\csname val@one.tool_agent.difference.ft\endcsname{$-3.2$}
\expandafter\def\csname val@one.tool_agent.failures.a\endcsname{1}
\expandafter\def\csname val@one.tool_agent.failures.f\endcsname{1}
\expandafter\def\csname val@one.tool_agent.failures.ft\endcsname{6}
\expandafter\def\csname val@one.tool_agent.high.a\endcsname{$+17.0$}
\expandafter\def\csname val@one.tool_agent.high.f\endcsname{$+14.3$}
\expandafter\def\csname val@one.tool_agent.high.ft\endcsname{$+5.6$}
\expandafter\def\csname val@one.tool_agent.identical_final_request.a\endcsname{0}
\expandafter\def\csname val@one.tool_agent.identical_final_request.all\endcsname{0}
\expandafter\def\csname val@one.tool_agent.identical_final_request.f\endcsname{0}
\expandafter\def\csname val@one.tool_agent.identical_final_request.ft\endcsname{0}
\expandafter\def\csname val@one.tool_agent.low.a\endcsname{$-3.0$}
\expandafter\def\csname val@one.tool_agent.low.f\endcsname{$-2.1$}
\expandafter\def\csname val@one.tool_agent.low.ft\endcsname{$-12.0$}
\expandafter\def\csname val@one.tool_agent.mcnemar_p.a\endcsname{0.248}
\expandafter\def\csname val@one.tool_agent.mcnemar_p.all\endcsname{0.237}
\expandafter\def\csname val@one.tool_agent.mcnemar_p.f\endcsname{0.210}
\expandafter\def\csname val@one.tool_agent.mcnemar_p.ft\endcsname{0.607}
\expandafter\def\csname val@one.tool_agent.repairs.a\endcsname{17}
\expandafter\def\csname val@one.tool_agent.repairs.all\endcsname{34}
\expandafter\def\csname val@one.tool_agent.repairs.f\endcsname{11}
\expandafter\def\csname val@one.tool_agent.repairs.ft\endcsname{6}
\expandafter\def\csname val@order.pair.changed_correctness.a\endcsname{0}
\expandafter\def\csname val@order.pair.changed_correctness.all\endcsname{0}
\expandafter\def\csname val@order.pair.changed_correctness.f\endcsname{0}
\expandafter\def\csname val@order.pair.changed_correctness.ft\endcsname{0}
\expandafter\def\csname val@order.pair.claims.a\endcsname{100}
\expandafter\def\csname val@order.pair.claims.all\endcsname{293}
\expandafter\def\csname val@order.pair.claims.f\endcsname{100}
\expandafter\def\csname val@order.pair.claims.ft\endcsname{93}
\expandafter\def\csname val@order.pair.correct_to_wrong.a\endcsname{0}
\expandafter\def\csname val@order.pair.correct_to_wrong.all\endcsname{0}
\expandafter\def\csname val@order.pair.correct_to_wrong.f\endcsname{0}
\expandafter\def\csname val@order.pair.correct_to_wrong.ft\endcsname{0}
\expandafter\def\csname val@order.pair.different_request.a\endcsname{0}
\expandafter\def\csname val@order.pair.different_request.all\endcsname{0}
\expandafter\def\csname val@order.pair.different_request.f\endcsname{0}
\expandafter\def\csname val@order.pair.different_request.ft\endcsname{0}
\expandafter\def\csname val@order.pair.first_accuracy.a\endcsname{35.0}
\expandafter\def\csname val@order.pair.first_accuracy.all\endcsname{51.2}
\expandafter\def\csname val@order.pair.first_accuracy.f\endcsname{64.0}
\expandafter\def\csname val@order.pair.first_accuracy.ft\endcsname{54.8}
\expandafter\def\csname val@order.pair.first_correct.a\endcsname{35}
\expandafter\def\csname val@order.pair.first_correct.all\endcsname{150}
\expandafter\def\csname val@order.pair.first_correct.f\endcsname{64}
\expandafter\def\csname val@order.pair.first_correct.ft\endcsname{51}
\expandafter\def\csname val@order.pair.first_failures.a\endcsname{2}
\expandafter\def\csname val@order.pair.first_failures.all\endcsname{9}
\expandafter\def\csname val@order.pair.first_failures.f\endcsname{1}
\expandafter\def\csname val@order.pair.first_failures.ft\endcsname{6}
\expandafter\def\csname val@order.pair.identical_final_request.a\endcsname{100}
\expandafter\def\csname val@order.pair.identical_final_request.all\endcsname{293}
\expandafter\def\csname val@order.pair.identical_final_request.f\endcsname{100}
\expandafter\def\csname val@order.pair.identical_final_request.ft\endcsname{93}
\expandafter\def\csname val@order.pair.mcnemar_p.a\endcsname{1.000}
\expandafter\def\csname val@order.pair.mcnemar_p.all\endcsname{1.000}
\expandafter\def\csname val@order.pair.mcnemar_p.f\endcsname{1.000}
\expandafter\def\csname val@order.pair.mcnemar_p.ft\endcsname{1.000}
\expandafter\def\csname val@order.pair.second_accuracy.a\endcsname{35.0}
\expandafter\def\csname val@order.pair.second_accuracy.all\endcsname{51.2}
\expandafter\def\csname val@order.pair.second_accuracy.f\endcsname{64.0}
\expandafter\def\csname val@order.pair.second_accuracy.ft\endcsname{54.8}
\expandafter\def\csname val@order.pair.second_correct.a\endcsname{35}
\expandafter\def\csname val@order.pair.second_correct.all\endcsname{150}
\expandafter\def\csname val@order.pair.second_correct.f\endcsname{64}
\expandafter\def\csname val@order.pair.second_correct.ft\endcsname{51}
\expandafter\def\csname val@order.pair.second_failures.a\endcsname{2}
\expandafter\def\csname val@order.pair.second_failures.all\endcsname{9}
\expandafter\def\csname val@order.pair.second_failures.f\endcsname{1}
\expandafter\def\csname val@order.pair.second_failures.ft\endcsname{6}
\expandafter\def\csname val@order.pair.verdict_changed.a\endcsname{0}
\expandafter\def\csname val@order.pair.verdict_changed.all\endcsname{0}
\expandafter\def\csname val@order.pair.verdict_changed.f\endcsname{0}
\expandafter\def\csname val@order.pair.verdict_changed.ft\endcsname{0}
\expandafter\def\csname val@order.pair.verdict_changed_on_identical_request.a\endcsname{0}
\expandafter\def\csname val@order.pair.verdict_changed_on_identical_request.all\endcsname{0}
\expandafter\def\csname val@order.pair.verdict_changed_on_identical_request.f\endcsname{0}
\expandafter\def\csname val@order.pair.verdict_changed_on_identical_request.ft\endcsname{0}
\expandafter\def\csname val@order.pair.wrong_to_correct.a\endcsname{0}
\expandafter\def\csname val@order.pair.wrong_to_correct.all\endcsname{0}
\expandafter\def\csname val@order.pair.wrong_to_correct.f\endcsname{0}
\expandafter\def\csname val@order.pair.wrong_to_correct.ft\endcsname{0}
\expandafter\def\csname val@order.reversed.changed_correctness.a\endcsname{9}
\expandafter\def\csname val@order.reversed.changed_correctness.all\endcsname{20}
\expandafter\def\csname val@order.reversed.changed_correctness.f\endcsname{4}
\expandafter\def\csname val@order.reversed.changed_correctness.ft\endcsname{7}
\expandafter\def\csname val@order.reversed.claims.a\endcsname{100}
\expandafter\def\csname val@order.reversed.claims.all\endcsname{293}
\expandafter\def\csname val@order.reversed.claims.f\endcsname{100}
\expandafter\def\csname val@order.reversed.claims.ft\endcsname{93}
\expandafter\def\csname val@order.reversed.correct_to_wrong.a\endcsname{6}
\expandafter\def\csname val@order.reversed.correct_to_wrong.all\endcsname{11}
\expandafter\def\csname val@order.reversed.correct_to_wrong.f\endcsname{3}
\expandafter\def\csname val@order.reversed.correct_to_wrong.ft\endcsname{2}
\expandafter\def\csname val@order.reversed.different_request.a\endcsname{1}
\expandafter\def\csname val@order.reversed.different_request.all\endcsname{1}
\expandafter\def\csname val@order.reversed.different_request.f\endcsname{0}
\expandafter\def\csname val@order.reversed.different_request.ft\endcsname{0}
\expandafter\def\csname val@order.reversed.first_accuracy.a\endcsname{38.0}
\expandafter\def\csname val@order.reversed.first_accuracy.all\endcsname{51.9}
\expandafter\def\csname val@order.reversed.first_accuracy.f\endcsname{66.0}
\expandafter\def\csname val@order.reversed.first_accuracy.ft\endcsname{51.6}
\expandafter\def\csname val@order.reversed.first_correct.a\endcsname{38}
\expandafter\def\csname val@order.reversed.first_correct.all\endcsname{152}
\expandafter\def\csname val@order.reversed.first_correct.f\endcsname{66}
\expandafter\def\csname val@order.reversed.first_correct.ft\endcsname{48}
\expandafter\def\csname val@order.reversed.first_failures.a\endcsname{3}
\expandafter\def\csname val@order.reversed.first_failures.all\endcsname{8}
\expandafter\def\csname val@order.reversed.first_failures.f\endcsname{0}
\expandafter\def\csname val@order.reversed.first_failures.ft\endcsname{5}
\expandafter\def\csname val@order.reversed.identical_final_request.a\endcsname{99}
\expandafter\def\csname val@order.reversed.identical_final_request.all\endcsname{292}
\expandafter\def\csname val@order.reversed.identical_final_request.f\endcsname{100}
\expandafter\def\csname val@order.reversed.identical_final_request.ft\endcsname{93}
\expandafter\def\csname val@order.reversed.mcnemar_p.a\endcsname{0.508}
\expandafter\def\csname val@order.reversed.mcnemar_p.all\endcsname{0.824}
\expandafter\def\csname val@order.reversed.mcnemar_p.f\endcsname{0.625}
\expandafter\def\csname val@order.reversed.mcnemar_p.ft\endcsname{0.453}
\expandafter\def\csname val@order.reversed.second_accuracy.a\endcsname{35.0}
\expandafter\def\csname val@order.reversed.second_accuracy.all\endcsname{51.2}
\expandafter\def\csname val@order.reversed.second_accuracy.f\endcsname{64.0}
\expandafter\def\csname val@order.reversed.second_accuracy.ft\endcsname{54.8}
\expandafter\def\csname val@order.reversed.second_correct.a\endcsname{35}
\expandafter\def\csname val@order.reversed.second_correct.all\endcsname{150}
\expandafter\def\csname val@order.reversed.second_correct.f\endcsname{64}
\expandafter\def\csname val@order.reversed.second_correct.ft\endcsname{51}
\expandafter\def\csname val@order.reversed.second_failures.a\endcsname{2}
\expandafter\def\csname val@order.reversed.second_failures.all\endcsname{9}
\expandafter\def\csname val@order.reversed.second_failures.f\endcsname{1}
\expandafter\def\csname val@order.reversed.second_failures.ft\endcsname{6}
\expandafter\def\csname val@order.reversed.verdict_changed.a\endcsname{16}
\expandafter\def\csname val@order.reversed.verdict_changed.all\endcsname{32}
\expandafter\def\csname val@order.reversed.verdict_changed.f\endcsname{5}
\expandafter\def\csname val@order.reversed.verdict_changed.ft\endcsname{11}
\expandafter\def\csname val@order.reversed.verdict_changed_on_identical_request.a\endcsname{16}
\expandafter\def\csname val@order.reversed.verdict_changed_on_identical_request.all\endcsname{32}
\expandafter\def\csname val@order.reversed.verdict_changed_on_identical_request.f\endcsname{5}
\expandafter\def\csname val@order.reversed.verdict_changed_on_identical_request.ft\endcsname{11}
\expandafter\def\csname val@order.reversed.wrong_to_correct.a\endcsname{3}
\expandafter\def\csname val@order.reversed.wrong_to_correct.all\endcsname{9}
\expandafter\def\csname val@order.reversed.wrong_to_correct.f\endcsname{1}
\expandafter\def\csname val@order.reversed.wrong_to_correct.ft\endcsname{5}
\expandafter\def\csname val@order.same.changed_correctness.a\endcsname{9}
\expandafter\def\csname val@order.same.changed_correctness.all\endcsname{20}
\expandafter\def\csname val@order.same.changed_correctness.f\endcsname{4}
\expandafter\def\csname val@order.same.changed_correctness.ft\endcsname{7}
\expandafter\def\csname val@order.same.claims.a\endcsname{100}
\expandafter\def\csname val@order.same.claims.all\endcsname{293}
\expandafter\def\csname val@order.same.claims.f\endcsname{100}
\expandafter\def\csname val@order.same.claims.ft\endcsname{93}
\expandafter\def\csname val@order.same.correct_to_wrong.a\endcsname{6}
\expandafter\def\csname val@order.same.correct_to_wrong.all\endcsname{11}
\expandafter\def\csname val@order.same.correct_to_wrong.f\endcsname{3}
\expandafter\def\csname val@order.same.correct_to_wrong.ft\endcsname{2}
\expandafter\def\csname val@order.same.different_request.a\endcsname{1}
\expandafter\def\csname val@order.same.different_request.all\endcsname{1}
\expandafter\def\csname val@order.same.different_request.f\endcsname{0}
\expandafter\def\csname val@order.same.different_request.ft\endcsname{0}
\expandafter\def\csname val@order.same.first_accuracy.a\endcsname{38.0}
\expandafter\def\csname val@order.same.first_accuracy.all\endcsname{51.9}
\expandafter\def\csname val@order.same.first_accuracy.f\endcsname{66.0}
\expandafter\def\csname val@order.same.first_accuracy.ft\endcsname{51.6}
\expandafter\def\csname val@order.same.first_correct.a\endcsname{38}
\expandafter\def\csname val@order.same.first_correct.all\endcsname{152}
\expandafter\def\csname val@order.same.first_correct.f\endcsname{66}
\expandafter\def\csname val@order.same.first_correct.ft\endcsname{48}
\expandafter\def\csname val@order.same.first_failures.a\endcsname{3}
\expandafter\def\csname val@order.same.first_failures.all\endcsname{8}
\expandafter\def\csname val@order.same.first_failures.f\endcsname{0}
\expandafter\def\csname val@order.same.first_failures.ft\endcsname{5}
\expandafter\def\csname val@order.same.identical_final_request.a\endcsname{99}
\expandafter\def\csname val@order.same.identical_final_request.all\endcsname{292}
\expandafter\def\csname val@order.same.identical_final_request.f\endcsname{100}
\expandafter\def\csname val@order.same.identical_final_request.ft\endcsname{93}
\expandafter\def\csname val@order.same.mcnemar_p.a\endcsname{0.508}
\expandafter\def\csname val@order.same.mcnemar_p.all\endcsname{0.824}
\expandafter\def\csname val@order.same.mcnemar_p.f\endcsname{0.625}
\expandafter\def\csname val@order.same.mcnemar_p.ft\endcsname{0.453}
\expandafter\def\csname val@order.same.second_accuracy.a\endcsname{35.0}
\expandafter\def\csname val@order.same.second_accuracy.all\endcsname{51.2}
\expandafter\def\csname val@order.same.second_accuracy.f\endcsname{64.0}
\expandafter\def\csname val@order.same.second_accuracy.ft\endcsname{54.8}
\expandafter\def\csname val@order.same.second_correct.a\endcsname{35}
\expandafter\def\csname val@order.same.second_correct.all\endcsname{150}
\expandafter\def\csname val@order.same.second_correct.f\endcsname{64}
\expandafter\def\csname val@order.same.second_correct.ft\endcsname{51}
\expandafter\def\csname val@order.same.second_failures.a\endcsname{2}
\expandafter\def\csname val@order.same.second_failures.all\endcsname{9}
\expandafter\def\csname val@order.same.second_failures.f\endcsname{1}
\expandafter\def\csname val@order.same.second_failures.ft\endcsname{6}
\expandafter\def\csname val@order.same.verdict_changed.a\endcsname{16}
\expandafter\def\csname val@order.same.verdict_changed.all\endcsname{32}
\expandafter\def\csname val@order.same.verdict_changed.f\endcsname{5}
\expandafter\def\csname val@order.same.verdict_changed.ft\endcsname{11}
\expandafter\def\csname val@order.same.verdict_changed_on_identical_request.a\endcsname{16}
\expandafter\def\csname val@order.same.verdict_changed_on_identical_request.all\endcsname{32}
\expandafter\def\csname val@order.same.verdict_changed_on_identical_request.f\endcsname{5}
\expandafter\def\csname val@order.same.verdict_changed_on_identical_request.ft\endcsname{11}
\expandafter\def\csname val@order.same.wrong_to_correct.a\endcsname{3}
\expandafter\def\csname val@order.same.wrong_to_correct.all\endcsname{9}
\expandafter\def\csname val@order.same.wrong_to_correct.f\endcsname{1}
\expandafter\def\csname val@order.same.wrong_to_correct.ft\endcsname{5}
\expandafter\def\csname val@repeat.full.changed_correctness.a\endcsname{8}
\expandafter\def\csname val@repeat.full.changed_correctness.all\endcsname{21}
\expandafter\def\csname val@repeat.full.changed_correctness.f\endcsname{4}
\expandafter\def\csname val@repeat.full.changed_correctness.ft\endcsname{9}
\expandafter\def\csname val@repeat.full.claims.a\endcsname{100}
\expandafter\def\csname val@repeat.full.claims.all\endcsname{293}
\expandafter\def\csname val@repeat.full.claims.f\endcsname{100}
\expandafter\def\csname val@repeat.full.claims.ft\endcsname{93}
\expandafter\def\csname val@repeat.full.correct_to_wrong.a\endcsname{3}
\expandafter\def\csname val@repeat.full.correct_to_wrong.all\endcsname{10}
\expandafter\def\csname val@repeat.full.correct_to_wrong.f\endcsname{0}
\expandafter\def\csname val@repeat.full.correct_to_wrong.ft\endcsname{7}
\expandafter\def\csname val@repeat.full.different_request.a\endcsname{1}
\expandafter\def\csname val@repeat.full.different_request.all\endcsname{1}
\expandafter\def\csname val@repeat.full.different_request.f\endcsname{0}
\expandafter\def\csname val@repeat.full.different_request.ft\endcsname{0}
\expandafter\def\csname val@repeat.full.first_accuracy.a\endcsname{36.0}
\expandafter\def\csname val@repeat.full.first_accuracy.all\endcsname{51.5}
\expandafter\def\csname val@repeat.full.first_accuracy.f\endcsname{62.0}
\expandafter\def\csname val@repeat.full.first_accuracy.ft\endcsname{57.0}
\expandafter\def\csname val@repeat.full.first_correct.a\endcsname{36}
\expandafter\def\csname val@repeat.full.first_correct.all\endcsname{151}
\expandafter\def\csname val@repeat.full.first_correct.f\endcsname{62}
\expandafter\def\csname val@repeat.full.first_correct.ft\endcsname{53}
\expandafter\def\csname val@repeat.full.first_failures.a\endcsname{0}
\expandafter\def\csname val@repeat.full.first_failures.all\endcsname{5}
\expandafter\def\csname val@repeat.full.first_failures.f\endcsname{1}
\expandafter\def\csname val@repeat.full.first_failures.ft\endcsname{4}
\expandafter\def\csname val@repeat.full.identical_final_request.a\endcsname{99}
\expandafter\def\csname val@repeat.full.identical_final_request.all\endcsname{292}
\expandafter\def\csname val@repeat.full.identical_final_request.f\endcsname{100}
\expandafter\def\csname val@repeat.full.identical_final_request.ft\endcsname{93}
\expandafter\def\csname val@repeat.full.mcnemar_p.a\endcsname{0.727}
\expandafter\def\csname val@repeat.full.mcnemar_p.all\endcsname{1.000}
\expandafter\def\csname val@repeat.full.mcnemar_p.f\endcsname{0.125}
\expandafter\def\csname val@repeat.full.mcnemar_p.ft\endcsname{0.180}
\expandafter\def\csname val@repeat.full.second_accuracy.a\endcsname{38.0}
\expandafter\def\csname val@repeat.full.second_accuracy.all\endcsname{51.9}
\expandafter\def\csname val@repeat.full.second_accuracy.f\endcsname{66.0}
\expandafter\def\csname val@repeat.full.second_accuracy.ft\endcsname{51.6}
\expandafter\def\csname val@repeat.full.second_correct.a\endcsname{38}
\expandafter\def\csname val@repeat.full.second_correct.all\endcsname{152}
\expandafter\def\csname val@repeat.full.second_correct.f\endcsname{66}
\expandafter\def\csname val@repeat.full.second_correct.ft\endcsname{48}
\expandafter\def\csname val@repeat.full.second_failures.a\endcsname{3}
\expandafter\def\csname val@repeat.full.second_failures.all\endcsname{8}
\expandafter\def\csname val@repeat.full.second_failures.f\endcsname{0}
\expandafter\def\csname val@repeat.full.second_failures.ft\endcsname{5}
\expandafter\def\csname val@repeat.full.verdict_changed.a\endcsname{16}
\expandafter\def\csname val@repeat.full.verdict_changed.all\endcsname{32}
\expandafter\def\csname val@repeat.full.verdict_changed.f\endcsname{5}
\expandafter\def\csname val@repeat.full.verdict_changed.ft\endcsname{11}
\expandafter\def\csname val@repeat.full.verdict_changed_on_identical_request.a\endcsname{16}
\expandafter\def\csname val@repeat.full.verdict_changed_on_identical_request.all\endcsname{32}
\expandafter\def\csname val@repeat.full.verdict_changed_on_identical_request.f\endcsname{5}
\expandafter\def\csname val@repeat.full.verdict_changed_on_identical_request.ft\endcsname{11}
\expandafter\def\csname val@repeat.full.wrong_to_correct.a\endcsname{5}
\expandafter\def\csname val@repeat.full.wrong_to_correct.all\endcsname{11}
\expandafter\def\csname val@repeat.full.wrong_to_correct.f\endcsname{4}
\expandafter\def\csname val@repeat.full.wrong_to_correct.ft\endcsname{2}
\expandafter\def\csname val@repeat.selected.changed_correctness.a\endcsname{3}
\expandafter\def\csname val@repeat.selected.changed_correctness.all\endcsname{11}
\expandafter\def\csname val@repeat.selected.changed_correctness.f\endcsname{5}
\expandafter\def\csname val@repeat.selected.changed_correctness.ft\endcsname{3}
\expandafter\def\csname val@repeat.selected.claims.a\endcsname{100}
\expandafter\def\csname val@repeat.selected.claims.all\endcsname{293}
\expandafter\def\csname val@repeat.selected.claims.f\endcsname{100}
\expandafter\def\csname val@repeat.selected.claims.ft\endcsname{93}
\expandafter\def\csname val@repeat.selected.correct_to_wrong.a\endcsname{2}
\expandafter\def\csname val@repeat.selected.correct_to_wrong.all\endcsname{7}
\expandafter\def\csname val@repeat.selected.correct_to_wrong.f\endcsname{3}
\expandafter\def\csname val@repeat.selected.correct_to_wrong.ft\endcsname{2}
\expandafter\def\csname val@repeat.selected.different_request.a\endcsname{0}
\expandafter\def\csname val@repeat.selected.different_request.all\endcsname{0}
\expandafter\def\csname val@repeat.selected.different_request.f\endcsname{0}
\expandafter\def\csname val@repeat.selected.different_request.ft\endcsname{0}
\expandafter\def\csname val@repeat.selected.first_accuracy.a\endcsname{65.0}
\expandafter\def\csname val@repeat.selected.first_accuracy.all\endcsname{73.4}
\expandafter\def\csname val@repeat.selected.first_accuracy.f\endcsname{81.0}
\expandafter\def\csname val@repeat.selected.first_accuracy.ft\endcsname{74.2}
\expandafter\def\csname val@repeat.selected.first_correct.a\endcsname{65}
\expandafter\def\csname val@repeat.selected.first_correct.all\endcsname{215}
\expandafter\def\csname val@repeat.selected.first_correct.f\endcsname{81}
\expandafter\def\csname val@repeat.selected.first_correct.ft\endcsname{69}
\expandafter\def\csname val@repeat.selected.first_failures.a\endcsname{1}
\expandafter\def\csname val@repeat.selected.first_failures.all\endcsname{5}
\expandafter\def\csname val@repeat.selected.first_failures.f\endcsname{1}
\expandafter\def\csname val@repeat.selected.first_failures.ft\endcsname{3}
\expandafter\def\csname val@repeat.selected.identical_final_request.a\endcsname{100}
\expandafter\def\csname val@repeat.selected.identical_final_request.all\endcsname{293}
\expandafter\def\csname val@repeat.selected.identical_final_request.f\endcsname{100}
\expandafter\def\csname val@repeat.selected.identical_final_request.ft\endcsname{93}
\expandafter\def\csname val@repeat.selected.mcnemar_p.a\endcsname{1.000}
\expandafter\def\csname val@repeat.selected.mcnemar_p.all\endcsname{0.549}
\expandafter\def\csname val@repeat.selected.mcnemar_p.f\endcsname{1.000}
\expandafter\def\csname val@repeat.selected.mcnemar_p.ft\endcsname{1.000}
\expandafter\def\csname val@repeat.selected.second_accuracy.a\endcsname{64.0}
\expandafter\def\csname val@repeat.selected.second_accuracy.all\endcsname{72.4}
\expandafter\def\csname val@repeat.selected.second_accuracy.f\endcsname{80.0}
\expandafter\def\csname val@repeat.selected.second_accuracy.ft\endcsname{73.1}
\expandafter\def\csname val@repeat.selected.second_correct.a\endcsname{64}
\expandafter\def\csname val@repeat.selected.second_correct.all\endcsname{212}
\expandafter\def\csname val@repeat.selected.second_correct.f\endcsname{80}
\expandafter\def\csname val@repeat.selected.second_correct.ft\endcsname{68}
\expandafter\def\csname val@repeat.selected.second_failures.a\endcsname{3}
\expandafter\def\csname val@repeat.selected.second_failures.all\endcsname{8}
\expandafter\def\csname val@repeat.selected.second_failures.f\endcsname{3}
\expandafter\def\csname val@repeat.selected.second_failures.ft\endcsname{2}
\expandafter\def\csname val@repeat.selected.verdict_changed.a\endcsname{6}
\expandafter\def\csname val@repeat.selected.verdict_changed.all\endcsname{20}
\expandafter\def\csname val@repeat.selected.verdict_changed.f\endcsname{6}
\expandafter\def\csname val@repeat.selected.verdict_changed.ft\endcsname{8}
\expandafter\def\csname val@repeat.selected.verdict_changed_on_identical_request.a\endcsname{6}
\expandafter\def\csname val@repeat.selected.verdict_changed_on_identical_request.all\endcsname{20}
\expandafter\def\csname val@repeat.selected.verdict_changed_on_identical_request.f\endcsname{6}
\expandafter\def\csname val@repeat.selected.verdict_changed_on_identical_request.ft\endcsname{8}
\expandafter\def\csname val@repeat.selected.wrong_to_correct.a\endcsname{1}
\expandafter\def\csname val@repeat.selected.wrong_to_correct.all\endcsname{4}
\expandafter\def\csname val@repeat.selected.wrong_to_correct.f\endcsname{2}
\expandafter\def\csname val@repeat.selected.wrong_to_correct.ft\endcsname{1}

%% file: sections/abstract.tex
\begin{abstract}
An evidence-grounded fact-checker can fail because it does not find the
evidence or because it does not use it; end-to-end
accuracy does not tell the two apart. We present a protocol that follows the
reference evidence from the store to the verdict and separates retrieval,
search control, judgement and verdict validation. We assess 293 claims
from FEVER, FEVER 2.0 and AVeriTeC with one open-weights 9B model under two
evidence conditions: small stores of annotated evidence, where
discovery is easy, and the full source collections. In the control
condition a single hybrid retrieval pass delivers complete reference evidence
for 226 of 228 evidence-bearing claims, and adaptive controllers and a tool
agent spend two to nine times as many model calls without a consistent gain. Of 34 decisive claims
labelled as lacking evidence, 33 have the complete evidence in the packet: in
15 the judge abstains, and in 18 validation rules overturn a
decisive verdict. On AVeriTeC these rules lower single-pass accuracy by six
points, to 65.0\% (balanced accuracy 57.6\%). Ten judges from eight model
families also show abstentions and validation conversions on fixed packets.
Against the full stores hybrid single-pass accuracy falls by 17 to 29 points.
Wikipedia evidence recovery declines; AVeriTeC recovery changes from annotated
answers to cited pages and is measured separately. No controller recovers the
accuracy loss. The agent also shows no resolved gain on 714 further Wikipedia
claims. Evidence recovery, judgement and
validation should be measured separately before an error is attributed to
search.
\end{abstract}

%% file: body.tex
\input{sections/introduction}
\input{sections/related}
\input{sections/method}
\input{sections/design}
\input{sections/results}
\input{sections/discussion}

%% file: sections/introduction.tex
\section{Introduction}\label{sec:intro}
An evidence-grounded fact-checker has two jobs. It must find sources that bear
on a claim, and it must decide what those sources establish. Much recent work
improves the first job: retrieval is made hybrid, iterative, graph-assisted or
agentic, on the expectation that a better investigation yields a better
verdict. That expectation holds only when the investigation is what limits the
verdict. If the decisive passage already lies in front of the judge and the
judge misreads it, a second search adds cost and corrects nothing. If the
passage was never retrieved, no judge can recover it. A system builder who has
to choose between a larger evidence store, a more elaborate search controller
and a different verdict stage therefore needs to know where the errors arise,
and end-to-end accuracy does not say.

One claim from our evaluation shows the problem. The FEVER claim states that
Labor Day is incapable of honouring the contributions that workers have made
to the prosperity of the country. Retrieval returns the annotated sentence
that says the holiday honours exactly those contributions. The judge reads it
correctly: its written reasoning concludes that the claim is refuted and it
marks the passage as refuting. Its verdict field nevertheless says
\emph{supported}. The validation step that follows finds no supporting passage
behind a supporting verdict and replaces the verdict with \emph{not enough
information}. The final answer is wrong, the retrieved evidence was complete,
so missing reference evidence does not explain this error. A retrieval
score would record a success, an accuracy score a failure, and neither would
show that the failure lies between the judge's reasoning and the system's own
safeguards.

This paper studies that boundary between finding evidence and using it. We
decompose verification into four operations with explicit interfaces:
\emph{retrieval}, which returns candidate passages for a query;
\emph{search control}, which decides what to ask and whether to ask again;
\emph{selection}, which fixes the bounded packet the judge may read; and
\emph{judgement}, which produces a verdict that is then checked by validation
rules. We hold the judge and the packet budget fixed and vary the first two
operations. Sparse, dense and hybrid retrieval are crossed with four search
controllers: a single pass, an adaptive gate, a question ledger and a
dependency controller. A graph-assisted retrieval condition and a tool agent
that chooses its own retrieval operations complete the comparison. For every
outcome we keep the evidence packet, so that the annotated reference evidence
can be traced through candidate pool, packet and verdict.

The comparison is run under two evidence conditions on the same 293 claims
from FEVER, FEVER 2.0 and AVeriTeC. In the first, each store contains only
passages selected through the reference annotations: 46 and 52 passages for
the two Wikipedia datasets and, for AVeriTeC, a separate store of annotated
question--answer evidence per claim. This is a control condition and not a
test of retrieval: the stores are so small that almost any search reaches the
evidence, which isolates what goes wrong after the evidence has been found. In
the second
condition the same claims are checked against the full source collections,
some twenty million Wikipedia passages and the complete archived web
collection of each AVeriTeC claim, with the model, the budgets and the judge
unchanged. This contrast shows whether conclusions drawn under easy discovery
survive when discovery becomes the hard part.

We ask four questions.
\begin{description}
\item[RQ1.] When evidence discovery is easy, how do the retrieval method and
the search controller change verdict accuracy, model-call cost and execution
reliability?
\item[RQ2.] When the complete reference evidence reaches the judge, where do
the remaining errors arise: in the judge's reading of the evidence, or in the
validation of its verdict?
\item[RQ3.] Do abstentions, validation conversions and inconsistent replies
persist when different models judge the same evidence packets?
\item[RQ4.] How do the answers to RQ1 and RQ2 change when the same claims are
checked against the full evidence stores?
\end{description}

The answers are as follows. In the control condition one hybrid retrieval pass
delivers a complete annotated evidence set for 226 of 228 evidence-bearing
claims, and no adaptive controller improves on its accuracy while each
multiplies the number of model calls. The errors that remain are downstream.
Of the 34 decisive claims that hybrid single pass labels \emph{not enough
information}, 33 have complete reference evidence in the packet; they divide
into abstentions of the judge and decisive verdicts that the validation rules
convert. On AVeriTeC the rules cost at least six accuracy points in every
matrix condition. Ten judges from eight model families, given identical
packets, also show abstentions and validation conversions; evidence
sufficiency was not audited in that block. Against the full stores the
accuracy of hybrid single pass falls by 17 to 29 points. Complete annotated
sentence sets reach fewer Wikipedia packets. For AVeriTeC the recovery
measure changes from annotated answers to cited pages, so its counts are
interpreted separately. The controllers
and the tool agent then do change what the judge reads, but none recovers the
loss, and their differences from a single pass stay within the uncertainty of
the cohort. A repetition of the Wikipedia comparison on 714 claims that the
study had not used before gives the same result for the agent. Repeated
executions also show that an identical request can receive a different
verdict. Server prompt caching is a possible explanation, but its causal role
was not isolated. In a further execution served by one loaded model instance,
variation on identical requests is reduced and the gate gains one to four
points; the ledger and the agent have no resolved gain. On Wikipedia the
accuracy loss falls on claims labelled not enough information. On AVeriTeC
it accompanies low recovery of cited pages from the full collections.

The contribution of the paper is a protocol for locating gaps in reference
evidence and changes to verdicts, and not only the negative result on adaptive search that it
produces here. It has four parts. The first is a controlled comparison of
retrieval methods and search controllers that reports accuracy together with
model-call cost and execution failures, so that the additional work of a
multi-step procedure is not mistaken for a better investigation. The second is
an audit that follows the reference evidence to the judge and separates the
judge's own abstentions from abstentions imposed by verdict validation, a
distinction that an end-to-end score hides. The third is the repetition of
the verdict audit with ten judge models on fixed packets, which measures how
abstentions and validation effects vary across models under shared evidence.
The fourth is the paired evaluation of the same claims, model and budgets under
annotation-selected and full evidence stores, which turns the usual caveat
about favourable evidence into a measured factor. The study is a diagnosis of
where a retrieval-augmented fact-checker loses verdicts; it does not propose a
new verification system and makes no comparison with benchmark leaderboards.

%% file: sections/related.tex
\section{Related work}\label{sec:related}
\subsection{Retrieval for verification}
Automated fact-checking is commonly organised as claim detection, evidence
retrieval and claim verification \citep{guo2022survey}, and
retrieval-augmented generation supplies a language model with retrieved text
at inference time \citep{lewis2020rag}. Sparse retrieval ranks passages by
lexical relevance; BM25 weights term occurrence and normalises for document
length \citep{robertson2009bm25}. Dense retrieval compares learned embeddings
of query and passage \citep{karpukhin2020dpr}. The two signals are
complementary, and hybrid search fuses them. Reciprocal-rank fusion combines
ranked lists without calibrating their scores \citep{cormack2009rrf};
\citet{bruch2023analysis} show that a convex combination of normalised scores
can rank better than reciprocal-rank fusion while recall is usually tied.
Because verification needs the passage to be present more than it needs it to
be ranked first, we use reciprocal-rank fusion followed by a cross-encoder
reranker. Dense encoders trained on verification benchmarks can overlap with
the evaluation data, so we use the Nomic encoder variant whose supervised
training mixture excludes FEVER \citep{nussbaum2025nomic}.

Graph-based retrieval adds structure extracted from the corpus. GraphRAG
builds an entity graph with community summaries for corpus-level questions
\citep{edge2024graphrag}; HippoRAG runs personalised PageRank over an
extracted knowledge graph and returns original passages
\citep{gutierrez2024hipporag}. Controlled comparisons agree that graphs are
not uniformly better than passage retrieval: plain retrieval matches or beats
graph methods on single-hop and detail questions, and graph methods help as
multi-hop reasoning becomes necessary
\citep{han2025rag,xiang2026when,zhou2025indepth}. Our graph-assisted condition
follows the passage-preserving design, in which graph paths nominate source
passages and the judge reads only source text.

The work reported here grew out of a pipeline for annotating disinformation
outlined as a simplified concept in
\ifanonymous earlier work \citep{pavlicek2026pipeline}\else our earlier work
\citep{pavlicek2026pipeline}\fi. That paper proposes decomposing a text into
claims, filtering them for check-worthiness and passing them to several
detectors chosen for their different failure modes, one of them
evidence-grounded, with a combiner that merges their outputs into an
annotation with a verdict, cited evidence and a rationale. Its first
measurement, which the paper itself calls a shakedown, ran four detectors on
300 FEVER claims and did not separate retrieval from judgement. The
fact-checker studied here is not the detector of that paper: its retrieval,
search control and verdict validation were developed since. The earlier paper
is the baseline concept from which this work started, and the present study
examines the evidence-grounded path in depth.

\subsection{Iterative, adaptive and agentic search}
A second line of work lets the model decide when and what to retrieve. IRCoT
interleaves retrieval with chain-of-thought steps and improves recall on
multi-hop questions over one-step retrieval \citep{trivedi2023ircot}. FLARE
retrieves when the next sentence is generated with low confidence and finds
that retrieving at every step is not best \citep{jiang2023flare}. Self-RAG
trains a model to emit tokens that request retrieval and critique its own
output \citep{asai2024selfrag}. ReAct interleaves reasoning with tool actions
and observations \citep{yao2023react}, the pattern behind current search
agents. Adaptive-RAG routes each query to no retrieval, single-step or
multi-step retrieval and reports that the multi-step strategy is the most
accurate but several times slower \citep{jeong2024adaptive}.

More recent studies measure what the additional search costs.
\citet{mccleary2026quantifying} vary the design of a budget-constrained search
agent and find that accuracy rises up to about three searches and then
plateaus, with hybrid retrieval and reranking giving the most consistent gain.
\citet{wu2025search} quantify searches that the model did not need and steps
where it should have searched and did not. \citet{liu2026search} report that
small models used as search agents call the search tool less often than larger
ones and break the tool-call syntax. These studies use open-domain question
answering. Our comparison asks the same cost question for claim verification,
where the output is a three- or four-way verdict that includes abstention, and
it adds a factor those studies hold fixed: how hard the evidence is to find.

In fact-checking, question-guided procedures are the common form of adaptive
search. QACheck chains a sufficiency check, question generation, question
answering and a final reasoner \citep{pan2023qacheck}. InFact, the winning
system of the first AVeriTeC shared task, runs six stages from claim
interpretation to justification \citep{rothermel2024infact,
schlichtkrull2024averitecshared}. Papelo pursues missing evidence with
follow-up questions and reports gains over generating all questions at once
\citep{malon2024papelo}; HerO retrieves with hypothetical fact-checking
passages and open-weights models \citep{yoon2024hero}. FIRE decides at each
step between a verdict and another web search and observes that small models
repeat near-identical queries and that forcing additional searches raises cost
without improving accuracy \citep{xie2025fire}. \citet{tian2024web} find that
a web-search agent improves verdicts on political claims and brings no gain on
Wikipedia-derived FEVER claims. Our gate, ledger and dependency controller are local
implementations of the design choices these systems embody: a sufficiency
decision, an explicit question plan and the binding of one answer into the
next query. They are not reproductions of the published systems, and their
scores are not comparable with shared-task results, which are obtained with
different models, evidence access and scoring.

\subsection{Evidence sufficiency and abstention}
Whether the evidence suffices is a question separate from whether it was
retrieved. \citet{atanasova2022insufficient} remove information from evidence
to test whether verification models recognise the resulting insufficiency,
and find that sensitivity depends on what was removed. \citet{joren2025sufficient} label retrieved context as
sufficient or insufficient for a query and find that small models abstain or
err even when the context is sufficient, while large models answer instead of
abstaining when it is not. \citet{madhusudhan2025abstention} report that
abstention remains unreliable on reasoning questions even for strong models.
In the second AVeriTeC shared task no system scored above 0.1 on the
not-enough-evidence and conflicting-evidence classes
\citep{akhtar2025averitec}. A further complication is that a model's stated
reasoning need not determine its answer \citep{turpin2023unfaithful}. We
measure sufficiency through the benchmark annotations instead of an automatic
rater: a packet that contains a complete annotated evidence set is treated as
having received what the annotators judged sufficient, and the verdict is
examined from there. This lets us separate two events that both surface as an
insufficient-evidence verdict, the judge abstaining and a validation rule
overriding the judge.

\subsection{Benchmarks and evidence conditions}
FEVER pairs Wikipedia-derived claims with sentence-level evidence and three
labels \citep{thorne2018fever}; FEVER 2.0 adds adversarially written claims
\citep{thorne2019fever2}. AVeriTeC contains real-world claims with
question--answer evidence from the web and a fourth label for conflicting
evidence or cherry-picking \citep{schlichtkrull2023averitec}. The evidence a
system may search strongly shapes what its score means.
\citet{glockner2022missing} argue that realistic evidence must be sufficient
and must not leak the fact-check itself, and find leaked evidence for most
misinformation claims in a web-evidence dataset. Our annotation-selected stores
are the favourable extreme of this spectrum, since they are built from the
evidence the annotators used; the full stores remove that advantage for
Wikipedia and replace annotation text by archived page text for AVeriTeC. We
report the two conditions side by side and do not equate either with open web
search.

%% file: sections/method.tex
\section{Retrieval, search control and judgement}\label{sec:method}
\begin{figure}[htbp]
\centering
\includegraphics[width=\linewidth]{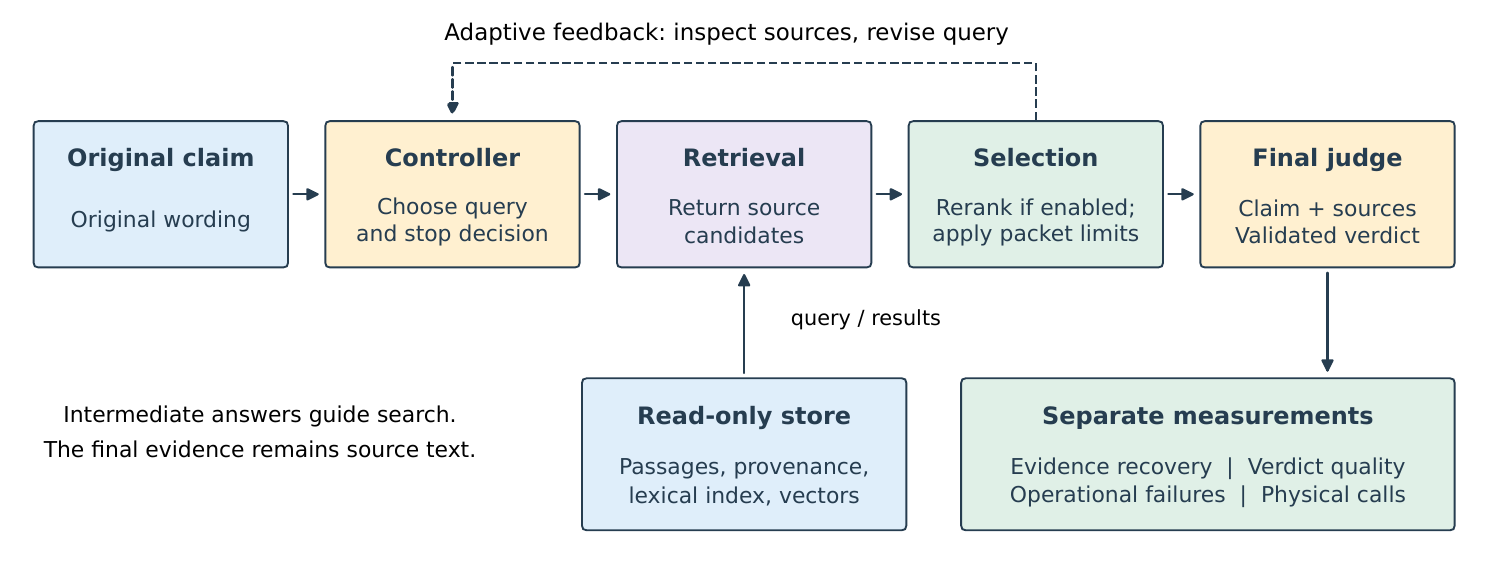}
\caption{Search control, retrieval, selection and judgement. The dashed
feedback loop exists only in the adaptive controllers. The judge reads source
passages after the investigation has ended.}
\label{fig:method-overview}
\end{figure}

For a claim $c$ and a store $S$, retrieval returns a ranked candidate list
$R(q,S)$ for a query $q$. A search controller chooses the queries and decides
when to stop. After round $t$ the accumulated candidates and the evidence
packet are
\begin{equation}
 U_t=\operatorname{dedup}\bigl(U_{t-1}\cup R(q_t,S)\bigr),\qquad
 E_t=\operatorname{select}(c,U_t;B),
\end{equation}
where $B$ is a fixed budget of at most eight passages, 2,400 characters per
passage and 16,000 characters in total. The judge reads the claim and the final
packet and returns a verdict $J(c,E)$, which validation rules may then change.
Figure~\ref{fig:method-overview} shows the interfaces. Two properties of this
decomposition matter for what follows. A controller's intermediate answers
guide its next query but never reach the judge, which reads source passages
only. And because selection is repeated against the original claim whenever
candidates arrive, a later packet is not necessarily a superset of an earlier
one: a newly retrieved passage can displace a premise that had been selected.

We use one running example. The claim \emph{The founder of Acme was born in
France} is constructed so that three passages are needed: $p_1$, \emph{Mira
founded Acme}; $p_2$, \emph{Mira was born in Alder}; and $p_3$, \emph{Alder is
a city in France}. A single query may or may not retrieve all three; $p_3$
shares no content word with the claim except the country.

\subsection{Retrieval methods}
Sparse retrieval uses BM25 over a full-text index. The query is the lowercased
claim with common words removed and the remaining terms joined disjunctively,
so a passage qualifies by matching any term. Dense retrieval embeds the claim
with \texttt{nomic-embed-text-v1-ablated}, a 768-dimensional encoder whose
supervised training mixture excludes FEVER \citep{nussbaum2025nomic}, and ranks
passages by cosine similarity. Hybrid retrieval fuses the two rankings by
reciprocal-rank fusion, $\sum_{L\ni d} 1/(60+r_L(d))$ with $r_L(d)$ the
zero-based rank of passage $d$ in list $L$; the common formulation counts
ranks from one, which shifts every denominator by one. Every search returns at most 40
candidates. Where reranking is enabled, the cross-encoder
\texttt{ms-marco-MiniLM-L-6-v2} scores the claim against each candidate's
title and text, and selection follows that order.

\begin{figure}[htbp]
\centering
\includegraphics[width=\linewidth]{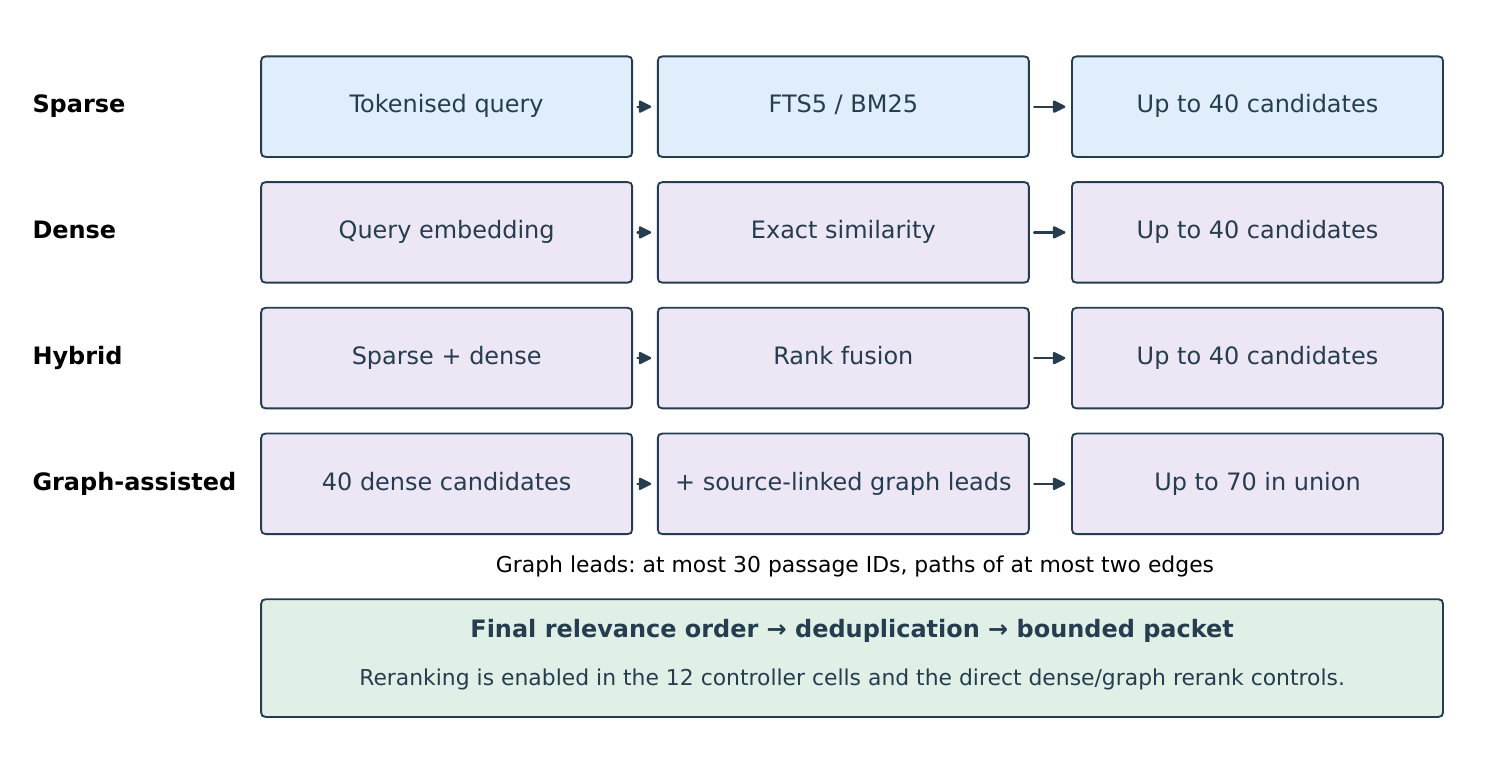}
\caption{The four retrieval methods within one search round. All end in the
same bounded packet.}
\label{fig:retrieval-methods}
\end{figure}

Graph-assisted retrieval starts from the 40 dense candidates. Entities in the
claim are matched to an entity graph extracted from the store, paths of at most
two edges are followed, and the passages from which the traversed relations
were extracted are added, up to 30 of them. The union of at most 70 passages is
reranked. In the example, the path Acme--Mira--Alder would nominate $p_1$ and
$p_2$. The graph only nominates passages; relations are never shown to the
judge as evidence. Figure~\ref{fig:retrieval-methods} summarises the four
methods.

\subsection{Search controllers}\label{sec:controllers}
\begin{figure}[htbp]
\centering
\includegraphics[width=\linewidth]{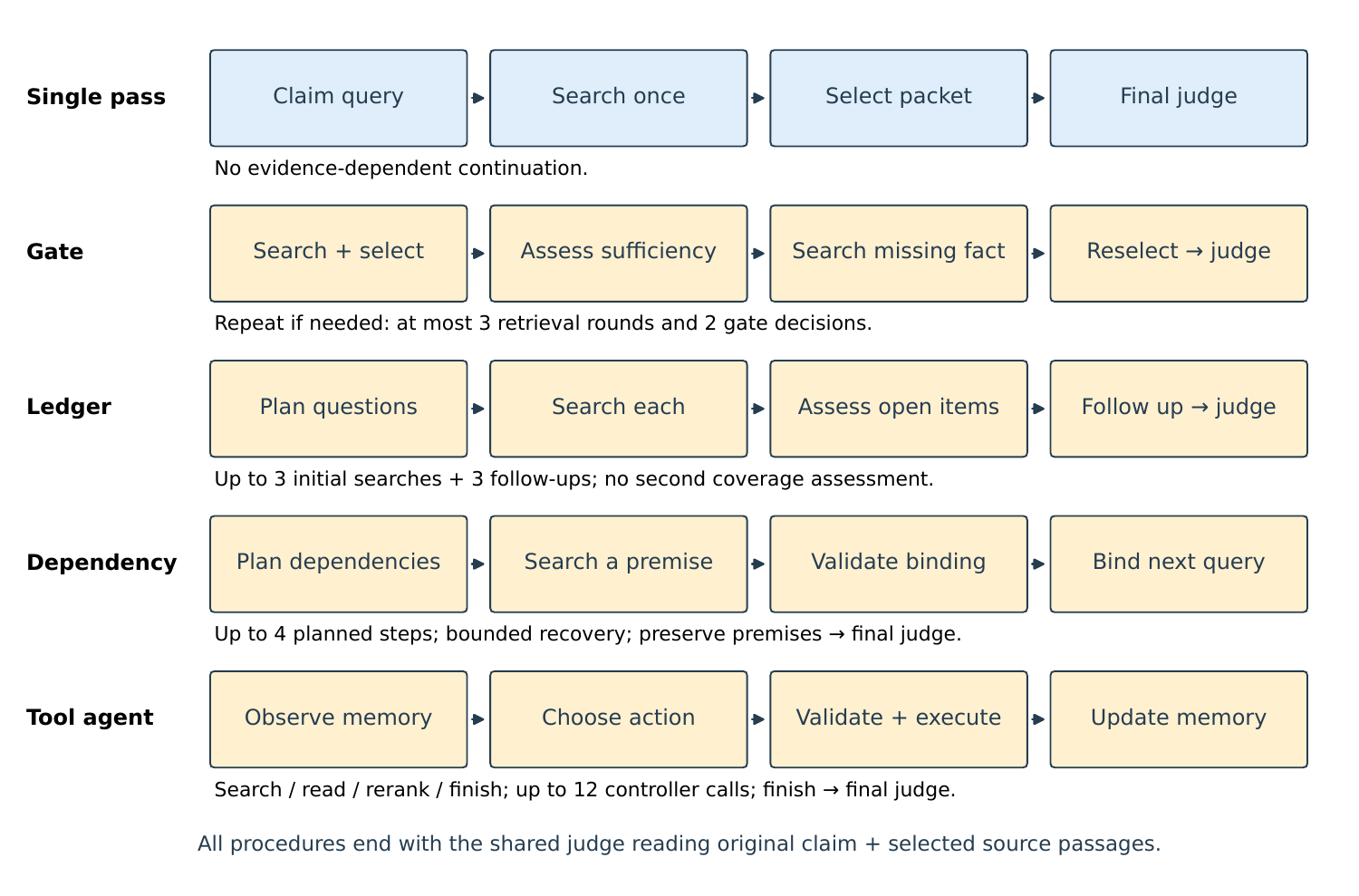}
\caption{The five search controllers and their bounds.}
\label{fig:controllers}
\end{figure}

The five controllers differ in the state they keep and in the decision they
delegate to the language model (Figure~\ref{fig:controllers}). All end with the
same judge call, except that the dependency controller passes its accepted
chain to the judge in an extended prompt and adds a validation rule for that
chain.

\emph{Single pass} issues the claim as the only query and stops.

The \emph{adaptive gate} searches with the claim, shows the packet to the model
and asks whether it suffices and, if not, which fact is missing. A proposed
follow-up query is checked for length, repetition and embedded instructions and
then executed. At most three retrieval rounds and two gate decisions are
allowed. In the example, a gate that has seen $p_1$ and $p_2$ can ask which
country Alder is in.

The \emph{question ledger} plans before it searches. The model writes up to
three verification questions, each is searched, and the model then marks which
questions the packet answers and proposes one follow-up query for each open
question. The ledger thus makes at most three initial and three follow-up
searches, with one model call for planning, up to two for assessment and up
to two that check a proposed answer.

The \emph{dependency controller} represents explicitly that one question needs
the answer to another. It plans up to four steps; after each search the model
proposes an answer with a supporting quotation, the program checks the
quotation against the source, and the accepted value is substituted into the
next query. \emph{Who founded Acme?} must yield \emph{Mira} from a source before
\emph{Where was Mira born?} is asked. An answer that fails the check is not
replaced by a guess; a bounded recovery step may reformulate the query.
Selection keeps the premise passages of an accepted chain together.

The \emph{tool agent} chooses its own operations. At each step it returns one
structured action: a sparse, dense or hybrid search, a read of retrieved
passages with optional neighbouring text, a rerank of the collected pool, or a
finish that nominates the final packet. It may use twelve controller calls,
six searches and three reranks per claim. A passage must have been read before
it can support a working finding or enter the packet. Because the agent
selects the retrieval method itself, it is one complete procedure and not a
cell of the retrieval-by-controller matrix.

\subsection{Judge and verdict validation}\label{sec:validation}
The judge receives the claim and the packet and returns a structured reply: a
verdict (supported, refuted, not enough information, or conflicting evidence),
a stance and confidence for each passage, the cited passages, the share of the
claim's assertions that the evidence covers, and a free-text justification.
Conflicting evidence is scored against AVeriTeC's fourth class and mapped to
not enough information on the three-label Wikipedia datasets.

A decisive verdict is accepted only if the reply supports it. Four validation
rules, applied by the program after the judge has answered, enforce this.
\begin{description}
\item[Stance alignment.] A supported or refuted verdict needs at least one
passage that the judge itself assessed with the matching stance at confidence
0.6 or higher.
\item[Coverage.] A supported verdict needs the judge to report that the
evidence covers the assertions of the claim.
\item[Conflict declaration.] Passages assessed in opposite directions require
the conflicting-evidence verdict, and that verdict requires such passages on
both sides.
\item[Attribution review.] If the claim contains a reporting construction
such as \emph{said}, \emph{wrote}, \emph{announced}, \emph{declared} or
\emph{according to}, and the verdict is decisive, a second model call audits
whether the cited passages establish the speaker, the statement and its time,
and may withhold permission for the verdict.
\end{description}
A verdict that fails a rule is replaced by not enough information. The rules
implement a conservative policy: the system prefers to abstain when it cannot
show the passage behind a verdict. Their effect on correct and incorrect
verdicts is measured in Section~\ref{sec:rq2}.

A judge reply that violates the output schema, or a controller that does not
complete within its limits, is an operational failure. Failures are kept as a
separate outcome and are never counted as abstentions.

%% file: sections/design.tex
\section{Experimental design}\label{sec:design}
\subsection{Claims}
The main evaluation uses 293 claims: 100 from FEVER, 93 from FEVER 2.0 and 100 from
AVeriTeC (Table~\ref{tab:cohorts}). The Wikipedia cohorts are balanced across
the three labels. The AVeriTeC cohort follows the skew of its source, with two
thirds of the claims refuted and only seven and six claims in the two smallest
classes; a constant \emph{refuted} prediction would score 66\% there, which is
the reference point for every AVeriTeC accuracy in this paper. Several
Wikipedia claims are variants of one another. We group such claims into
families and resample whole families when estimating uncertainty. The FEVER
and AVeriTeC claims were drawn from the publishers' training splits and the
FEVER 2.0 claims from its development set, in each case stratified by label.
Seven of the 100 FEVER 2.0 claims drawn were removed before any prediction was
made, because their claim families occur in training data used elsewhere in
the project. All 293 claims
were also used while the system was developed, and errors on the FEVER 2.0 and
AVeriTeC claims had been read when earlier judge rules were written; those
rules are switched off here. The
results on them therefore describe
behaviour on known material and are not an estimate for unseen claims; the
714 further Wikipedia claims of Section~\ref{sec:setup} were not used before.
Development exposure can affect both absolute accuracy and comparisons
between components. Its direction and magnitude were not measured, so the
negative findings on control and validation are exploratory.

\begin{table}[htbp]
\centering\small
\begin{tabular}{lrrrrrr}
\toprule Dataset & Claims & Supported & Refuted & NEI & Conflicting & Families \\
\midrule
FEVER & 100 & 33 & 33 & 34 & -- & 43 \\
FEVER 2.0 & 93 & 30 & 32 & 31 & -- & 68 \\
AVeriTeC & 100 & 21 & 66 & 7 & 6 & 100 \\
\bottomrule
\end{tabular}
\caption{Evaluation claims by reference label.}
\label{tab:cohorts}
\end{table}

\subsection{Two evidence conditions}\label{sec:stores}
Each claim is checked against two stores (Table~\ref{tab:stores}).

The \emph{annotation-selected} stores are the control condition of the study.
They contain only passages that the reference annotations point to, so that
discovery is nearly certain by construction and the errors that remain can be
assigned to the steps after it. A result on retrieval or search control in
these stores is therefore expected to be null, and we read it as a check that
the control works, not as evidence about search. For each Wikipedia dataset one store holds the annotated
passages of all its claims, 46 passages for FEVER and 52 for FEVER 2.0. A query
can therefore retrieve passages that belong to other claims, but the candidate
limit of 40 covers most of the store. For AVeriTeC every claim has its own
store with the annotators' question--answer pairs for that claim, 282 units in
total and between one and ten per claim. Only one claim has more units than
the packet of eight can hold, so for the other 99 a search that reaches the
whole store can change the order of the evidence and not its content. A unit
holds the question, the
answer and, where present, the annotator's explanation; it is not the text of
the cited web page. The reference label is never part of a store, a query or a
judge input, but an annotator's answer can state the fact-checker's
conclusion in its own words.

Wikipedia claims labelled not enough information contribute no passage to
their store. For 39 of these 65 claims the store therefore holds nothing on
the subject of the claim; the other 26 belong to the family of an
evidence-bearing claim whose annotated sentences are in the store.

The \emph{full} stores remove the selection. For both Wikipedia datasets the
store is the complete Wikipedia snapshot distributed with FEVER, 5.4 million
pages split into 20.8 million passages of two consecutive sentences. It
contains the annotated sentences among the rest, so the Wikipedia contrast
changes the number of distractors and the unit of a passage, and leaves the
evidence itself in place. For AVeriTeC the store of a claim is its collection of archived
web documents, 72,768 documents and 55.0 million passages in total across the
hundred claims. Here the
evidence appears as page text and no longer as an annotator's answer, so the
AVeriTeC contrast changes the representation of the evidence together with the
size of the store. We therefore treat the Wikipedia contrast as the test of
store growth and report the AVeriTeC contrast separately as a change of
evidence source.
The reference evidence survives the change of store completely for Wikipedia
and partly for AVeriTeC. All 128 evidence-bearing Wikipedia claims have at
least one complete annotated set in the snapshot. Of the 99 AVeriTeC claims
whose annotations cite a web page, 39 have every cited page in their
collection and 66 at least one. Recovery for AVeriTeC is therefore counted at
the level of the cited page in this condition.

\begin{table}[htbp]
\centering\small
\begin{tabularx}{\linewidth}{lXX}
\toprule
 & Annotation-selected & Full \\
\midrule
FEVER & one store, 46 passages & \multirow{2}{=}{one shared store, 5,396,106 pages, 20,845,882 passages} \\
FEVER 2.0 & one store, 52 passages & \\
AVeriTeC & 100 stores, 282 question--answer units & 100 stores, 72,768 web documents \\
Evidence text & annotated sentences; annotators' answers & article passages; archived page text \\
Entity graph & available & not built \\
\bottomrule
\end{tabularx}
\caption{The two evidence conditions. Claims, model, candidate limit and packet
budget are identical in both.}
\label{tab:stores}
\end{table}

\subsection{Conditions}
In the annotation-selected stores we run the complete comparison. Sparse, dense
and hybrid retrieval are crossed with the single-pass, gate, ledger and
dependency controllers, with reranking enabled in all twelve cells. Four
further single-pass conditions isolate individual components: sparse and dense
retrieval without reranking, dense retrieval with reranking through the direct
search path, and graph-assisted retrieval with reranking. These sixteen
conditions on 293 claims give 4,688 verification tasks. The tool agent adds
one condition and 293 tasks. It was executed separately with a longer task
limit, so we compare its model calls with the other conditions and not its
elapsed time.

In the full stores we run the five conditions that the first comparison shows
to be informative: sparse, dense and hybrid single pass, the hybrid gate and
the hybrid ledger, all with reranking. The dependency controller is left out
because its failures in the selected stores are failures of execution that a
larger store cannot repair, and the graph-assisted condition because an entity
graph of the full collections was not built. The tool agent is run on the
full stores as well, with its settings unchanged. Hybrid single pass is the
reference condition in both stores.

\subsection{Judge comparison on fixed packets}\label{sec:judges-design}
To test whether the downstream errors belong to one model, ten judges read
identical evidence. For each of the 293 claims one packet was produced by
graph-assisted retrieval with reranking over annotation-selected evidence and
then frozen; every judge received exactly that packet. The judges are three
Qwen models of 9, 27 and 122 billion parameters and seven models from other
families between 14 and 32 billion parameters (Table~\ref{tab:judges}); we
name them by the tags under which they were installed. All ran with the
generation settings of the main comparison. This block was run before the main
comparison, with an earlier version of the judge prompt and without the
attribution review; the other three validation rules were in force. Its
Wikipedia packets were drawn from one store shared by FEVER and FEVER 2.0, and
the reference evidence in its packets was not traced. For these reasons the block is analysed on its own: its judges are compared with
one another, and its numbers are not pooled with those of the matrix.

\subsection{Model and settings}\label{sec:setup}
Outside the judge comparison, all controller and judge calls use Qwen 3.5 with 9 billion parameters in 4-bit
quantisation, served locally. Generation is deterministic in its settings
(temperature zero, fixed seed) with a context of 16,384 tokens and at most
3,200 output tokens for the judge. A matrix task may take 600 seconds and a
single model request 240 seconds; the agent may take 1,800 seconds per claim.
The main tables report one execution per task. Hybrid single pass, the
reference condition, was executed a second time on all claims in each store,
to measure how far two executions of the same condition differ
(Sections~\ref{sec:rq1} and~\ref{sec:rq4}). The tool agent's run on the
selected stores was repeated on the machine that served the full stores, and
the repeat is the run we report (Section~\ref{sec:discussion}). The five
full-store conditions and the agent were executed once more with equal batch
sizes for all calls, which lets one loaded instance of the model serve the
whole comparison (Section~\ref{sec:rq4}). To test the Wikipedia results
of the full-store condition on more claims, hybrid single pass, twice, and
the tool agent were also run on 714 further claims that no earlier step of
the study had used: 216 from the FEVER development set and 498 from FEVER
2.0, in equal classes, against the same Wikipedia store and with unchanged
settings (Section~\ref{sec:rq4}). Equal settings did not make the replies
reproducible across conditions: where two conditions sent the judge an
identical request, the reply was sometimes different
(Section~\ref{sec:rq1}). The identities of model, encoder and reranker are
listed in Appendix~\ref{app:identities}.

\subsection{Measures}
\paragraph{Verdict accuracy.}
Accuracy is the share of claims whose final verdict equals the reference label.
An operational failure counts as incorrect, so every condition is scored on all
claims of the dataset. Where class balance matters we add balanced accuracy,
the mean recall over the reference classes.

\paragraph{Cost and reliability.}
Cost is the number of language-model calls, including controller calls,
attribution reviews and retries, divided by the number of claims. It leaves out
embedding, reranking and indexing. Reliability is the number of operational
failures, reported by the component that failed.

\paragraph{Paired change.}
Against hybrid single pass, a \emph{repair} is a claim the reference condition
gets wrong and the alternative gets right, and a \emph{damage} is the reverse;
the accuracy difference is repairs minus damages over the number of claims. We
give 95\% percentile intervals from 10,000 bootstrap samples of claim families.
The intervals are pointwise and not corrected for the number of comparisons,
and we read them as a description of these claims. For the repairs and
damages of a controller we also state the exact McNemar probability in the text. Its resolution
on a hundred claims is coarse: with ten discordant claims it separates a
difference from zero only from a net eight claims, with twenty from ten and
with thirty from twelve. A statement that a controller brings no gain
therefore means no gain of that size, and not that an effect is shown to be
absent. The paired analysis and
the two audits below were defined after the runs had completed.

\paragraph{Evidence recovery.}
Every saved passage carries its source identity, which we match against the
annotated evidence. FEVER annotations list one or more alternative evidence
sets per claim. With $\mathcal{A}_i$ the alternatives for claim $i$ and $P_i$
the annotated sentences found at a given stage, \emph{complete recovery} is
\begin{equation}
 C_i=\mathbf{1}\bigl(\exists A\in\mathcal{A}_i:\ A\subseteq P_i\bigr).
\end{equation}
For AVeriTeC in the selected stores, complete recovery means that all
annotated question--answer units of the claim are present. In the full stores
it means that at least one passage from every cited page is present, for the
99 claims with a page reference. This page-level measure does not establish
that the relevant text within each page was retrieved and is not directly
comparable with recovery of annotated answers. We evaluate recovery on the candidate pool and on the final packet,
which separates evidence that retrieval did not find from evidence that
selection dropped. Wikipedia claims labelled not enough information have no
annotated evidence and are excluded from recovery; all AVeriTeC claims carry
annotated evidence. This leaves 66, 62 and 100 \emph{evidence-bearing} claims;
the AVeriTeC claim without a page reference has no assessable full-store
recovery. Recovery of a complete sentence set or all annotated answers
records delivery of the reference evidence, not an independent judgement of
entailment. Page recovery provides a weaker source-coverage measure.

\paragraph{Validation audit.}
For every completed task we compare the judge's verdict before validation with
the final verdict. A \emph{conversion} is a verdict changed by one of the rules
of Section~\ref{sec:validation}. A conversion is harmful if the judge's verdict
matched the reference and the final verdict does not, beneficial if the
reverse holds, and neutral otherwise. Since the rules only ever convert a
verdict to not enough information, a beneficial conversion is one that turns a
wrong verdict on a not-enough-information claim into the right abstention. Accuracy before validation is the accuracy the same run would have
obtained with the rules switched off.

\paragraph{Residual abstentions.}
The hybrid single-pass audit covers the saved traces of every claim that has a
supporting or refuting reference, complete recovery in the packet and a final
verdict of not enough information. The split between the judge's own
abstention and a validation conversion is computed from the saved replies.
Claude (Anthropic) proposed the finer categories from the judge's
justifications, passage assessments and evidence packets. These categories
were developed during inspection rather than specified in advance and are
listed with every claim in Appendix~\ref{app:cases}. They are exploratory,
AI-assisted interpretations, not an independently validated annotation set;
the numerical split between judge abstentions and conversions does not
depend on them.

\paragraph{Store contrast.}
The effect of the store on a condition is its paired accuracy change from the
selected to the full store on the same claims. The question of RQ4 is an
interaction: whether the gain of an adaptive controller or of the tool agent
over single pass differs between the stores. We estimate it as the difference of the two
paired changes, with the same family bootstrap. For the full stores we also
classify each error on an evidence-bearing claim by the first observed gap in
reference recovery: absent from the store, absent from the candidate pool,
dropped by selection, or present in the packet with a wrong verdict;
operational failures are recorded separately. This locates reference-evidence
gaps but does not establish their causal contribution to an error, since
unannotated alternative evidence may suffice.

%% file: sections/results.tex
\section{Results}\label{sec:results}
\subsection{RQ1: retrieval and search control when discovery is easy}\label{sec:rq1}
Table~\ref{tab:core} gives the main conditions in the annotation-selected
stores, and Appendix~\ref{app:matrix} lists all seventeen. Of the 4,688 matrix
tasks, 4,426 completed and 262 ended in an operational failure; the matrix
used 15,829 model calls and the agent a further \val{agent.selected.calls.all}.

\begin{table}[htbp]
\centering\small
\begin{tabular}{lrrrrrr}
\toprule
 & \multicolumn{3}{c}{Accuracy, \% (failures)} & \multicolumn{3}{c}{Model calls per claim} \\
\cmidrule(lr){2-4}\cmidrule(lr){5-7}
Condition & FEVER & FEVER 2.0 & AVeriTeC & FEVER & FEVER 2.0 & AVeriTeC \\
\midrule
Sparse, no reranking & 83.0 (2) & 68.8 (5) & 62.0 (2) & 1.02 & 1.00 & 1.13 \\
Dense, no reranking & 81.0 (2) & 71.0 (1) & 61.0 (1) & 1.02 & 1.00 & 1.13 \\
Sparse single pass & 80.0 (1) & 69.9 (3) & 63.0 (1) & 1.02 & 1.00 & 1.13 \\
Dense single pass & 81.0 (1) & 74.2 (3) & 65.0 (1) & 1.02 & 1.00 & 1.13 \\
Hybrid single pass & 81.0 (1) & 74.2 (3) & 65.0 (1) & 1.02 & 1.00 & 1.13 \\
Graph-assisted & 81.0 (2) & 74.2 (3) & 65.0 (1) & 1.02 & 1.00 & 1.13 \\
\midrule
Hybrid gate & 81.0 (4) & 68.8 (2) & 65.0 (1) & 2.47 & 2.53 & 2.13 \\
Hybrid ledger & 81.0 (2) & 69.9 (2) & 64.0 (3) & 4.42 & 4.43 & 4.61 \\
Hybrid dependency & 67.0 (16) & 51.6 (27) & 45.0 (15) & 8.44 & 7.89 & 9.83 \\
\input{generated/rq1-agent-row.tex} \bottomrule
\end{tabular}
\caption{Annotation-selected stores. Accuracy counts operational failures as
incorrect. All conditions below the first two rows use the reranker, except
the tool agent, which calls it as a tool and did so in \val{agent.selected.claims_with_rerank.all} of 293
investigations. Single pass exceeds one call per claim through attribution
reviews.}
\label{tab:core}
\end{table}

\paragraph{Retrieval method.}
Once the reranker is applied, the choice between dense and hybrid retrieval
has no effect on the verdicts. The two conditions agree on the correctness of
every one of the 293 claims, and they deliver the identical ordered packet for
97 of 100 FEVER claims, 88 of 93 FEVER 2.0 claims and all AVeriTeC claims.
Graph-assisted retrieval returns exactly the packets of reranked dense
retrieval for all 293 claims: the graph nominates passages, but none of them
displaces a passage that dense retrieval and the reranker had already placed
in the packet. Dense single pass, direct dense retrieval with reranking
(Appendix~\ref{app:matrix}) and graph-assisted retrieval are thus one
procedure on these stores, and the single FEVER failure by which their rows
differ arose on an identical judge request. Sparse retrieval with reranking is lower by 1.0, 4.3 and 2.0
points, with intervals that reach zero on all three datasets. Without the
reranker the picture is mixed: sparse retrieval is the best condition on FEVER
(83.0\%) and among the weakest on FEVER 2.0 (68.8\%). No retrieval method is
better on all three datasets, and no interval excludes zero at this sample
size.

These stores are too small for retrieval to matter much, and one comparison
shows how small the remaining differences are. On AVeriTeC, dense retrieval
with and without the reranker puts the same set of passages in the packet for
99 of 100 claims and orders them differently for 31. The judge reads the same
evidence in a different order, and accuracy is 65.0\% with the reranker and
61.0\% without it: removing the reranker repairs one verdict and damages five.
Order is not the only source of such changes. Where two conditions sent the
judge a byte-identical request, the reply was not always the same: between
hybrid single pass and the hybrid gate this happened to the correctness of
two of the 100 AVeriTeC verdicts. The four points therefore combine the effect
of passage order with the variation between two executions. A second
execution of hybrid single pass on the same stores measures the latter
directly. The judge received an identical request for \val{repeat.selected.identical_final_request.all} of 293 claims and
returned a different verdict for \val{repeat.selected.verdict_changed.all} of them. Correctness changed for \val{repeat.selected.changed_correctness.f}
FEVER claims, \val{repeat.selected.changed_correctness.ft} FEVER 2.0 claims and \val{repeat.selected.changed_correctness.a} AVeriTeC claims, and accuracy
moved by one claim on each dataset, since the changes nearly cancel. We
therefore do not interpret differences of up to four points between
conditions as evidence of a method advantage on their own. Server logs show
reuse of cached prompt states, which suggests request history as one possible
source of variation; this mechanism was not isolated experimentally. Some
executions agree exactly: two consecutive executions returned identical replies for all
\val{fresh.identical_replies.all} further claims of Section~\ref{sec:rq4}, and for all 293 claims of the
full-store condition when these were issued once in the original and once in
the reversed order. That pair nevertheless changed
\val{order.same.verdict_changed.all} of 293 verdicts against an earlier
execution on identical requests. The earlier execution remains unexplained,
and the order test does not establish a general cause of the variation.
In the main comparison the controllers and the judge requested different batch sizes,
which made the server reload the model, and with it empty its store of
states, about a hundred times an hour. Section~\ref{sec:rq4} repeats the
full-store conditions without these reloads.

\begin{figure}[htbp]
\centering
\includegraphics[width=\linewidth]{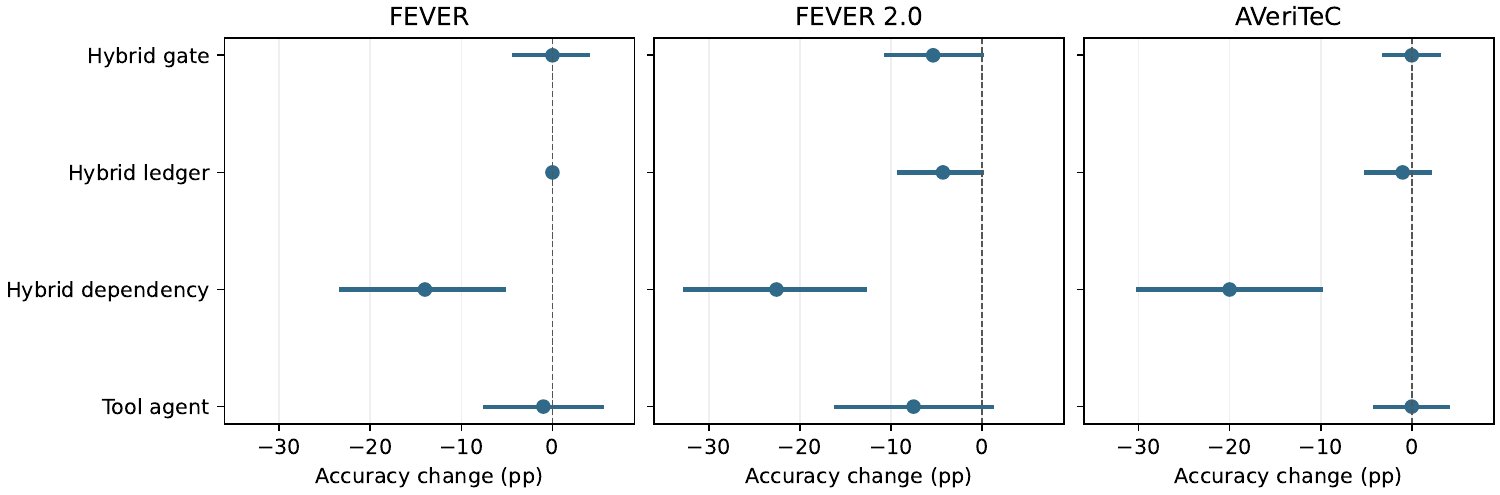}
\caption{Paired accuracy change against hybrid single pass in the
annotation-selected stores, with 95\% family-bootstrap intervals.}
\label{fig:paired-changes}
\end{figure}

\begin{figure}[htbp]
\centering
\includegraphics[width=\linewidth]{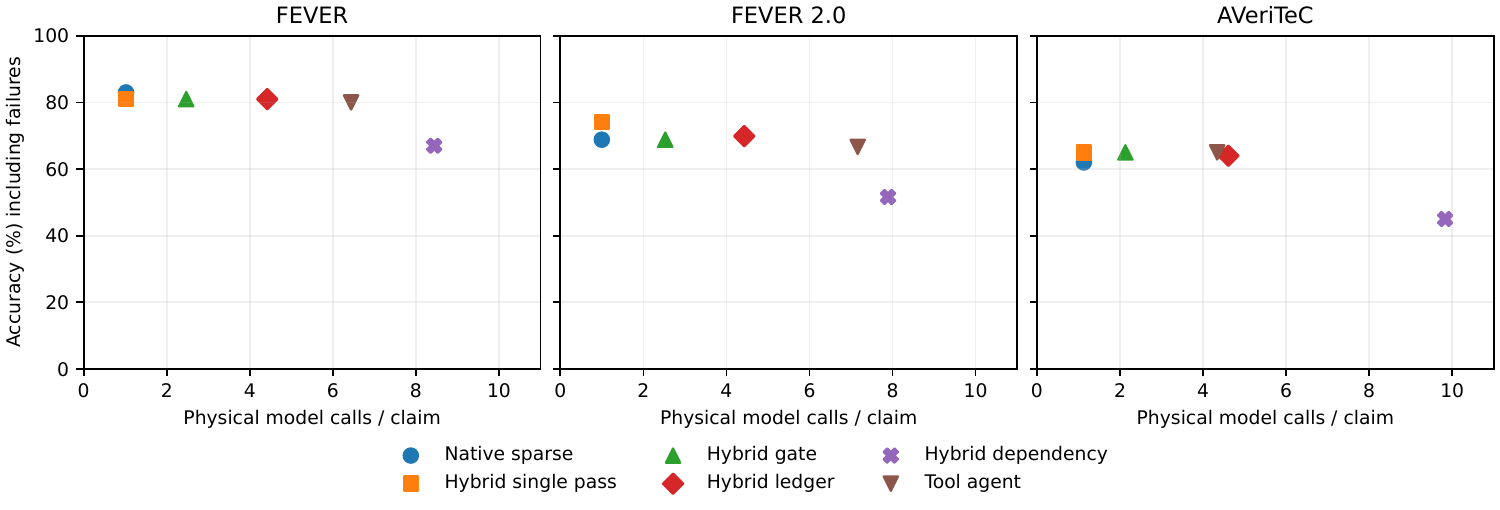}
\caption{Accuracy against model calls per claim in the annotation-selected
stores.}
\label{fig:cost-quality}
\end{figure}

\paragraph{Search controllers.}
No adaptive controller improves on single pass (Table~\ref{tab:paired},
Figure~\ref{fig:paired-changes}). The gate leaves accuracy unchanged on FEVER
and AVeriTeC and loses 5.4 points on FEVER 2.0, where it repairs one verdict
and damages six. The ledger changes the correctness of no FEVER verdict, and loses
4.3 and 1.0 points on the other two datasets. The tool agent is lower by
\val{agent.selected.deficit.f} and \val{agent.selected.deficit.ft} points on FEVER and FEVER 2.0 and level with
single pass on AVeriTeC. For all three controllers the intervals include zero. What does
change is the cost: the gate needs 2.1 to 2.5 calls per claim, the ledger 4.4
to 4.6 and the agent \val{agent.selected.calls_per_claim.a} to \val{agent.selected.calls_per_claim.ft}, against 1.0 to 1.1 for single pass
(Figure~\ref{fig:cost-quality}).

\input{tables/paired}

The reason is visible in the evidence. Hybrid single pass already has a
complete annotated evidence set in its candidate pool for every one of the 228
evidence-bearing claims and in its packet for 226 of them
(Table~\ref{tab:full-recovery}, columns \emph{Selected}). The gate and the ledger end with complete evidence
in the packet for the same number of claims on every dataset. Their additional
searches cannot add reference evidence that the first search has already
delivered. Nor do they change what the judge reads in most cases: the final
judge request of the gate is identical to that of single pass for 90, 83 and
100 claims of the three datasets, and that of the ledger for 69, 66 and 99.
Of the 22 verdicts whose correctness the gate or the ledger changes in
Table~\ref{tab:paired}, 21 lie among these claims. The repairs and damages of
the two controllers are therefore, with one exception, differences between two
executions of the same judge request and not effects of the evidence the
controller gathered. The agent shows the redundancy from the inside: on the
two Wikipedia datasets \val{agent.selected.stop.repeated_action.wikipedia} of \val{agent.selected.claims.wikipedia} investigations end because the agent repeats
an action it has already taken, and on AVeriTeC all 100 investigations finish
explicitly while \val{agent.selected.incomplete.a} of them still list an open question.

\begin{figure}[htbp]
\centering
\includegraphics[width=\linewidth]{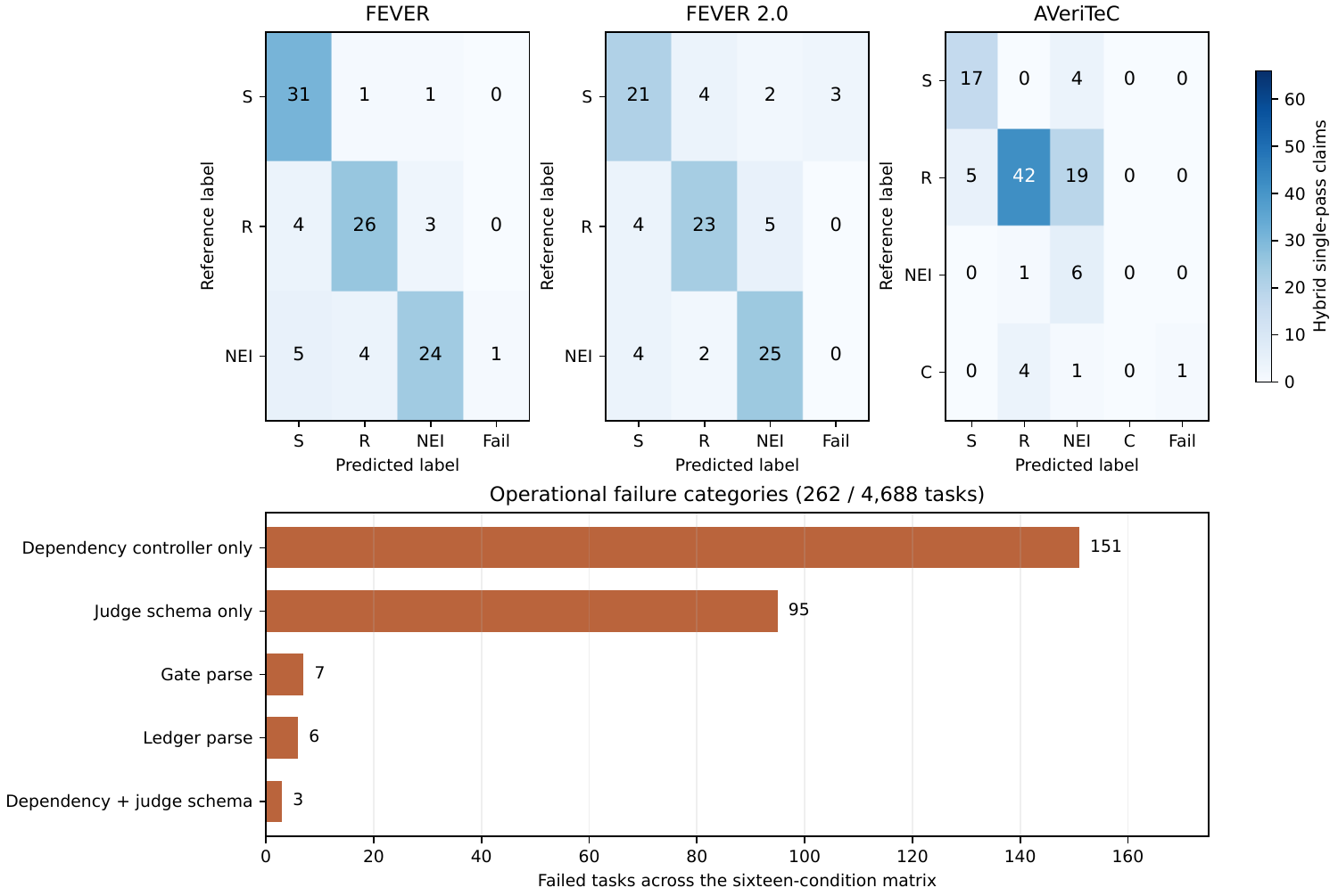}
\caption{Top: verdicts of hybrid single pass against the reference label, with
operational failures as a separate column (S supported, R refuted, C
conflicting evidence). Bottom: operational failures of the sixteen matrix
conditions by failing component.}
\label{fig:reliability}
\end{figure}

\paragraph{Reliability.}
The dependency controller is the one condition that is clearly worse. With
hybrid retrieval it repairs 5, 3 and 5 verdicts and damages 19, 24 and 25, a
loss of 14.0, 22.6 and 20.0 points whose intervals exclude zero on all three
datasets, at eight to ten calls per claim. Its three conditions account for
176 of the 262 failures in the matrix. By component, 154 failures arise in the
dependency controller, 98 in the judge's output schema, seven in the gate and
six in the ledger, three tasks failing in two components
(Figure~\ref{fig:reliability}). The controller has the most places to
fail: the plan, each intermediate answer and each binding to a source quotation
must be a valid structured reply. A task in which one of them fails counts as
a failure although the judge is still called, and in all 154 such tasks the
judge returned a verdict that the score does not use. The dependency
conditions also end with an extended judge prompt and a further validation
rule (Section~\ref{sec:controllers}), so the deficit combines planning,
failure scoring and a different verdict stage. The result is a
statement about this controller with this model; it does not show that
dependency planning cannot work.

\paragraph{Answer to RQ1.}
When the evidence is easy to find, one retrieval pass finds it, the retrieval
method hardly matters once a reranker is applied, and adaptive search control
adds between two and nine times the model calls for no gain in accuracy and, in
the case of the dependency controller, a substantial loss.

\subsection{RQ2: errors after the evidence has been found}\label{sec:rq2}
Hybrid single pass is wrong on 78 of the 293 claims. Sixteen of these are
Wikipedia claims labelled not enough information, fifteen with a decisive
verdict and one operational failure. Fifteen of the sixteen are among the 26
such claims whose store contains the annotated sentences of a related claim;
of the 39 with nothing on their subject in the store, one is wrong. The other 62 are evidence-bearing claims, and for them the evidence
trace gives a clear account of reference recovery
(Table~\ref{tab:location}): all have complete reference evidence in the
candidate pool, two follow a packet from which selection dropped part of the
evidence, four are operational failures, and 56 are wrong verdicts on a packet
that contains the complete reference evidence.

\begin{table}[htbp]
\centering\small
\begin{tabular}{lrrrr}
\toprule
Errors on evidence-bearing claims & FEVER & FEVER 2.0 & AVeriTeC & Total \\
\midrule
Reference evidence not in candidate pool & 0 & 0 & 0 & 0 \\
Reference evidence dropped by selection & 1 & 0 & 1 & 2 \\
Operational failure & 0 & 3 & 1 & 4 \\
Complete evidence in packet, wrong verdict & 8 & 15 & 33 & 56 \\
\midrule
Total & 9 & 18 & 35 & 62 \\
\bottomrule
\end{tabular}
\caption{Hybrid single pass in the annotation-selected stores: errors on
evidence-bearing claims by reference-evidence gap or operational failure,
with wrong verdicts on complete packets shown separately.}
\label{tab:location}
\end{table}

The most frequent of these downstream errors is an abstention against a
decisive reference label.
Thirty-four claims with a supporting or refuting reference end as not enough
information, 4 on FEVER, 7 on FEVER 2.0 and 23 on AVeriTeC, and 33 of the 34
have complete reference evidence in the packet. On AVeriTeC the consequence is
visible in the class totals (Figure~\ref{fig:reliability}): the system returns
not enough information for 30 claims, of which six carry that reference label
and 19 are refuted claims. Its accuracy of 65.0\% is below the 66\% of a
constant \emph{refuted} answer, although its balanced accuracy of 57.6\% is
far above the 25\% of that baseline.

\paragraph{Validation conversions.}
The saved model replies show that many of these abstentions are not the
judge's. Table~\ref{tab:validation} compares the judge's verdict before
validation with the final verdict. On the 293 claims the rules convert 24
verdicts to not enough information. Ten conversions are harmful, removing a
verdict that matched the reference; five are beneficial, turning a wrong
verdict on a not-enough-information claim into the right abstention;
nine are neutral, because the judge's verdict was wrong in a way the
conversion does not repair.

\begin{table}[htbp]
\centering\small\setlength{\tabcolsep}{4.5pt}
\begin{tabular}{lrrrrrrr}
\toprule
 & \multicolumn{4}{c}{Conversions} & \multicolumn{3}{c}{Effect} \\
\cmidrule(lr){2-5}\cmidrule(lr){6-8}
Rule & FEVER & FEVER 2.0 & AVeriTeC & Total & Harmful & Beneficial & Neutral \\
\midrule
Stance alignment & 3 & 2 & 4 & 9 & 3 & 1 & 5 \\
Attribution review & 1 & 0 & 9 & 10 & 7 & 1 & 2 \\
Conflict declaration & 0 & 0 & 5 & 5 & 0 & 3 & 2 \\
\midrule
Total & 4 & 2 & 18 & 24 & 10 & 5 & 9 \\
\midrule
Correct before validation & 80 & 69 & 71 & 220 & & & \\
Correct after validation & 81 & 69 & 65 & 215 & & & \\
\bottomrule
\end{tabular}
\caption{Verdict validation in hybrid single pass, annotation-selected stores.
A conversion replaces the judge's verdict by not enough information. The
coverage rule did not fire in this condition.}
\label{tab:validation}
\end{table}

The three rules behave differently. The conflict rule is the only one that
does no harm: all five verdicts it converts were wrong, and three conversions
produce the reference label. Stance alignment is close to neutral. Attribution
review is the costly rule. It is triggered for 15 claims, withholds permission
for ten, and in seven of those ten the withheld verdict was the reference
label. In two of the seven the claim is not an attribution at all. \emph{The
first case of coronavirus in Milwaukee County, Wisconsin, had been declared as
of March 11, 2020} is reviewed because it contains the word \emph{declared}, and the review
then rejects a refutation that rests on a passage stating there were no
confirmed cases on that date. \emph{35\% of revenue goes to states. (In
Nigeria)} is reviewed
because \emph{states} is read as a reporting verb.

The effect is concentrated on AVeriTeC, where validation lowers the number of
correct verdicts from 71 to 65, and it is stable across the comparison. In
each of the thirteen matrix conditions without the dependency controller,
validation costs between six and nine correct AVeriTeC verdicts; in the three
dependency conditions, where a further rule checks the dependency chain, it
costs thirteen. On FEVER and FEVER 2.0 the net effect lies between $-2$ and
$+2$ verdicts in those thirteen conditions. The validation rules were written
to stop the system from asserting what it cannot ground, and on the Wikipedia
claims they do so at no cost. On real-world claims, where the reference
verdict often rests on an inference from the evidence and many claims are
attributions, the same rules cost more correct verdicts than they gain.

\paragraph{The 33 residual abstentions.}
Table~\ref{tab:residual} classifies the 33 claims with complete evidence, a
decisive reference and a final abstention by where the abstention originated;
Appendix~\ref{app:cases} lists every claim. Eighteen are conversions and
fifteen are abstentions by the judge.

\begin{table}[htbp]
\centering\small
\begin{tabularx}{\linewidth}{lXrrrr}
\toprule
 & Origin of the abstention & FEVER & FEVER 2.0 & AVeriTeC & Total \\
\midrule
\multicolumn{6}{l}{\emph{Converted by validation}} \\
V1 & Verdict field contradicts the judge's own reasoning & 1 & 2 & 2 & 5 \\
V2 & Correct verdict, no passage assessed with the matching stance & 1 & 0 & 2 & 3 \\
V3 & Attribution review withholds the verdict & 0 & 0 & 8 & 8 \\
V4 & Conflict declared without grounded passages on both sides & 0 & 0 & 2 & 2 \\
\midrule
\multicolumn{6}{l}{\emph{Abstention by the judge}} \\
J1 & Declines to infer the verdict from an incompatible stated property & 2 & 3 & 0 & 5 \\
J2 & Leaves an unusual paraphrase unresolved & 0 & 2 & 0 & 2 \\
J3 & Does not accept evidence about the claim's origin as refutation & 0 & 0 & 4 & 4 \\
J4 & Does not take an arithmetic or evaluative step & 0 & 0 & 4 & 4 \\
\midrule
 & Total & 4 & 7 & 22 & 33 \\
\bottomrule
\end{tabularx}
\caption{Hybrid single pass: claims with a supporting or refuting reference,
complete reference evidence in the packet and a final verdict of not enough
information.}
\label{tab:residual}
\end{table}

Among the conversions, category V1 is the Labor Day pattern of the
introduction. In five cases the judge's reasoning reaches the reference
verdict, its passage stances agree with the reasoning, and the verdict field
states the opposite label. All five claims are refuted in the reference, and
in all five the verdict field says \emph{supported} although the reasoning
concludes that the claim is refuted; two of the claims are
themselves phrased with a negation (\emph{is incapable of}, \emph{is not the
approximate time at which}). Stance alignment catches the inconsistency, which is what it is for,
but it can only abstain; a repair that re-derives the verdict from the stances
would have produced the reference label in all five. In V2 the verdict is
right and the bookkeeping is not: in two cases two passages jointly support a
claim, the judge marks each as partial support, and no single passage carries
the required stance; in the third the reply contains no passage assessments. V3 holds the eight attribution reviews, seven of which
withhold a verdict that matches the reference.

The judge's own abstentions follow the dataset. On the Wikipedia datasets the
judge demands an explicit statement where the annotators accepted an
inference. A passage that describes \emph{Conquest of the Planet of the Apes}
as a science-fiction film does not, for the judge, refute that the film is
horror; a passage that defines benzodiazepines as a class of drugs does not
refute that there is a music genre of that name. Two of these FEVER 2.0 claims use the
phrase \emph{started living} for birth, and the judge treats the phrase as
ambiguous. On AVeriTeC half of the judge's abstentions concern provenance. The
evidence establishes that the claim originated on a satirical or fabricated
news site, the reference label is refuted, and the judge answers that the
passages say nothing about the event itself. The other half need a step the
judge does not take, such as dividing two stated counts to check a percentage.
The exploratory reading of these fifteen cases is that the judge applies a
stricter standard of proof than the reference annotation. Whether that
standard is wrong is a question of label policy; missing annotated evidence
does not explain these abstentions.

\paragraph{Answer to RQ2.}
Among the 33 residual abstentions with complete reference evidence, the
judge's own abstentions and validation conversions form groups of comparable
size. The fine categories suggest differences in accepted inference, while
the saved replies establish which verdicts validation changed. On AVeriTeC the
second source is the larger one, and most of its cost comes from a single
rule.

\subsection{RQ3: the same evidence, ten judges}\label{sec:rq3}
\begin{figure}[htbp]
\centering
\includegraphics[width=.92\linewidth]{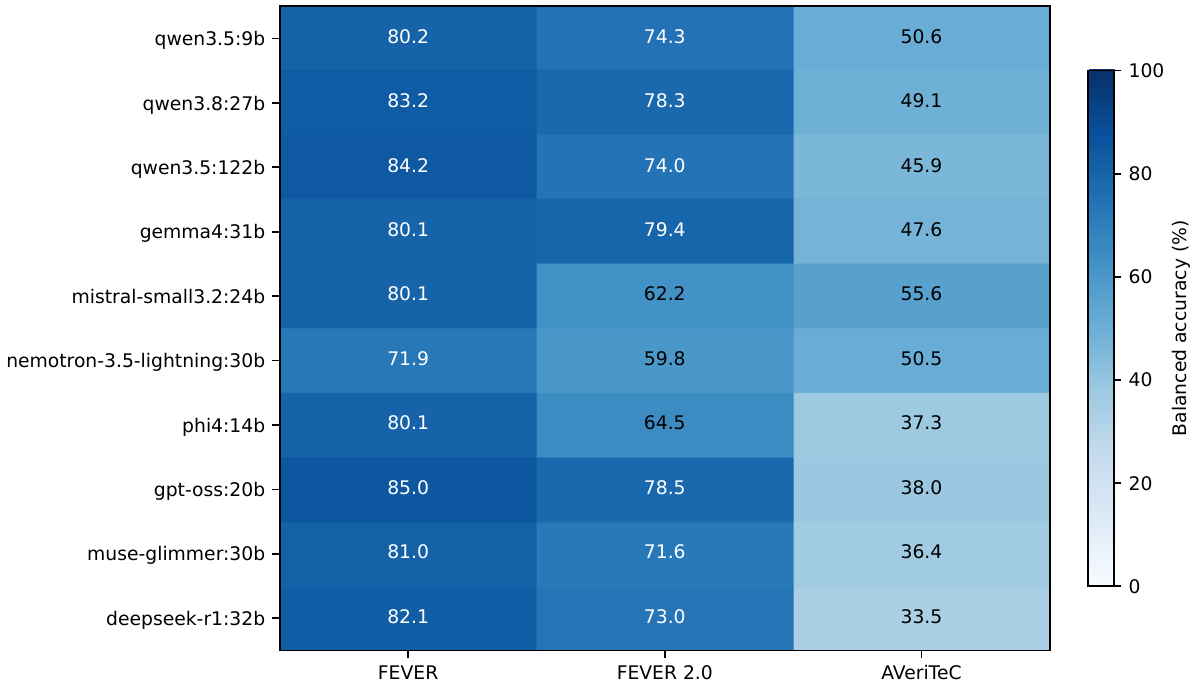}
\caption{Balanced accuracy of the ten judges on identical evidence packets.}
\label{fig:judges}
\end{figure}

Table~\ref{tab:judges} repeats the audit of Section~\ref{sec:rq2} for ten
judges that read identical packets, and Figure~\ref{fig:judges} shows their
balanced accuracy. Accuracy ranges from 72.0\% to 85.0\% on FEVER, from 60.2\%
to 79.6\% on FEVER 2.0 and from 48.0\% to 72.0\% on AVeriTeC. No judge leads on
all three datasets, and within the Qwen family the largest model is the best of
the three on FEVER and the weakest on AVeriTeC. Five judges complete every
task; the others fail on up to 22 of the 293 claims.

\input{tables/judges}

\paragraph{Abstentions recur across judges.}
Every judge turns claims with a supporting or refuting reference into
abstentions. The number ranges from 17 to 82 of the 215 such claims, that is
from 8\% to 38\%, and the 9B model of the main comparison has the lowest count
of the ten. A larger model does not abstain less: the 27B and 122B Qwen models
do so on 37 and 28 claims. Of the 476 such abstentions over all judges, 85 are
validation conversions and 391 are the judge's own verdict, so that for nine
of the ten judges the judge itself is the main source. The abstentions also
concentrate on particular claims. For 76 of the 215 claims no judge abstains,
and for 26 claims at least seven of the ten judges do; 15 of these 26 are
AVeriTeC claims. This concentration is consistent with shared difficulty in
the claims, packets or label interpretation. Because reference recovery in
these packets was not traced, it does not distinguish insufficient evidence
from a stricter standard of proof or a shared model error.

\paragraph{The cost of validation depends on the judge.}
Validation converts 114 verdicts over the ten judges. Fifty-seven conversions
are harmful, 22 beneficial and 35 neutral. The net effect is negative for
seven judges, zero for one and positive for two, and it ranges from four
correct verdicts gained to fifteen lost. The loss of fifteen belongs to one
judge whose supported verdicts often come with a reported coverage below the
required level; the coverage rule alone converts 15 of its verdicts. For the
9B model, validation gains four verdicts in this block, where the attribution
review was not yet part of the judge. This agrees with the main comparison,
in which the three rules without the attribution review are close to neutral
for that model (three harmful and four beneficial conversions,
Table~\ref{tab:validation}) and the attribution review accounts for the loss.
A validation rule therefore has no fixed cost. Its cost depends on how
reliably a given model fills the fields the rule inspects.

\paragraph{The contradiction between reasoning and verdict is general.}
The Labor Day pattern is not confined to the 9B model. In 27 conversions the
verdict field names one decisive label while every passage the judge assessed
with confidence points to the other. This occurs in eight of the ten judges.
In 19 of the 27 cases the label implied by the stances is the reference label,
and this holds for all 14 cases in the three Qwen models, four of them in the
122B model.

\paragraph{Answer to RQ3.}
Abstentions and validation conversions persist when different models read
the same packets, and inconsistent replies occur across model families.
Every judge abstains on claims with a decisive reference label, and many of
these abstentions fall on the same claims. The 9B model has the lowest such
count in this block. Since packet sufficiency was not audited, these counts
do not establish errors despite complete evidence or show that larger models
cannot improve judgement. What changes from judge to judge is the size of
each observed effect.

\subsection{RQ4: the same claims against the full stores}\label{sec:rq4}
The full stores test whether these findings depend on easy discovery. The
claims, the model, the candidate limit and the packet budget are those of
Section~\ref{sec:rq1}; only the store changes.

\begin{figure}[htbp]
\centering
\includegraphics[width=\linewidth]{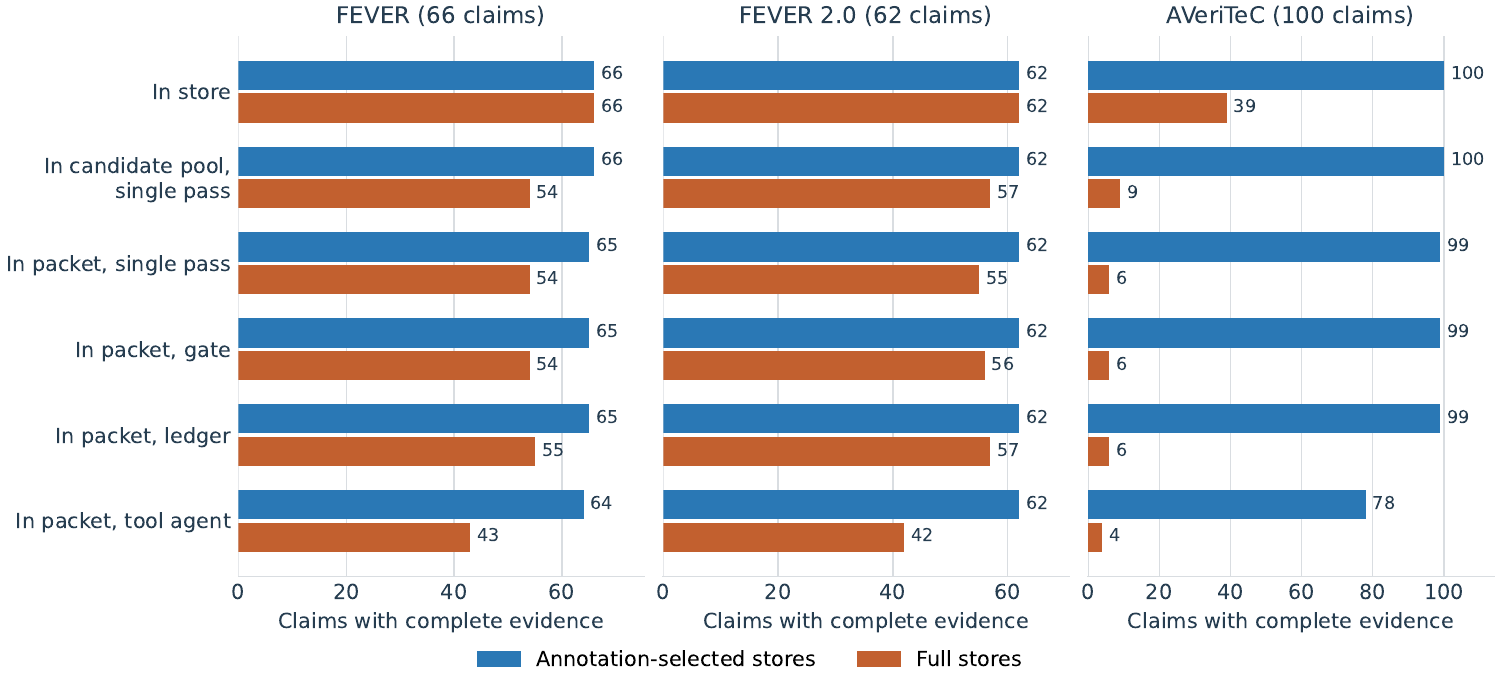}
\caption{Evidence-bearing claims whose complete reference evidence is present
at each stage, in the annotation-selected and in the full stores. For AVeriTeC
in the full stores evidence is counted at the level of the cited page.}
\label{fig:store-recovery}
\end{figure}

\paragraph{Evidence recovery.}
The reference evidence is harder to recover in the full stores, and the two
sources differ in how much (Table~\ref{tab:full-recovery}). In the Wikipedia
snapshot every evidence-bearing claim has a complete annotated set. Hybrid
single pass brings it into the candidate pool for 54 of 66 FEVER claims and 57
of 62 FEVER 2.0 claims, and into the packet for 54 and 55, against 65 and 62 in
the selected stores. Dense single pass ends with the same counts within two
claims (54 and 57). Sparse single pass falls further, to 43 and 43: lexical
search alone loses the evidence of 11 FEVER and 14 FEVER 2.0 claims that hybrid
search keeps. For AVeriTeC, evidence is counted at the level of the cited page.
Only 39 of the 99 claims that cite a page have every cited page in their
collection. Hybrid single pass has all of them in the candidate pool for nine
claims and in the packet for six. In the selected stores all annotated
question--answer units reached the packet for 99 claims. These AVeriTeC
counts measure different objects: cited-page coverage and annotated-answer
recovery. The controllers hardly change the respective recovery counts. Against hybrid
single pass the gate completes the evidence of one further FEVER 2.0 claim;
the ledger completes it for two FEVER and three FEVER 2.0 claims and loses it
for one of each. Neither changes a single AVeriTeC claim. The tool agent ends
with less: complete evidence is in its packet for 43 FEVER, 42 FEVER 2.0 and
four AVeriTeC claims, although it had come across it during the investigation
for 48, 50 and nine (Figure~\ref{fig:store-recovery}).

\begin{table}[htbp]
\centering\small\setlength{\tabcolsep}{4pt}
\begin{tabular}{lrrrrrr}
\toprule
 & \multicolumn{3}{c}{Selected} & \multicolumn{3}{c}{Full} \\
\cmidrule(lr){2-4}\cmidrule(lr){5-7}
Complete evidence & F & F2 & AV & F & F2 & AV \\
\midrule
\input{generated/rq4-full-recovery-rows.tex} 
\bottomrule
\end{tabular}
\caption{Complete recovery of the reference evidence with hybrid retrieval and
by the tool agent in the two evidence conditions. F: FEVER; F2: FEVER 2.0;
AV: AVeriTeC. Wikipedia recovery matches annotated sentences in both stores;
AVeriTeC matches annotated answers in the selected stores and cited pages
in the full stores. Full-store AVeriTeC recovery is assessable for 99 of the
100 evidence-bearing claims.}
\label{tab:full-recovery}
\end{table}

\begin{figure}[htbp]
\centering
\includegraphics[width=\linewidth]{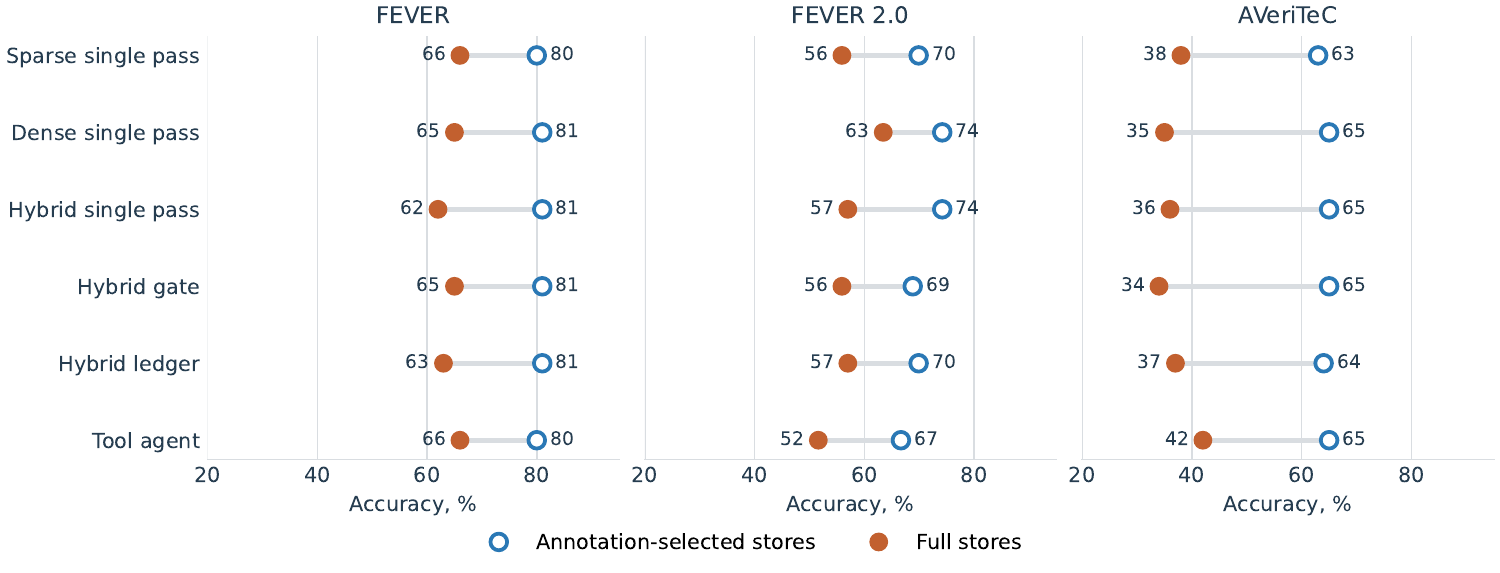}
\caption{Accuracy of each condition on the same claims in the
annotation-selected stores (open markers) and in the full stores (filled
markers).}
\label{fig:store-accuracy}
\end{figure}

\paragraph{Accuracy and cost.}
Accuracy falls in every condition and on every dataset
(Table~\ref{tab:full}, Figure~\ref{fig:store-accuracy}). For hybrid single pass the paired change from the
selected to the full store is $-19.0$ points on FEVER (95\% interval $-30.3$
to $-9.8$), $-17.2$ on FEVER 2.0 ($-29.7$ to $-4.4$) and $-29.0$ on AVeriTeC
($-41.0$ to $-17.0$). Over the six conditions the change lies between $-14$
and $-19$ points on FEVER, $-11$ and $-17$ on FEVER 2.0 and $-23$ and $-31$ on
AVeriTeC; only two intervals reach zero, those of the dense single pass and the
gate on FEVER 2.0. No retrieval method leads consistently. Sparse single pass
is the most accurate on FEVER and AVeriTeC, with 66 and 38 per cent, and dense
single pass on FEVER 2.0 with 63.4, and the three single-pass conditions lie
within four points of one another except for dense on FEVER 2.0. Sparse search
is thus no less accurate on FEVER although it recovers complete evidence for
eleven fewer claims. Of the 1,465 tasks of the five matrix conditions 41 end
in an operational failure, none of them in retrieval, against 29 for the same
conditions in the selected stores; the agent fails on \val{agent.full.failures.all} of its 293 claims,
against \val{agent.selected.failures.all}. Model calls per claim stay as they were for the matrix conditions:
about one for single pass, 2.3 to 2.7 for the gate and 4.6 for the ledger. Those
of the agent rise to \val{agent.full.calls_per_claim.f}, \val{agent.full.calls_per_claim.ft} and \val{agent.full.calls_per_claim.a}, from \val{agent.selected.calls_per_claim.f}, \val{agent.selected.calls_per_claim.ft} and \val{agent.selected.calls_per_claim.a}.

On the Wikipedia datasets the loss is almost entirely a loss on claims whose
reference label is not enough information. Hybrid single pass is correct on 56
of 66 evidence-bearing FEVER claims in the full store and on 57 in the selected
store, and on 45 against 44 of the 62 evidence-bearing FEVER 2.0 claims. On the
claims labelled not enough information it is correct on 6 of 34 against 24 for
FEVER and on 8 of 31 against 25 for FEVER 2.0. In the selected stores 39 of
these 65 claims had nothing on their subject in the store. The full snapshot
holds the pages on their subject, and the judge then returns a decisive
verdict: 48 of the 51 errors on these claims are supported or refuted verdicts
and three are failures. On AVeriTeC the loss falls on the decisive classes:
20 of 66 refuted claims are correct against 42, and 14 of 21 supported claims
against 17.

\begin{table}[htbp]
\centering\small\setlength{\tabcolsep}{4.5pt}
\begin{tabular}{lrrrrrrrrr}
\toprule
 & \multicolumn{3}{c}{Selected, \%} & \multicolumn{3}{c}{Full, \%} & \multicolumn{3}{c}{Calls, full} \\
\cmidrule(lr){2-4}\cmidrule(lr){5-7}\cmidrule(lr){8-10}
Condition & F & F2 & AV & F & F2 & AV & F & F2 & AV \\
\midrule
\input{generated/rq4-full-rows.tex} 
\bottomrule
\end{tabular}
\caption{Accuracy in the two evidence conditions and model calls per claim in
the full stores. F: FEVER; F2: FEVER 2.0; AV: AVeriTeC.}
\label{tab:full}
\end{table}

\paragraph{Does adaptive search help when discovery is hard?}
It does not, within the resolution of this comparison
(Table~\ref{tab:interaction}). In the full stores the gate repairs three
FEVER verdicts of hybrid single pass and damages none, repairs two and damages
three on FEVER 2.0, and damages two on AVeriTeC without a repair. The ledger
repairs four and damages three on FEVER, seven and seven on FEVER 2.0, and
seven and six on AVeriTeC. The paired differences lie between $-2$ and $+3$
points. The interaction, the controller's gain in the full store minus its
gain in the selected store, is at most $+4.3$ points and its interval includes
zero for both controllers on all three datasets. Unlike in the selected
stores, the controllers now do change what the judge reads. The ledger's final
judge request equals that of single pass for 25 FEVER, 18 FEVER 2.0 and 14
AVeriTeC claims, where it was 69, 66 and 99; for the gate the counts are 94,
79 and 72. The ledger therefore sends a different packet for about four claims
in five and repairs as many verdicts as it damages. Differences of this size
are no larger than those between two executions of hybrid single pass itself
(Table~\ref{tab:repeat}).
Executed a second time on the full stores, it received an identical judge
request for \val{repeat.full.identical_final_request.all} of 293 claims and returned a different verdict for \val{repeat.full.verdict_changed.all}. It
repaired \val{repeat.full.wrong_to_correct.f} FEVER verdicts and damaged \val{repeat.full.correct_to_wrong.f}, repaired \val{repeat.full.wrong_to_correct.ft} and damaged \val{repeat.full.correct_to_wrong.ft}
on FEVER 2.0, and repaired \val{repeat.full.wrong_to_correct.a} and damaged \val{repeat.full.correct_to_wrong.a} on AVeriTeC, so that its
accuracy was \val{repeat.full.second_accuracy.f}, \val{repeat.full.second_accuracy.ft} and \val{repeat.full.second_accuracy.a} per cent where the first execution gave
\val{repeat.full.first_accuracy.f}, \val{repeat.full.first_accuracy.ft} and \val{repeat.full.first_accuracy.a}. The exact McNemar
probability is 0.25 or more for the repair and damage counts of the gate and
the ledger, at least \val{repeat.full.mcnemar_p.f} for those of the repeated execution, and 0.26 or
more for those of the tool agent below.

The tool agent is the one condition whose search changes visibly with the
store. In the selected stores \val{agent.selected.stop.controller_budget.all} of its 293 investigations ended at the step
limit; in the full stores \val{agent.full.stop.controller_budget.all} do, \val{agent.full.stop.repeated_action.all} end because the agent repeats an action
and \val{agent.full.stop.finished.all} because it declares itself finished. It issues \val{agent.full.calls.all} model calls where
it issued \val{agent.selected.calls.all}, and hands the judge five to eight passages per claim where it
handed about two. Its accuracy is \val{agent.full.accuracy.f}, \val{agent.full.accuracy.ft} and \val{agent.full.accuracy.a} per cent. Against hybrid
single pass in the full stores it repairs \val{agent.full.repairs.f} verdicts and damages \val{agent.full.damages.f} on
FEVER, repairs \val{agent.full.repairs.ft} and damages \val{agent.full.damages.ft} on FEVER 2.0, and repairs \val{agent.full.repairs.a} and damages
\val{agent.full.damages.a} on AVeriTeC: \val{agent.full.difference.f}, \val{agent.full.difference.ft} and \val{agent.full.difference.a} points, each with an interval
that includes zero (\val{agent.full.low.f} to \val{agent.full.high.f}, \val{agent.full.low.ft} to \val{agent.full.high.ft}, \val{agent.full.low.a} to \val{agent.full.high.a}).
Its interaction with the store is the largest of the three controllers, \val{agent.interaction.a}
points on AVeriTeC, with an interval from \val{agent.interaction.low.a} to \val{agent.interaction.high.a}. The agent is thus the
most accurate condition on AVeriTeC, level with sparse single pass on FEVER
and the least accurate on FEVER 2.0, at eight to eleven model calls per
claim against one. Where it is ahead, it does not recover more reference evidence: its
packets hold complete reference evidence for fewer claims than those of single
pass on all three datasets. A large part of its margin lies on claims without
a decisive reference. On FEVER it is correct on 12 of the 34 claims labelled
not enough information, against six for single pass, and on the same number
of supported claims and two fewer refuted ones. On AVeriTeC it is correct on
five of seven such claims against one, on two of six conflicting claims
against one, on three more refuted claims and two fewer supported ones.

\begin{table}[htbp]
\centering\small
\begin{tabular}{llrrr}
\toprule
Controller & Dataset & Selected, pp & Full, pp & Interaction, pp [95\% interval] \\
\midrule
\input{generated/rq4-interaction-rows.tex} 
\bottomrule
\end{tabular}
\caption{Accuracy change of the adaptive controllers and the tool agent
against hybrid single pass in each evidence condition, and the difference
between the two.}
\label{tab:interaction}
\end{table}

\paragraph{Where the errors arise.}
More errors accompany gaps in reference recovery in the full stores
(Table~\ref{tab:full-location}). Hybrid single pass makes 90 errors on
evidence-bearing claims in the full stores, against 62. For 37 of them the
reference evidence is not in the store, for 28 it is in the store and absent
from the candidate pool, three lose it at selection and two are operational
failures. Twenty errors occur with complete reference recovery in the packet,
where the selected stores had 56; for AVeriTeC this means cited-page coverage,
not verified passage sufficiency. The increase in errors comes from AVeriTeC. The number of
errors on evidence-bearing Wikipedia claims is 27 in both stores, and all 37
errors without evidence in the store are AVeriTeC claims with a cited page
missing from the collection. These categories locate gaps in the annotated
reference path, not proven causes of the wrong verdicts: alternative evidence
could support a decision even where a cited page is missing. Validation converts 37 verdicts in the full
stores: 20 conversions are harmful, six beneficial and eleven neutral, against
ten, five and nine. The cost on AVeriTeC persists when the evidence is page
text and grows: 14 harmful conversions and one beneficial, where the selected
stores gave nine and three.

\begin{table}[htbp]
\centering\small
\begin{tabular}{lrr}
\toprule
Errors on evidence-bearing claims, hybrid single pass & Selected & Full \\
\midrule
\input{generated/rq4-full-location-rows.tex} 
\bottomrule
\end{tabular}
\caption{Errors classified by reference-evidence gaps and operational
failures, all three datasets together. AVeriTeC recovery matches annotated
answers in the selected stores and cited pages in the full stores; the
columns therefore do not measure an unchanged notion of sufficiency.}
\label{tab:full-location}
\end{table}

The AVeriTeC loss also occurs where all cited pages are present
(Table~\ref{tab:in-store}). On the 39 claims whose collection holds every
cited page, hybrid single pass is correct on 26 in the selected stores and on
13 in the full stores, a paired change of $-33.3$ points ($-51.3$ to $-15.4$).
On the 60 claims with a cited page missing it is correct on 38 and 23, a
change of $-25.0$ points ($-41.7$ to $-8.3$). The six conditions are correct
on 12 to 16 of the 39 claims and on 21 to 26 of the 60. The observed accuracy
drop is thus at least as large where all cited pages are present as where one
is missing; these subsets do not isolate a causal effect of page availability. On the
39 claims the cited pages are all in the candidate pool for eight to 15 and
all in the packet for four to seven, so the evidence that is present is
rarely delivered, and on the 60 claims more than a third of the verdicts are
correct although a cited page cannot be delivered at all.

\begin{table}[htbp]
\centering\small
\begin{tabular}{lrrrrr}
\toprule
 & \multicolumn{3}{c}{All cited pages in store (39)} & \multicolumn{2}{c}{A cited page missing (60)} \\
\cmidrule(lr){2-4}\cmidrule(lr){5-6}
Condition & Selected, \% & Full, \% & In packet & Selected, \% & Full, \% \\
\midrule
\input{generated/rq4-in-store-rows.tex} 
\bottomrule
\end{tabular}
\caption{AVeriTeC accuracy in the two evidence conditions, by whether the
claim's full collection holds every cited page. In packet: claims with every
cited page in the packet in the full stores.}
\label{tab:in-store}
\end{table}

\paragraph{One model instance.}
To reduce reloads as a possible source of execution variation
(Section~\ref{sec:rq1}), the five matrix conditions and the tool agent were
executed once more on the full stores with equal batch sizes for all calls,
so that the server loaded the model once and served all 5,612 calls from that
instance. Table~\ref{tab:one-instance} gives the controllers against single
pass. The three single-pass conditions reproduce the first execution: the
replies of hybrid single pass are identical for 292 of 293 claims. The
comparison of the controllers changes in one respect. In the first execution
the gate sent the judge a request identical to that of single pass for
\val{many.adaptive_gate.identical_final_request.all} claims, and the correctness of the verdict nevertheless changed for \val{many.adaptive_gate.changes_on_identical_request.all} of them,
half of its ten repairs and damages. With one instance this happens for \val{one.adaptive_gate.changes_on_identical_request.all} of
\val{one.adaptive_gate.identical_final_request.all} such claims, and for none of the ledger's \val{one.question_ledger.identical_final_request.all}.
Variation on identical requests is reduced but not eliminated; neither the
effect of evidence nor the role of prompt caching is isolated by this run.
The gate then repairs \val{one.adaptive_gate.repairs.all} verdicts and damages \val{one.adaptive_gate.damages.all}
over the three datasets, \val{one.adaptive_gate.difference.f}, \val{one.adaptive_gate.difference.ft} and \val{one.adaptive_gate.difference.a} points; the interval on FEVER
2.0 excludes zero (\val{one.adaptive_gate.low.ft} to \val{one.adaptive_gate.high.ft}), while the exact McNemar probability of
the pooled split is \val{one.adaptive_gate.mcnemar_p.all}. The ledger repairs \val{one.question_ledger.repairs.all} and damages \val{one.question_ledger.damages.all}, and the tool
agent repairs \val{one.tool_agent.repairs.all} and damages \val{one.tool_agent.damages.all} (\val{one.tool_agent.difference.f}, \val{one.tool_agent.difference.ft} and \val{one.tool_agent.difference.a} points, each interval
including zero). The gate shows a small gain in this execution, whereas the
first comparison did not resolve one; the ledger and the agent stay without a
resolved difference, and no controller recovers more than a few points of
the loss between the stores.

\begin{table}[htbp]
\centering\footnotesize
\begin{tabular}{llrrrrrr}
\toprule
 & & First, & \multicolumn{5}{c}{One model instance} \\
\cmidrule(lr){4-8}
Controller & Dataset & pp & Repair & Damage & $\Delta$ pp [95\% interval] & Same req. & changed \\
\midrule
\input{generated/one-instance-rows.tex} 
\bottomrule
\end{tabular}
\caption{Controllers against hybrid single pass in the full stores, in the
first execution and in the execution served by one loaded instance of the
model. \emph{Same request}: claims for which the controller's final judge
request is identical to that of single pass; \emph{changed}: how many of
them changed in correctness.}
\label{tab:one-instance}
\end{table}

\paragraph{A larger cohort of unused claims.}
The Wikipedia comparison of single pass and the tool agent was repeated on
\val{fresh.claims.all} claims that had not been used before (Section~\ref{sec:setup}), again
against the full store. Table~\ref{tab:fresh} gives the result. Hybrid
single pass is correct on \val{fresh.single.accuracy.f} per cent of the FEVER claims and \val{fresh.single.accuracy.ft} per cent
of the FEVER 2.0 claims, close to its results on the first cohort, and its
second execution, served by the same instance of the model, returned the same
reply for every claim. The tool agent reaches \val{fresh.agent.accuracy.f} and \val{fresh.agent.accuracy.ft} per cent at \val{fresh.agent.calls_per_claim.f} and \val{fresh.agent.calls_per_claim.ft} model
calls per claim. It repairs \val{fresh.agent.repairs.f} verdicts and damages \val{fresh.agent.damages.f} on FEVER and repairs
\val{fresh.agent.repairs.ft} and damages \val{fresh.agent.damages.ft} on FEVER 2.0; over all \val{fresh.claims.all} claims the difference is \val{fresh.agent.difference.all}
points with an interval from \val{fresh.agent.low.all} to \val{fresh.agent.high.all}. The signs on the two datasets
are those of the first cohort and both intervals include zero, so five times
as many FEVER 2.0 claims do not turn the agent's deficit there into a resolved
difference, and over the two datasets together they exclude a gain of more
than about two points. The pattern behind the figures is also the same. The
agent's packets hold complete reference evidence for \val{fresh.agent.packet.f} of \val{fresh.evidence_bearing.f} and \val{fresh.agent.packet.ft} of
\val{fresh.evidence_bearing.ft} evidence-bearing claims, against \val{fresh.single.packet.f} and \val{fresh.single.packet.ft} for single pass, and it is
correct on more claims labelled not enough information (\val{fresh.agent.correct.nei.f} against \val{fresh.single.correct.nei.f} of
\val{fresh.class.nei.f}, and \val{fresh.agent.correct.nei.ft} against \val{fresh.single.correct.nei.ft} of \val{fresh.class.nei.ft}) and on fewer decisive ones. The FEVER part
of this cohort is small in independent material: its \val{fresh.claims.f} claims come from
\val{fresh.families.f} claim families. In each condition about five per cent of the tasks
failed because the judge's reply reached the output limit while it was still
weighing its verdict (\val{fresh.single.failures.all} tasks of single pass, \val{fresh.single.failures.ft} of them on FEVER 2.0, and
\val{fresh.agent.failures.all} of the agent); they are counted as errors. Had the verdict stated at
the start of each such reply been accepted, single pass would gain \val{fresh.single.stated_correct.all} correct
claims and the agent \val{fresh.agent.stated_correct.all}, which leaves the comparison as it is.

\begin{table}[htbp]
\centering\footnotesize
\begin{tabular}{lrrrrrrrr}
\toprule
 & & & \multicolumn{2}{c}{Accuracy, \%} & Calls & & & \\
Dataset & Claims & Families & single & agent & per claim & Repair & Damage & $\Delta$ pp [95\% interval] \\
\midrule
\input{generated/fresh-rows.tex} 
\bottomrule
\end{tabular}
\caption{Hybrid single pass and the tool agent on 714 previously unused
Wikipedia claims against the full store. Repairs, damages and the difference
are those of the agent against single pass; intervals resample claim
families.}
\label{tab:fresh}
\end{table}

\paragraph{Answer to RQ4.}
Checked against the full stores, the same claims are verified with 11 to 31
points lower accuracy. Complete annotated sentence sets reach the packet for
54 FEVER and 55 FEVER 2.0 claims, against 65 and 62 in the selected stores.
For AVeriTeC all cited pages reach six packets in the full stores, whereas all
annotated answers reach 99 in the selected stores; these are different
recovery measures. Adaptive control does not recover the accuracy difference:
the gate and the ledger change the evidence the judge reads, complete it for
at most three further claims per dataset, and repair about as many verdicts as
they damage; served by one model instance, the gate gains one to four points
and the ledger has no resolved gain. The tool agent searches longest, at eight to eleven model calls
per claim, delivers complete evidence for fewer claims than single pass, and
ends four and six points above it on FEVER and AVeriTeC and five below on
FEVER 2.0, differences this cohort does not resolve; on 714 further Wikipedia
claims it ends \val{fresh.agent.deficit.all} points below single pass, again without a resolved
difference. On Wikipedia the total number of errors on evidence-bearing
claims is unchanged, while the accuracy loss falls on claims labelled not
enough information. We did not audit whether the additional decisive
verdicts on those claims were grounded. On AVeriTeC most errors accompany
missing or unretrieved cited pages, and a large accuracy loss also occurs on
claims whose cited pages are all present. This identifies poor reference-page
recovery alongside the verdict loss, without assigning that loss solely to
retrieval.

%% file: generated/rq1-agent-row.tex
Tool agent & 80.0 (3) & 66.7 (7) & 65.0 (2) & 6.44 & 7.16 & 4.34 \\

%% file: tables/paired.tex
\begin{table}[htbp]
\centering\small
\begin{tabular}{llrrrr}
\toprule
Procedure & Dataset & Repair & Damage & $\Delta$ pp & 95\% interval \\
\midrule
Hybrid gate & FEVER & 2 & 2 & +0.0 & [-4.2, +3.8] \\
Hybrid gate & FEVER 2.0 & 1 & 6 & -5.4 & [-10.5, +0.0] \\
Hybrid gate & AVeriTeC & 1 & 1 & +0.0 & [-3.0, +3.0] \\
Hybrid ledger & FEVER & 0 & 0 & +0.0 & [+0.0, +0.0] \\
Hybrid ledger & FEVER 2.0 & 1 & 5 & -4.3 & [-9.1, +0.0] \\
Hybrid ledger & AVeriTeC & 1 & 2 & -1.0 & [-5.0, +2.0] \\
Hybrid dependency & FEVER & 5 & 19 & -14.0 & [-23.2, -5.3] \\
Hybrid dependency & FEVER 2.0 & 3 & 24 & -22.6 & [-32.6, -12.9] \\
Hybrid dependency & AVeriTeC & 5 & 25 & -20.0 & [-30.0, -10.0] \\
\input{generated/rq1-agent-paired-rows.tex} \bottomrule
\end{tabular}
\caption{Paired change against hybrid single pass in the annotation-selected stores: repairs, damages, accuracy difference in percentage points and its 95\% family-bootstrap interval.}
\label{tab:paired}
\end{table}

%% file: generated/rq1-agent-paired-rows.tex
Tool agent & FEVER & 4 & 5 & -1.0 & [-7.4, +5.4] \\
Tool agent & FEVER 2.0 & 4 & 11 & -7.5 & [-16.0, +1.1] \\
Tool agent & AVeriTeC & 2 & 2 & +0.0 & [-4.0, +4.0] \\

%% file: tables/judges.tex
\begin{table}[htbp]
\centering\small\setlength{\tabcolsep}{4.5pt}
\begin{tabular}{lrrrrrrrrrr}
\toprule
 & \multicolumn{3}{c}{Accuracy, \%} & & \multicolumn{2}{c}{S/R $\rightarrow$ \nei{}} & \multicolumn{4}{c}{Validation} \\
\cmidrule(lr){2-4}\cmidrule(lr){6-7}\cmidrule(lr){8-11}
Judge & F & F2 & AV & Fail & all & conv. & harm. & benef. & neutr. & net \\
\midrule
\texttt{qwen3.5:9b} & 80.0 & 74.2 & 72.0 & 8 & 17 & 10 & 2 & 6 & 9 & $+4$ \\
\texttt{qwen3.8:27b} & 83.0 & 78.5 & 68.0 & 0 & 37 & 10 & 5 & 1 & 5 & $-4$ \\
\texttt{qwen3.5:122b} & 84.0 & 74.2 & 66.0 & 11 & 28 & 8 & 4 & 1 & 5 & $-3$ \\
\texttt{gemma4:31b} & 80.0 & 79.6 & 60.0 & 2 & 50 & 7 & 7 & 1 & 0 & $-6$ \\
\texttt{mistral-small3.2:24b} & 80.0 & 62.4 & 64.0 & 0 & 44 & 9 & 8 & 1 & 2 & $-7$ \\
\texttt{nemotron-3.5-lightning:30b} & 72.0 & 60.2 & 53.0 & 0 & 82 & 3 & 2 & 2 & 2 & $+0$ \\
\texttt{phi4:14b} & 80.0 & 64.5 & 56.0 & 0 & 60 & 5 & 2 & 5 & 4 & $+3$ \\
\texttt{gpt-oss:20b} & 85.0 & 78.5 & 60.0 & 13 & 43 & 4 & 3 & 1 & 1 & $-2$ \\
\texttt{muse-glimmer:30b} & 81.0 & 72.0 & 54.0 & 0 & 75 & 21 & 17 & 2 & 4 & $-15$ \\
\texttt{deepseek-r1:32b} & 82.0 & 73.1 & 48.0 & 22 & 40 & 8 & 7 & 2 & 3 & $-5$ \\
\bottomrule
\end{tabular}
\caption{Ten judges on identical evidence packets. \emph{Fail}: operational
failures over the 293 claims. \emph{S/R $\rightarrow$} \nei{}: claims with a
supporting or refuting reference (215 in total) that end as not enough
information, and how many of them through a validation conversion.
\emph{Validation}: harmful, beneficial and neutral conversions and the net
change in correct verdicts. F: FEVER; F2: FEVER 2.0; AV: AVeriTeC.}
\label{tab:judges}
\end{table}

%% file: generated/rq4-full-recovery-rows.tex
Evidence-bearing claims & 66 & 62 & 100 & 66 & 62 & 100 \\
In store & 66 & 62 & 100 & 66 & 62 & 39 \\
In candidate pool, single pass & 66 & 62 & 100 & 54 & 57 & 9 \\
In packet, single pass & 65 & 62 & 99 & 54 & 55 & 6 \\
In packet, gate & 65 & 62 & 99 & 54 & 56 & 6 \\
In packet, ledger & 65 & 62 & 99 & 55 & 57 & 6 \\
In packet, tool agent & 64 & 62 & 78 & 43 & 42 & 4 \\

%% file: generated/rq4-full-rows.tex
Sparse single pass & 80.0 & 69.9 & 63.0 & 66.0 & 55.9 & 38.0 & 1.02 & 1.00 & 1.07 \\
Dense single pass & 81.0 & 74.2 & 65.0 & 65.0 & 63.4 & 35.0 & 1.02 & 1.00 & 1.11 \\
Hybrid single pass & 81.0 & 74.2 & 65.0 & 62.0 & 57.0 & 36.0 & 1.02 & 1.00 & 1.10 \\
Hybrid gate & 81.0 & 68.8 & 65.0 & 65.0 & 55.9 & 34.0 & 2.26 & 2.34 & 2.71 \\
Hybrid ledger & 81.0 & 69.9 & 64.0 & 63.0 & 57.0 & 37.0 & 4.62 & 4.57 & 4.56 \\
Tool agent & 80.0 & 66.7 & 65.0 & 66.0 & 51.6 & 42.0 & 7.73 & 9.24 & 10.88 \\

%% file: generated/rq4-interaction-rows.tex
Hybrid gate & FEVER & $+0.0$ & $+3.0$ & $+3.0$ [-1.0, +8.0] \\
Hybrid gate & FEVER 2.0 & $-5.4$ & $-1.1$ & $+4.3$ [-2.2, +10.8] \\
Hybrid gate & AVeriTeC & $+0.0$ & $-2.0$ & $-2.0$ [-6.0, +2.0] \\
Hybrid ledger & FEVER & $+0.0$ & $+1.0$ & $+1.0$ [-5.0, +7.1] \\
Hybrid ledger & FEVER 2.0 & $-4.3$ & $+0.0$ & $+4.3$ [-5.4, +14.6] \\
Hybrid ledger & AVeriTeC & $-1.0$ & $+1.0$ & $+2.0$ [-6.0, +10.0] \\
Tool agent & FEVER & $-1.0$ & $+4.0$ & $+5.0$ [-3.5, +13.3] \\
Tool agent & FEVER 2.0 & $-7.5$ & $-5.4$ & $+2.2$ [-11.2, +14.7] \\
Tool agent & AVeriTeC & $+0.0$ & $+6.0$ & $+6.0$ [-3.0, +16.0] \\

%% file: generated/rq4-full-location-rows.tex
Reference evidence not in store & 0 & 37 \\
Reference evidence not in candidate pool & 0 & 28 \\
Reference evidence dropped by selection & 2 & 3 \\
Operational failure & 4 & 2 \\
Complete evidence in packet, wrong verdict & 56 & 20 \\
\midrule
Total & 62 & 90 \\
\midrule
Validation conversions: harmful / beneficial / neutral & 10 / 5 / 9 & 20 / 6 / 11 \\

%% file: generated/rq4-in-store-rows.tex
Sparse single pass & 66.7 & 35.9 & 7 & 60.0 & 40.0 \\
Dense single pass & 66.7 & 30.8 & 6 & 63.3 & 38.3 \\
Hybrid single pass & 66.7 & 33.3 & 6 & 63.3 & 38.3 \\
Hybrid gate & 66.7 & 30.8 & 6 & 63.3 & 36.7 \\
Hybrid ledger & 66.7 & 41.0 & 6 & 61.7 & 35.0 \\
Tool agent & 66.7 & 41.0 & 4 & 63.3 & 43.3 \\

%% file: generated/one-instance-rows.tex
Hybrid gate & FEVER & $+3.0$ & 2 & 1 & $+1.0$ [-2.3, +4.7] & 92 & 1 \\
Hybrid gate & FEVER 2.0 & $-1.1$ & 4 & 0 & $+4.3$ [+1.0, +8.6] & 81 & 0 \\
Hybrid gate & AVeriTeC & $-2.0$ & 1 & 0 & $+1.0$ [+0.0, +3.0] & 75 & 0 \\
Hybrid ledger & FEVER & $+1.0$ & 3 & 3 & $+0.0$ [-4.7, +5.0] & 24 & 0 \\
Hybrid ledger & FEVER 2.0 & $+0.0$ & 8 & 10 & $-2.2$ [-11.1, +6.9] & 18 & 0 \\
Hybrid ledger & AVeriTeC & $+1.0$ & 10 & 6 & $+4.0$ [-4.0, +12.0] & 12 & 0 \\
Tool agent & FEVER & $+4.0$ & 11 & 5 & $+6.0$ [-2.1, +14.3] & 0 & 0 \\
Tool agent & FEVER 2.0 & $-5.4$ & 6 & 9 & $-3.2$ [-12.0, +5.6] & 0 & 0 \\
Tool agent & AVeriTeC & $+6.0$ & 17 & 10 & $+7.0$ [-3.0, +17.0] & 0 & 0 \\

%% file: generated/fresh-rows.tex
FEVER & 216 & 89 & 63.4 & 65.3 & 8.6 & 20 & 16 & $+1.9$ [-4.0, +8.1] \\
FEVER 2.0 & 498 & 438 & 52.4 & 49.4 & 8.9 & 55 & 70 & $-3.0$ [-7.7, +1.6] \\
All & 714 & 527 & 55.7 & 54.2 & 8.8 & 75 & 86 & $-1.5$ [-5.2, +2.1] \\

%% file: sections/discussion.tex
\section{Discussion}\label{sec:discussion}
\subsection{More search is not more evidence}
In the control condition there is nothing left for a second search to find.
One retrieval pass delivers the complete reference evidence for all but two
claims; the gate and the ledger end with exactly as much of it in the packet
as single pass, and the tool agent with less. The result agrees with
cost-aware studies of search agents for question answering, where accuracy
plateaus after a few searches \citep{mccleary2026quantifying} and a
considerable share of search steps is unnecessary \citep{wu2025search}, and
with the observation that forced additional searches do not improve claim
verification \citep{xie2025fire}. It adds the measurement of the evidence
itself, which those studies do not make. For the same reason the paired
changes of the gate and the ledger carry no information about control: nearly
all of them occur on claims where the judge received the request it received
under single pass.

The comparison also shows what such a result does not license. Graph-assisted
retrieval ties with dense retrieval because it delivers the same packets, and
the tie says nothing about graphs. A controller can be judged only where it
has an opportunity to change the evidence, and the annotation-selected stores
offer almost none.
The full stores supply the opportunity. There
the ledger sends the judge a packet different from that of single pass for
about four claims in five, and the gate for one in six. Neither gains in the
first execution: the
paired differences lie between $-2$ and $+3$ points and the interaction with
the store has an interval that includes zero on every dataset. When the
comparison is repeated with one loaded instance of the model, variation on
identical requests is reduced. The gate repairs \val{one.adaptive_gate.repairs.all} verdicts and
damages \val{one.adaptive_gate.damages.all}, a gain of one to four points at more than twice the model
calls, and the ledger has a small net gain with intervals that include zero.
Neither execution shows these controllers recovering the full accuracy loss. A
second search had something to find, since single pass missed complete
evidence for 19 Wikipedia claims that have it in the store, but the follow-up
searches completed it for at most three claims per dataset. The effects are
also of the size of the four-point difference involving passage order and
execution variation in Section~\ref{sec:rq1} and
of the two to five points by which a second execution of single pass differs
from the first in the full stores, so small effects require care: the gate's
FEVER 2.0 interval excludes zero in the one-instance run, but the other
controller intervals do not. The observed gains fall short of the 11 to 31
points lost across conditions with the larger stores. Recovery and verdict
audits are needed to distinguish contributions from the retriever and the judge.

The tool agent qualifies this without overturning it. It is the only
controller that responds to the harder store by searching more: its
investigations run to the step limit nine times as often and cost eight to
eleven model calls per claim. The additional search does not bring more of the
reference evidence to the judge. The agent meets complete evidence during the
investigation for no more claims than single pass retrieves, and keeps it in
the packet for fewer. Its verdicts are nevertheless four points better on
FEVER and six on AVeriTeC and five worse on FEVER 2.0, and its interaction
with the store on AVeriTeC, \val{agent.interaction.points.a} points with an interval from \val{agent.interaction.low.a} to \val{agent.interaction.high.a}, is
the nearest this study comes to a sign that a controller may be worth more
when discovery is hard. The sign is weak. On 714 further Wikipedia claims the
agent ended \val{fresh.agent.deficit.all} points below single pass, with an interval from \val{fresh.agent.low.all} to
\val{fresh.agent.high.all}, so on Wikipedia a larger cohort confirms neither a gain nor a loss;
AVeriTeC, where the sign appeared, has no such second cohort. A second execution of single pass moved its own full-store
accuracy by four, five and two points on the three datasets, which is the
size of the agent's margins, and the class counts suggest a different
reading from better discovery: much of the agent's margin lies on claims whose
reference is not enough information or conflicting evidence, and not on the
claims for which evidence had to be found. A comparison with more claims per
dataset would be needed to decide between the two readings.

\subsection{Abstention has two authors}
The second finding concerns the verdict \emph{not enough information}. It is
usually read as the judge's assessment of the evidence, and work on sufficiency
and abstention studies it as such
\citep{joren2025sufficient,madhusudhan2025abstention}. In the system studied
here it has two authors. In 15 of the 33 residual abstentions the judge
declines to decide. In 18 the judge decides and a rule overrides it. The two
cases need different remedies, and the usual confusion matrix cannot tell them
apart. Both occur with every one of the ten judges we compared, in different
proportions: for most judges the larger share is their own abstention, and the
model used in the main comparison abstains least.

The exploratory case reading points to the standard of proof. The judge
wants a passage that states the refutation, where the annotators accept that a
film described as science fiction is not horror, or that a claim traced to a
satirical site is false. Neither position is unreasonable. A fact-checker that
refuses to treat provenance as refutation is cautious in a way a human
fact-checker would not be, and a prompt or a label policy that states which
inferences are permitted could address these cases. The annotated passage is
already present, so its absence does not explain the abstention. The fixed-packet
comparison does not show that larger models abstain less: the 27B and 122B
judges abstain on more decisive claims than the 9B judge, and 26 claims draw an
abstention from at least seven of ten judges. However, those packets were not
audited for sufficiency, and the comparison cannot establish that more
evidence or a larger model would not help an individual case.

The conversions are a matter of system design. Validation rules exist so that
a verdict can be traced to a passage, and on the Wikipedia claims they achieve
this without loss. On AVeriTeC they turn a pre-validation accuracy of 71\% into
65\%, and the pattern repeats in every condition of the comparison. Across ten
judges the rules remove a correct verdict more than twice as often as they
produce one. Two
observations follow. The first is that a rule which can only abstain throws
away information the reply already contains. When the verdict field says
\emph{supported} and every assessed passage says \emph{refutes}, the reply is
inconsistent, but it is not uninformative; deriving the verdict from the
stances would have recovered the reference label in all five such cases, and
in 19 of the 27 such cases across the ten judges. The
second is that a rule needs a trigger as careful as its check. The attribution
review asks a sensible question, whether the passages establish who said what
and when, but it is switched on by a list of reporting verbs, applies a
standard under which only an explicit denial can refute an attribution, and so
blocks refutations that rest on a satirical origin or on a contradicting
figure. Its cost is seven correct verdicts out of ten interventions.

These rules are one implementation, and a trigger that fires on a list of
verbs can fairly be called a defect to repair and not a property of
validation. We report it as a case for that reason. The defect was invisible
in end-to-end accuracy, where it appeared as a judge that abstains too often,
and it became visible only when the verdict was recorded before and after
validation.

We do not conclude that validation should be removed. Accuracy against a
benchmark label is not the only purpose of a fact-checker, and a verdict that
cannot be tied to a passage is of little use to a reader however often it is
right. The conclusion is narrower: a validation layer is a component with its
own error rate, and that rate should be measured and reported like any other.
In this system the measurement changes the reading of the AVeriTeC results
substantially, since what looked like a judge that over-abstains on refuted
claims is in part a judge that refutes and a rule that takes the refutation
back.

\subsection{Implications for evaluation}
Four practices follow from the study. A comparison of search procedures
should report evidence recovery at the packet next to accuracy, because a tie
in accuracy means something different when the packets are identical and when
they differ. It should report model calls and operational failures per
condition, because the procedures differ far more in these than in accuracy.
And a pipeline that post-processes its judge should report the verdict before
and after post-processing. None of this requires new annotation; all of it was
obtained here from the evidence packets and model replies saved during the
run.

A fourth concerns the interpretation of the reference data. FEVER defines
not enough information relative to Wikipedia and allowed annotators to search
beyond the initially displayed pages; FEVER 2.0 retains that task definition
\citep{thorne2018fever,thorne2019fever2}. The full Wikipedia store therefore
does not by itself invalidate those labels. Nearly the entire accuracy loss
on Wikipedia falls on this class, but we did not assess the grounding of the
additional decisive predictions. The result establishes lower agreement with
the reference labels, not whether the predictions expose annotation errors.
For AVeriTeC, the full-store audit matches passages to cited page addresses.
It shows whether a page was reached, not whether the relevant text within it
was delivered. These distinctions matter when interpreting a recovery score
as evidence sufficiency or a verdict disagreement as an annotation problem.
Whether the store itself should carry more structure, such as entities,
subject--predicate--object relations and source metadata, is a separate
question that this study leaves open. The retrieval losses on AVeriTeC, where
all cited pages reach the packet for six of the 39 claims that have them, mark
the place where such structure would have to prove itself; the one
graph-assisted condition we ran delivered the packets of dense retrieval and
was not run on the full stores. Structure in the store would in any case not
directly repair a contradiction between a judge's passage assessments and its
verdict field.

\subsection{Limitations}
The main comparison has about one hundred claims per dataset and one model.
The main tables report one execution per task; repeats cover hybrid single
pass, the selected-store agent and the full-store conditions. Only the
Wikipedia comparison of single pass and the agent was repeated on a larger
cohort. Differences of a few points
between conditions are within the variation that passage order and a repeated
execution produce: two executions of hybrid single pass differ in the
correctness of three to five claims per dataset in the selected stores and of
four to nine in the full stores, on identical judge requests. Server logs
suggest prompt-cache state as a possible source of variation, but this was
not tested by a controlled intervention and one repeat remains unexplained
(Section~\ref{sec:rq1}). Differing batch sizes made the server reload the
model between calls in the main comparison. The repetition of the full-store
conditions with one instance reduced, but did not eliminate, changes in
correctness on identical requests. That execution control was
not repeated for the selected stores, the judges of
Section~\ref{sec:rq3} or the further cohort's agent run, whose agent calls
used the earlier batch size. In the first executions only the dependency
controller's deficit has intervals excluding zero among the controller
comparisons. In the one-instance run the gate's FEVER 2.0 interval also
excludes zero, although its pooled exact McNemar probability is
\val{one.adaptive_gate.mcnemar_p.all}. The intervals are not
corrected for multiple comparisons, and the exact McNemar test resolves a
difference only from a net eight to twelve claims in a hundred
(Section~\ref{sec:design}); in the first executions, for the gate, the ledger
and the tool agent its probability is at least 0.11 in either store. The absence of a gain is thus an
absence at this resolution. The 293 claims were used during development,
so both their absolute accuracies and the component comparisons may reflect
development exposure. Its direction and size are unknown.

The AVeriTeC cohort has seven and six claims in its two smallest classes.
Hybrid single pass predicts none of the six conflicting-evidence claims
correctly, which says little with six claims. Complete recovery is defined through the annotations:
it shows that the annotated evidence reached the judge, and it does not
certify that this evidence entails the label or that the label is beyond
dispute. The finer categories of the 33 residual abstentions are exploratory
interpretations proposed with Claude, without independent human validation.
The split between conversions and the judge's abstentions is computed from
the saved replies and does not depend on those categories.

The annotation-selected stores are a favourable extreme. For AVeriTeC they
contain the annotators' answers, which are written with the verdict in view.
The full stores remove this advantage, but they remain closed collections: no
condition searches the open web, deals with documents published after the
claim, or is exposed to text written to mislead a retriever. The agent was run
with a longer time limit than the matrix, and model calls leave out the cost
of embedding and reranking, which grows with the store. The agent's run
on the selected stores was first executed on a different machine from its
run on the full stores, and was then repeated on the same machine; the
repeat is the run reported here. The first execution gave
\val{agent.first_host.selected.accuracy.f}, \val{agent.first_host.selected.accuracy.ft} and \val{agent.first_host.selected.accuracy.a} per cent against \val{agent.selected.accuracy.f}, \val{agent.selected.accuracy.ft} and \val{agent.selected.accuracy.a}; \val{host.verdict_changed.all} of 293 verdicts
differ, and \val{host.different_request.all} claims reached the judge with a different request. Computed
with the first execution, the agent's loss between the stores would be \val{agent.first_host.store_loss.f}, \val{agent.first_host.store_loss.ft} and
\val{agent.first_host.store_loss.a} points instead of \val{agent.store_loss.f}, \val{agent.store_loss.ft} and \val{agent.store_loss.a}, and its interaction with the store
on AVeriTeC \val{agent.first_host.interaction.a} points (\val{agent.first_host.interaction.low.a} to \val{agent.first_host.interaction.high.a}) instead of \val{agent.interaction.a}. The
conclusions on the agent do not depend on which of the two runs is used
(Table~\ref{tab:agent-host}).
The full-store comparison has limits of its own. For 60 of the 99 AVeriTeC
claims at least one cited page is absent from the collection, so their
reference evidence cannot be recovered completely by any method, and presence
is counted for whole pages. These page-level counts cannot establish passage
sufficiency or be directly equated with selected-store answer recovery.
We did not manually assess the additional decisive predictions on Wikipedia
claims labelled not enough information, so their disagreement with the labels
cannot be attributed to annotation errors. The dependency controller, graph-assisted retrieval and the
conditions without reranking were not run on the full stores.

Finally, the search controllers were run with a 9B model only. The judge
comparison shows abstentions and validation conversions across models,
but it says nothing about control: a larger model may plan and bind more
reliably, which would change the dependency controller's result. The judge
comparison has the restrictions stated in Section~\ref{sec:judges-design} and
supports comparisons between judges, not with the matrix.

\section{Conclusion}\label{sec:conclusion}
We separated retrieval, search control and judgement in an evidence-grounded
fact-checker, followed the reference evidence from the store to the verdict,
and asked four questions.

\emph{RQ1.} When the evidence is easy to find, neither the retrieval method nor
the controller improves on a single pass. One hybrid retrieval pass delivers
the complete reference evidence for 226 of 228 evidence-bearing claims; after
reranking, dense and hybrid retrieval agree on the correctness of every claim,
and no method is better on all three datasets. An adaptive gate, a question
ledger, a dependency controller and a tool agent add between two and nine
times the model calls without improving accuracy. The dependency controller is
clearly worse and is also the least reliable: its conditions account for 176
of the 262 operational failures.

\emph{RQ2.} The errors that remain lie after retrieval, and they arise in both
places. Of the 62 errors on evidence-bearing claims 56 occur with the complete
reference evidence in the packet. The largest group, abstentions against
decisive reference labels, divides between the judge, whose abstentions
suggest a difference in accepted inference, and the validation rules, which
convert decisive verdicts into abstentions; on real-world claims the rules
cost more correct verdicts than they gain.

\emph{RQ3.} These patterns persist across models. Ten judges of eight families
all abstain on decisive claims given the same packets, validation costs more
correct verdicts than it gains for seven of them, and inconsistent replies
occur across families. Packet sufficiency was not audited in that block, so it
does not establish that those abstentions occur despite complete evidence.

\emph{RQ4.} Against the full stores the answer to RQ1 holds and the answer to
RQ2 shifts. Accuracy falls by 17 to 29 points for hybrid single pass and by 11
to 31 points over all six conditions, and no controller recovers the loss: in
the one-instance execution the gate gains one to four points and the ledger
and the tool agent have no resolved gain, and the agent shows none on 714
further Wikipedia claims either. Errors are no longer mostly downstream. Hybrid single pass
makes 90 errors on evidence-bearing claims against 62; for 37 the reference
evidence is not in the store, for 28 it is in the store and not retrieved, and
20 occur with complete recovery, which for AVeriTeC means coverage of cited
pages and not verified sufficiency. Validation still costs more than it gains,
with 20 harmful conversions against ten. On Wikipedia the loss is concentrated
on claims labelled not enough information, which without an audit of those
predictions establishes reduced agreement with the labels and not a defect in
them.

For the builder of such a system the practical conclusion is to locate an
error before treating it. Evidence recovery at the packet, the judge's verdict
before validation and the final verdict are three different measurements, they
are cheap to keep, and in this study they distinguish failures that an
end-to-end accuracy score alone cannot explain.

These results come from cohorts and evidence stores that we prepared from
public datasets, and should not be generalised beyond them. The largest effect
came from the store and not from the search procedure, and a store that is
updated continuously moves from the first condition towards the second. For
such a store the results suggest two requirements, neither of which was tested
here: reference evidence recorded at the level of the passage and not of the
page, and a fixed set of claims on which recovery, judgement and validation
are measured again after every update.

\section*{Data and code availability}
FEVER, FEVER 2.0 and AVeriTeC, including the Wikipedia snapshot and the
archived web collections, are public; their sources are cited in
Section~\ref{sec:related}. The claim identifiers of all cohorts, the prompts,
the per-claim outcomes with their evidence packets and model replies, and the
analysis code that produces every table and figure of this paper are
available from the corresponding author on reasonable request.

\section*{Declaration of competing interest}
The authors declare that they have no competing interests.

\section*{Acknowledgements}
\ifanonymous
Withheld for review.
\else
This work was supported by the project ``Biography of Fake News with a Touch
of AI: Dangerous Phenomenon through the Prism of Modern Human Sciences'',
project no.\ CZ.02.01.01/00/23\_025/0008724, funded through the Operational
Programme Jan \'{A}mos Komensk\'{y}.
\fi

\section*{Declaration of generative AI and AI-assisted technologies in the manuscript preparation process}
During the preparation of this work the authors used ChatGPT (OpenAI) and
Claude (Anthropic) to assist with drafting, translation into English,
revision and coherence checking. Claude's use in proposing qualitative
case categories is described in Section~\ref{sec:design}. After using these tools, the authors reviewed
and edited the content as needed and take full responsibility for the content
of the published article.

%% file: back.tex
\bibliography{references}
\appendix
\input{sections/appendix}

%% file: sections/appendix.tex
\section{All conditions in the annotation-selected stores}\label{app:matrix}
\input{tables/full-matrix}

\section{Residual abstentions}\label{app:cases}
Table~\ref{tab:cases} lists the claims summarised in Table~\ref{tab:residual}.
The column \emph{First} and the assignment to the V or the J group are read
from the saved model replies. The category within each group is an
exploratory assignment proposed with Claude from the judge's justification,
passage assessments and evidence packet, as described in
Section~\ref{sec:design}.

\input{tables/cases}

\section{Judge comparison by dataset}\label{app:judges}
\input{tables/judges-detail}

\section{Repeated executions}\label{app:repeat}
Table~\ref{tab:repeat} compares two executions of hybrid single pass on the
same claims, stores and settings. Table~\ref{tab:agent-host} compares the two
executions of the tool agent on the annotation-selected stores, which ran on
different machines, and gives the agent's store contrast computed with each.

\begin{table}[htbp]
\centering\small
\begin{tabular}{llrrrrrrrr}
\toprule
 & & & \multicolumn{2}{c}{Correct} & Same & Verdict & Correct & Wrong & \\
Store & Dataset & Claims & first & second & request & changed & to wrong & to correct & $p$ \\
\midrule
\input{generated/repeat-rows.tex} 
\bottomrule
\end{tabular}
\caption{Hybrid single pass executed twice. \emph{Same request}: claims whose
final judge request is identical in both executions; $p$: exact McNemar
probability of the split between the last two columns.}
\label{tab:repeat}
\end{table}

\begin{table}[htbp]
\centering\footnotesize
\begin{tabular}{lrrrrrrr}
\toprule
 & \multicolumn{2}{c}{Accuracy, \%} & Verdict & \multicolumn{2}{c}{Store change, pp} & \multicolumn{2}{c}{Interaction, pp [95\% interval]} \\
Dataset & first & second & changed & first & second & first & second \\
\midrule
\input{generated/agent-host-rows.tex} 
\bottomrule
\end{tabular}
\caption{The tool agent on the annotation-selected stores, executed on two
machines. \emph{Store change} and \emph{interaction} are those of
Section~\ref{sec:rq4}, computed with the first and with the second execution;
the second ran on the machine that served the full stores.}
\label{tab:agent-host}
\end{table}

\section{Model and store identities}\label{app:identities}
Table~\ref{tab:identities} gives the identities of the models used in the
main comparison; the ten judges of Section~\ref{sec:rq3} are identified by
their tags in Table~\ref{tab:judges}. The
language model is served by Ollama; generation uses temperature zero and seed
42, and the judge prompt is version \texttt{grounded-judge/v8}; the
dependency conditions use its extension
\texttt{grounded-judge/v8/dependencies-v4}.

\begin{table}[htbp]
\centering\footnotesize
\begin{tabularx}{\linewidth}{lX}
\toprule
Component & Identity \\
\midrule
Language model & \texttt{qwen3.5:9b}, \texttt{Q4\_K\_M} quantisation \\
 & \texttt{6488c96fa5faab64bb65cbd30d4289e20e6130ef535a93ef9a49f42eda893ea7} \\
Encoder & \texttt{nomic-ai/nomic-embed-text-v1-ablated} \\
 & revision \texttt{7d948905c5d5d3874fa55a925d68e49dbf411e5f} \\
Reranker & \texttt{cross-encoder/ms-marco-MiniLM-L-6-v2} \\
 & revision \texttt{c5ee24cb16019beea0893ab7796b1df96625c6b8} \\
\bottomrule
\end{tabularx}
\caption{Model identities.}
\label{tab:identities}
\end{table}

The full Wikipedia store holds 20,845,882 passages, each with one
768-dimensional vector of the encoder above and one full-text index row
(SHA-256 of the store file \texttt{ddb1f843794f2d40\ldots 96c0b21a}). The
hundred AVeriTeC collections hold 54,967,218 passages, between 100,220 and
1,989,235 per claim. Dense search over the full stores is exact cosine search
over all vectors, read from a matrix on disk; no approximate index is used.

%% file: tables/full-matrix.tex
\begin{table}[htbp]
\centering\small
\begin{tabular}{lrrr}
\toprule
Procedure & FEVER & FEVER 2.0 & AVeriTeC \\
\midrule
Gate / dense + rerank & 81.0 / 4 / 246 & 69.9 / 2 / 235 & 65.0 / 1 / 213 \\
Gate / hybrid + rerank & 81.0 / 4 / 247 & 68.8 / 2 / 235 & 65.0 / 1 / 213 \\
Gate / sparse + rerank & 77.0 / 4 / 235 & 68.8 / 4 / 210 & 62.0 / 1 / 218 \\
Direct dense & 81.0 / 2 / 102 & 71.0 / 1 / 93 & 61.0 / 1 / 113 \\
Direct dense + rerank & 81.0 / 2 / 102 & 74.2 / 3 / 93 & 65.0 / 1 / 113 \\
Dependency / dense + rerank & 67.0 / 16 / 844 & 51.6 / 27 / 738 & 45.0 / 15 / 983 \\
Dependency / hybrid + rerank & 67.0 / 16 / 844 & 51.6 / 27 / 734 & 45.0 / 15 / 983 \\
Dependency / sparse + rerank & 68.0 / 17 / 830 & 48.4 / 27 / 742 & 42.0 / 16 / 985 \\
Dense + graph + rerank & 81.0 / 2 / 102 & 74.2 / 3 / 93 & 65.0 / 1 / 113 \\
Ledger / dense + rerank & 81.0 / 2 / 442 & 69.9 / 2 / 412 & 64.0 / 3 / 461 \\
Ledger / hybrid + rerank & 81.0 / 2 / 442 & 69.9 / 2 / 412 & 64.0 / 3 / 461 \\
Ledger / sparse + rerank & 82.0 / 2 / 438 & 68.8 / 4 / 411 & 62.0 / 3 / 459 \\
Single pass / dense + rerank & 81.0 / 1 / 102 & 74.2 / 3 / 93 & 65.0 / 1 / 113 \\
Single pass / hybrid + rerank & 81.0 / 1 / 102 & 74.2 / 3 / 93 & 65.0 / 1 / 113 \\
Single pass / sparse + rerank & 80.0 / 1 / 102 & 69.9 / 3 / 93 & 63.0 / 1 / 113 \\
Direct sparse & 83.0 / 2 / 102 & 68.8 / 5 / 93 & 62.0 / 2 / 113 \\
\input{generated/rq1-agent-matrix-row.tex} \bottomrule
\end{tabular}
\caption{All conditions in the annotation-selected stores. Each cell gives accuracy (\%) / operational failures / model calls. The datasets have 100, 93 and 100 claims. \emph{Direct} conditions call the retrieval method without a controller; direct sparse and direct dense use no reranker.}
\label{tab:full-matrix}
\end{table}

%% file: generated/rq1-agent-matrix-row.tex
Tool agent & 80.0 / 3 / 644 & 66.7 / 7 / 666 & 65.0 / 2 / 434 \\

%% file: tables/cases.tex
\begin{table}[htbp]
\centering\footnotesize\setlength{\tabcolsep}{4pt}
\begin{tabularx}{\linewidth}{llllX}
\toprule
Claim & Ref. & First & Cat. & Observation \\
\midrule
\texttt{fever1-225196} & R & NEI & J1 & Does not infer refutation from the stated release information \\
\texttt{fever1-9514} & R & NEI & J1 & Science-fiction film; does not exclude horror \\
\texttt{fever1-213921} & R & S & V1 & Reasoning concludes refutation; verdict field says SUPPORTED \\
\texttt{fever1-75552} & S & S & V2 & Correct SUPPORTED from two passages each assessed PARTIAL \\
\texttt{fever2-500642} & R & NEI & J1 & Drug class; does not exclude a music genre of that name \\
\texttt{fever2-500914} & R & NEI & J1 & Novel; does not exclude a bar of that name \\
\texttt{fever2-500984} & S & NEI & J1 & Actress; does not support the negated brand claim \\
\texttt{fever2-500084} & R & NEI & J2 & Reads ``started living'' as ambiguous despite the birth date \\
\texttt{fever2-500882} & S & NEI & J2 & Does not map ``started living'' to birth \\
\texttt{fever2-500329} & R & S & V1 & Reasoning concludes refutation of a negated claim; verdict field says SUPPORTED \\
\texttt{fever2-500849} & R & S & V1 & Reasoning concludes refutation; verdict field says SUPPORTED \\
\texttt{averitec-train-1124} & R & NEI & J3 & Edited image and satirical origin not accepted as refutation \\
\texttt{averitec-train-1266} & R & NEI & J3 & Fabricated-news origin not accepted as refutation \\
\texttt{averitec-train-2269} & R & NEI & J3 & Satirical origin not accepted as refutation \\
\texttt{averitec-train-656} & R & NEI & J3 & Satirical origin not accepted as refutation \\
\texttt{averitec-train-1944} & S & NEI & J4 & Does not compute the share from the two stated counts \\
\texttt{averitec-train-385} & R & NEI & J4 & No regional figure exists; does not refute the stated figure \\
\texttt{averitec-train-404} & R & NEI & J4 & Does not weigh listed provisions against a qualitative claim \\
\texttt{averitec-train-515} & R & NEI & J4 & Does not judge ``very strict'' from the described rules \\
\texttt{averitec-train-394} & R & S & V1 & Reasoning concludes refutation; verdict field says SUPPORTED \\
\texttt{averitec-train-966} & R & S & V1 & Reasoning concludes refutation; verdict field says SUPPORTED \\
\texttt{averitec-train-2817} & R & R & V2 & Correct REFUTED without passage assessments \\
\texttt{averitec-train-757} & S & S & V2 & Correct SUPPORTED from two passages each assessed PARTIAL \\
\texttt{averitec-train-1127} & R & R & V3 & Correct REFUTED; review withheld (claim is not an attribution) \\
\texttt{averitec-train-1492} & R & R & V3 & Correct REFUTED; review withheld as unresolved \\
\texttt{averitec-train-2180} & R & R & V3 & Correct REFUTED from satire origin; review withheld \\
\texttt{averitec-train-2233} & R & R & V3 & Correct REFUTED; review withheld as unresolved \\
\texttt{averitec-train-2475} & R & R & V3 & Correct REFUTED from satire origin; review withheld \\
\texttt{averitec-train-557} & R & R & V3 & Correct REFUTED from chronology; review withheld \\
\texttt{averitec-train-6} & R & R & V3 & Correct REFUTED; review withheld (claim is not an attribution) \\
\texttt{averitec-train-638} & S & R & V3 & REFUTED against a SUPPORTS reference; review withheld \\
\texttt{averitec-train-2198} & R & C & V4 & CONFLICTING for a two-part claim; one side not grounded \\
\texttt{averitec-train-692} & S & C & V4 & CONFLICTING; passages assessed PARTIAL only \\
\bottomrule
\end{tabularx}
\caption{The 33 hybrid single-pass claims with a supporting or refuting reference, complete reference recovery and a final \nei{} verdict. \emph{First} is the judge's verdict before validation. Categories are defined in Table~\ref{tab:residual}.}
\label{tab:cases}
\end{table}

%% file: tables/judges-detail.tex
\begin{table}[htbp]
\centering\small
\begin{tabular}{lrrr}
\toprule
Judge & FEVER & FEVER 2.0 & AVeriTeC \\
\midrule
\texttt{qwen3.5:9b} & 78 / 80 / 2 & 68 / 69 / 8 & 71 / 72 / 7 \\
\texttt{qwen3.8:27b} & 86 / 83 / 3 & 73 / 73 / 4 & 69 / 68 / 4 \\
\texttt{qwen3.5:122b} & 84 / 84 / 0 & 70 / 69 / 3 & 68 / 66 / 7 \\
\texttt{gemma4:31b} & 84 / 80 / 4 & 74 / 74 / 0 & 62 / 60 / 4 \\
\texttt{mistral-small3.2:24b} & 82 / 80 / 4 & 61 / 58 / 4 & 66 / 64 / 3 \\
\texttt{nemotron-3.5-lightning:30b} & 72 / 72 / 0 & 56 / 56 / 0 & 53 / 53 / 6 \\
\texttt{phi4:14b} & 77 / 80 / 5 & 60 / 60 / 5 & 56 / 56 / 1 \\
\texttt{gpt-oss:20b} & 86 / 85 / 1 & 74 / 73 / 1 & 60 / 60 / 3 \\
\texttt{muse-glimmer:30b} & 83 / 81 / 2 & 74 / 67 / 10 & 60 / 54 / 11 \\
\texttt{deepseek-r1:32b} & 82 / 82 / 2 & 68 / 68 / 0 & 53 / 48 / 10 \\
\bottomrule
\end{tabular}
\caption{Fixed-packet judge comparison by dataset: correct verdicts before
validation / correct verdicts after validation / conversions. The datasets
have 100, 93 and 100 claims.}
\label{tab:judges-detail}
\end{table}

%% file: generated/repeat-rows.tex
Selected & FEVER & 100 & 81 & 80 & 100 & 6 & 3 & 2 & 1.000 \\
Selected & FEVER 2.0 & 93 & 69 & 68 & 93 & 8 & 2 & 1 & 1.000 \\
Selected & AVeriTeC & 100 & 65 & 64 & 100 & 6 & 2 & 1 & 1.000 \\
Selected & All & 293 & 215 & 212 & 293 & 20 & 7 & 4 & 0.549 \\
Full & FEVER & 100 & 62 & 66 & 100 & 5 & 0 & 4 & 0.125 \\
Full & FEVER 2.0 & 93 & 53 & 48 & 93 & 11 & 7 & 2 & 0.180 \\
Full & AVeriTeC & 100 & 36 & 38 & 99 & 16 & 3 & 5 & 0.727 \\
Full & All & 293 & 151 & 152 & 292 & 32 & 10 & 11 & 1.000 \\

%% file: generated/agent-host-rows.tex
FEVER & 80.0 & 80.0 & 4 & $-14.0$ & $-14.0$ & $+5.0$ [-3.3, +13.7] & $+5.0$ [-3.5, +13.3] \\
FEVER 2.0 & 68.8 & 66.7 & 8 & $-17.2$ & $-15.1$ & $+0.0$ [-12.6, +12.1] & $+2.2$ [-11.2, +14.7] \\
AVeriTeC & 61.0 & 65.0 & 16 & $-19.0$ & $-23.0$ & $+10.0$ [+0.0, +21.0] & $+6.0$ [-3.0, +16.0] \\